%% file: neurips_2026.tex
\documentclass{article}

\usepackage[nonatbib,preprint]{neurips_2026}
\usepackage[numbers,sort&compress]{natbib}

\usepackage[utf8]{inputenc} 
\usepackage[T1]{fontenc}    
\usepackage{hyperref}       
\usepackage{url}            
\usepackage{booktabs}       
\usepackage{amsfonts}       
\usepackage{nicefrac}       
\usepackage{microtype}      
\usepackage{xcolor}         
\definecolor{gold}{HTML}{D4AF37}
\definecolor{silver}{HTML}{C0C0C0}
\definecolor{bronze}{HTML}{CD7F32}

\usepackage{amsmath}
\usepackage{graphicx}
\usepackage[inline]{enumitem}
\usepackage{algorithm}
\usepackage[capitalize,noabbrev]{cleveref}
\usepackage{pifont}
\definecolor{darkgreen}{RGB}{41,166,41}

\usepackage{dsfont}
\newcommand{\indic}[1]{\mathds{1}\{#1\}}
\usepackage{algpseudocode}
\usepackage{subcaption}  
\usepackage{wrapfig}
\usepackage{titletoc}
\usepackage{hyperref}
\usepackage{siunitx}
\usepackage{placeins}

\title{TabPrep: Closing the Feature Engineering Gap in Tabular Benchmarks}

\author{%
  \textbf{Andrej Tschalzev}\textsuperscript{1$\dagger$},
  \textbf{Nick Erickson}\textsuperscript{2$\dagger$},
  \textbf{Yuyang Wang}\textsuperscript{3},
  \textbf{Huzefa Rangwala}\textsuperscript{4$\dagger$}, \\
  \textbf{Stefan Lüdtke}\textsuperscript{\textbf{5}},
  \textbf{Heiner Stuckenschmidt}\textsuperscript{\textbf{1}},
  \textbf{Christian Bartelt}\textsuperscript{\textbf{6}} \\
  \textsuperscript{1}University of Mannheim \quad
  \textsuperscript{2}Prior Labs \quad
  \textsuperscript{3}AWS \quad 
  \textsuperscript{4}Siemens \quad
  \textsuperscript{5}University of Rostock \\
  \textsuperscript{6}Technical University of Clausthal
  \textsuperscript{$\dagger$}Part of the work done at AWS
}

\begin{document}

\maketitle

\begin{abstract}
Progress in tabular machine learning has largely focused on increasingly sophisticated model architectures. At the same time, feature engineering remains a critical yet underexplored component of real-world modeling pipelines that is entirely absent from modern benchmarks, which creates an unquantified evaluation gap. In this work, we introduce \textbf{TabPrep}, a lightweight preprocessing pipeline composed of feature generators that are carefully designed to target three specific structural data patterns. We show that many widely used model classes exhibit predictable blind spots to these patterns and that systematic feature engineering alone can establish new peak performance. Across the TabArena benchmark, integrating TabPrep into model training and tuning consistently improves performance for tree-based, neural, linear, and foundation models, often surpassing gains achieved by model-centric innovations alone. TabPrep outperforms previous automated feature engineering approaches in performance, efficiency, and applicability across datasets, enabling integration into large-scale benchmarks. By releasing TabPrep (see \href{https://github.com/atschalz/tabprep}{https://github.com/atschalz/tabprep}), we enable researchers to integrate feature engineering into their benchmarking setup, filling a longstanding gap in tabular evaluations. 
\end{abstract}


\begin{figure}[!h]
    \centering
    \includegraphics[width=\columnwidth]{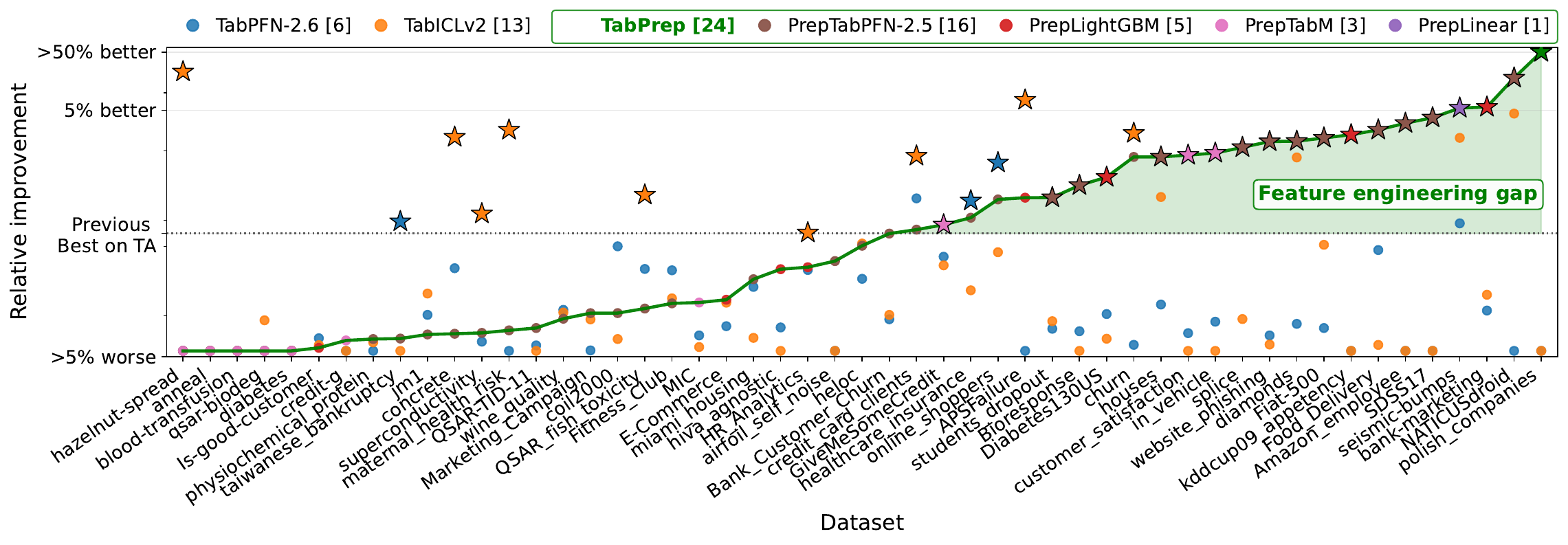}
    \caption{\textbf{Illustration of how TabPrep exposes the peak performance gap missed by models.} We zoom into the top of the leaderboard per dataset on the TabArena benchmark and compare recent model-centric improvements (TabPFN-2.6 \cite{priorlabs_tabpfn_2_6}, TabICLv2 \cite{qu2026tabiclv2}) along with our data-centric improvements from augmenting four models with TabPrep. Improvement is relative to the previous best out of 25 TabArena models tuned for 200 configurations, including the non-TabPrep-augmented versions of the  models. The y-axis is on a symmetric log scale. Datasets are ordered by the highest gain from TabPrep. Improvement is averaged over folds. Numbers in the legend correspond to the number of datasets where a model improves peak performance. Stars mark new peak performance.
    }
    \label{fig:performance_across_datasets}
\end{figure}

\section{Introduction} \label{sec:intro}

Recent years have seen rapid progress in tabular machine learning, with a growing number of models, architectures, and training paradigms proposed to improve predictive performance \cite{hollmann2022tabpfn,marton2023grande,holzmuller2024better,gorishniy2024tabm,tschalzev2024enabling,ma2024tabdpt,ye2024modern,qu2025tabicl,grinsztajn2025tabpfn,zhang2025limix,qu2026tabiclv2,zhang2025mitra,spinaci2025contexttab,joseph2021pytorch,thielmann2024efficiency,gardner2024large}. Much of this progress has focused on modeling choices, while preprocessing is often treated as a secondary implementation detail that is typically not benchmarked explicitly \cite{tschalzev2024data, tschalzev2025unreflected}. This contrasts with extensive empirical evidence from competitive machine learning \cite{tunguz2023kaggle,tschalzev2024data} and industry \cite{rubachev2024tabred} showing that feature engineering (FE) can have a decisive impact on performance. \\
The reasons for this imbalance are multi-faceted: 
\begin{enumerate*}[label=(\textbf{\arabic*})]
    \item Successful FE pipelines are highly dataset-specific and typically driven by domain knowledge and manual crafting, which is infeasible in academic benchmarking. 
    \item Although automated FE methods have been proposed \cite{zhang2023openfe,horn2019autofeat}, existing methods rely on exhaustive search procedures and are limited in usability \cite{schafer2025usable}. As we will demonstrate, integrating them into modern benchmarking pipelines is difficult due to frequent out-of-memory failures and extreme training time increases. 
    \item There is no common understanding of which feature engineering methods remain relevant in the modern model landscape with novel tabular deep learning and foundation models \cite{gorishniy2024tabm, hollmann-nature25a}.
    \item Even though many historically useful operations are known, benchmark-ready feature generators targeting specific operation families are missing. \\
\end{enumerate*}
Surprisingly, no simple, ready-to-use feature engineering baseline has been proposed for tabular benchmarking so far. As a result, recent tabular benchmarks omit feature engineering baselines, leaving a gap in the evaluation of peak performance \cite{erickson2025tabarena, grinsztajn2022tree, mcelfresh2023neural,bischl2025openml}.
We call this discrepancy the feature engineering gap: the difference between the performance frontier measured by model-only benchmarking and the performance frontier attainable when feature engineering is included in the evaluation protocol.
We address this gap with the following contributions:

\begin{enumerate}[topsep=2pt,itemsep=2pt,parsep=0pt]
    \item We identify three recurring structural patterns in tabular datasets that are not consistently captured by current model classes, thereby offering guidance on which forms of feature engineering remain relevant in the modern tabular model landscape.
    \item We propose automated feature generators targeted toward these patterns and combine them into \textbf{TabPrep}, a lightweight preprocessing pipeline that outperforms existing automated feature engineering approaches in performance, efficiency, and applicability, enabling large-scale benchmarking of improvements through feature engineering for the first time.
    \item We integrate TabPrep into the TabArena evaluation protocol and show that feature engineering can establish new peak performance at the benchmark level (\autoref{fig:elo_full_medium}) and dataset level (\autoref{fig:performance_across_datasets}), making the feature engineering gap in current tabular evaluations explicit.
\end{enumerate}
By making the feature engineering gap explicit, TabPrep shows that purely model-centric comparisons can underestimate peak performance and that current models show systematic blind spots. Thereby, TabPrep can guide the development of future tabular foundation models and offers a baseline for measuring whether future models close the feature engineering gap.

\section{Related Work} \label{sec:rel_work}
Although a large variety of \textbf{tabular benchmarking studies} has been conducted in the past \cite{bischl2017openml,bischl-arxiv17a,gorishniy2021revisiting,purucker2023assembled,gardner2023benchmarking,grinsztajn2022tree,mcelfresh2023neural,salinas2024tabrepo,tschalzev2024data,rubachev2024tabred,kohli2024towards,shmuel2024comprehensive,zabergja2024tabular,erickson2025tabarena}, none ever included general feature engineering baselines. As a result, current benchmarking evaluations, including the industry-standard benchmark TabArena \cite{erickson2025tabarena}, leave a gap in evaluating peak performance. We offer a methodology and empirical evidence to make this gap explicit in tabular benchmarks.
The most related study by \citet{tschalzev2024data} analyzed the impact of feature engineering on performance in Kaggle competitions, but only highlighted dataset-specific solutions, whereas we propose a generalizable feature engineering methodology. \\
Existing machine learning libraries such as scikit-learn \cite{scikit-learn}, \href{https://github.com/skrub-data/skrub}{skrub}, feature-engine \cite{galli2021feature}, and category\_encoders \cite{mcginnis2018category} provide large \textbf{collections of preprocessing techniques} for various purposes.
In contrast, we focus exclusively on feature engineering (FE) to improve predictive performance, rather than preprocessing for data compatibility or cleaning \cite{verdonck2024special}.
Recent work suggests that longstanding preprocessing techniques such as rebalancing for class imbalance or missing value imputation are already sufficiently well handled by modern models \cite{liu2025climb,morvan2024imputation}.
Instead, we focus on identifying feature engineering techniques that remain impactful in the current tabular model landscape.
Although some operations that we build on are well known for feature engineering, to our knowledge, only target encoding is implemented as a dedicated generator in the referenced libraries. Moreover, our proposed composition  of techniques is novel and has not been operationalized for benchmarking before.
\\
Most existing \textbf{automated feature engineering frameworks} rely on the expand-and-reduce paradigm \cite{zhang2023openfe} of large-scale feature generation followed by (supervised) feature selection \cite{horn2019autofeat,zhang2023openfe,yuanfei2019autocross,kanter2015deep,lam2021automated,fan2010generalized,shi2020safe,nargesian2017learning}. In contrast, we propose a paradigm shift toward budgeted pattern-informed generation and show that despite its simplicity our approach outperforms supervised expand-and-reduce selection, avoiding its high computational costs and risks of failure. 
Some other approaches are specialized to certain domains \cite{bonidia2022bioautoml, lin2022adafs}, while we provide a generally applicable approach. 
In addition, a large body of work focuses on feature selection \cite{guyon2003introduction,bakheet2023hybrid,yu2003feature,kursa2010feature,balin2019concrete,yamada2020feature,moslemi2023tutorial,cherepanova2023performance,barbieri2024analysis}. In contrast, we focus on principled feature generation.
Although LLM-based solutions for automated feature engineering have been proposed \cite{hollmann2023large,abhyankar2025llm}, they do not always suggest useful features \cite{kuken2024large} and raise concerns about data leakage, hindering a fair evaluation \cite{bordtelephants}. TabPrep relies on a simple, generalizable pipeline and therefore can be used for benchmarking without risking leakage. 
Finally, \citet{schafer2025usable} found that only a small fraction of existing frameworks can be used in practice, calling for reliable autoFE approaches.
TabPrep fills this gap as a simple baseline that is easy to use and can augment any model.

\section{TabPrep: Pattern-Informed Feature Generators for Structural Patterns} \label{sec:latent_structure}
Previous approaches for automated feature engineering (autoFE) predominantly follow an expand-and-reduce paradigm of exhaustive model-based selection over concrete operations \cite{zhang2023openfe,horn2019autofeat}. 
To enable large-scale benchmarking of the impact of feature engineering, we propose a paradigm shift abandoning expand-and-reduce in favor of \textbf{budgeted pattern-informed generation}\footnote{See Appendix \ref{appendix:expand_and_reduce}, where we distinguish the two paradigms with a focus on practicability for large-scale benchmarking.}.
Our core premise is that feature engineering is not exclusively helpful because of exact operations (e.g. multiplication of two specific features), but that datasets have underlying structural properties such that whole families of operations (e.g. arithmetic interactions) are helpful. 
Therefore, instead of exhaustive model-based selection over concrete operations, we propose to reduce the search problem to a \textbf{search over generators targeting specific patterns} that reoccur in tabular data. 
To motivate this paradigm, we start by defining the role of feature engineering in tabular learning.

\subsection{The Role of Feature Engineering: Externalize Structural Patterns to Reduce Model Bias} \label{ssec:role_fe}

\begin{figure}[t]
    \centering
    \includegraphics[width=\columnwidth]{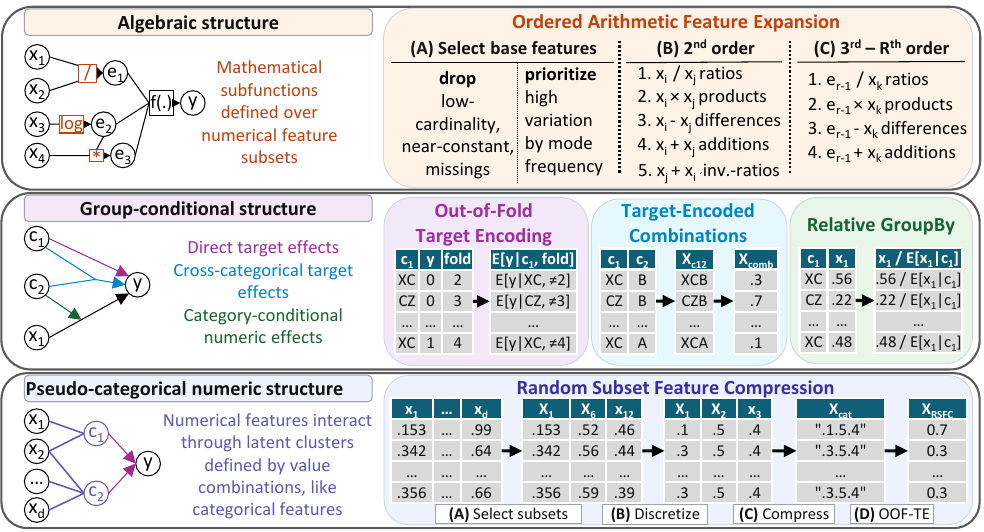}
    \caption{\textbf{Overview of the three structural patterns targeted by TabPrep and the feature generators addressing them}. The generator illustration focuses on the types of the generated features. The underlying procedures for feature filtering and prioritization under limited budgets are in Appendix \ref{appendix:fe-method-details}.}
    \label{fig:system}
\end{figure}

Let $\mathcal{D} = (\mathbf{X}_{\mathrm{num}}, \mathbf{X}_{\mathrm{cat}}, \mathbf{y})$ denote a dataset, where
$\mathbf{X}_{\mathrm{num}} \in \mathbb{R}^{n \times d_{\mathrm{num}}}$ contains the numerical features, and
$
\mathbf{X}_{\mathrm{cat}} = [\mathbf{c}_1,\dots,\mathbf{c}_{d_{\mathrm{cat}}}]
$
contains the categorical features, with $\mathbf{c}_k \in \mathcal{C}_k^n$ for $k = 1,\dots,d_{\mathrm{cat}}$, where each $\mathcal{C}_k$ is a
finite set and $|\mathcal{C}_k|$ may differ between $k$.
The target vector $\mathbf{y} \in \mathcal{Y}^n$ corresponds to either a regression or classification task. \\
Although many models commonly used for tabular data are universal function approximators in principle \cite{hornik1989multilayer}, learning a well-generalizing function $y = f^\star(\mathbf{x})$ from an infinite hypothesis space $\mathcal{F}$ with limited data is challenging and prone to overfitting.
Different model classes, such as neural networks (NNs) or gradient-boosted decision trees (GBDTs), have distinct inductive biases, representing assumptions of the learning algorithm that favor particular structural properties of functions \cite{mitchell80, baxter2000model}, effectively imposing a prior on $\mathcal{F}$ to favor a model-dependent function subset $\mathcal{F}^m$. \\
For tabular data, the input space is naturally factorized by features, allowing the target
function to be expressed as a composition of feature transformations:
\[
f^\star(\mathbf{x})
=
g^\star\!\left(
f^\star_1(\mathbf{x}_{S_1}),
\dots,
f^\star_L(\mathbf{x}_{S_L})
\right),
\]
where $S_1,\dots,S_L$ denote (possibly overlapping) groups of features,
$f^\star_\ell : \mathcal{X}_{S_\ell} \to \mathbb{R}^{d_\ell}$ maps each group to a latent
representation, and
$g^\star : \mathbb{R}^{m_1} \times \cdots \times \mathbb{R}^{m_L} \to \mathcal{Y}$.
We refer to a component $f^{\star}_{\ell}$ as a structural data pattern.
In this paper, we view feature engineering as a methodology to explicitly represent structural patterns to compensate for systematic limitations from model inductive biases.
If a component $f^\star_l$ can be approximated explicitly via a feature transformation $\Phi$, s.t.  $\Phi(\mathbf{x}_{S_\ell}) = \mathbf{x}_{\Phi} \approx f^\star_\ell(\mathbf{x}_{S_\ell})$, the model no longer needs to implicitly represent this component.
To reduce generalization error, a feature transformation $\Phi$ needs to
(i) address a subfunction $f^\star_\ell$ that is difficult to learn implicitly under limited data or a misaligned inductive bias, and
(ii) provide a sufficiently accurate approximation of $f^\star_l$. \\
As discussed in Section~\ref{sec:intro}, current benchmarks lack dataset-agnostic feature engineering baselines that target patterns modern models still miss. We therefore identify such structural patterns and introduce pattern-driven generators for them\footnote{The identified patterns were selected based on evidence from prior work; see Appendix~\ref{appendix:ind_bias}.}. 
For an overview of the identified patterns and proposed generators, see \autoref{fig:system}.

\subsection{Ordered Arithmetic Feature Expansion for Algebraic Structure in Numeric Features}
\textbf{Data Structure.} \hspace{1.5mm} Many tabular datasets exhibit structured algebraic relationships between numerical features defined by fixed compositions of mathematical operations.
This structure naturally arises in domains governed by physical laws, financial rules, or engineered systems, where numerical features represent measurements and algebraic expressions provide a meaningful description of the underlying relationships.
This results in subfunctions $f^\star_\ell$ such as arithmetic combinations or univariate transformations of numerical features, e.g.,
\[
f^\star_\ell(\mathbf{x}_{S_\ell}) = \mathbf{x}_i + \mathbf{x}_j,\qquad
f^\star_\ell(\mathbf{x}_{S_\ell}) = \frac{\mathbf{x}_i}{\mathbf{x}_j},\qquad
f^\star_\ell(\mathbf{x}_{S_\ell}) = \log(\mathbf{x}_i),\qquad
f^\star_\ell(\mathbf{x}_{S_\ell}) = \mathbf{x}_i^{\,2},
\]
where $x_i$ and $x_j$ denote individual numerical features. 
Real-world datasets rarely follow a single global function. Instead, structured relationships often hold only between subsets of features, while the target and other features may depend on additional factors. 
Although the possible feature transformations $\Phi$ are known mathematical operations, finding the right operations on high-dimensional datasets is challenging due to combinatorial explosion.

\textbf{Inductive Bias Alignment.} \hspace{1.5mm} Tree-based models rely on axis-aligned partitions, and therefore lack an inductive bias for approximating smooth algebraic functions. However, even neural networks have been shown to struggle at learning exact arithmetic operations \cite{trask2018neural, madsen2020neural, nogueira2021investigating}. In Appendix \ref{appendix:ind_bias_arithmetic}, we study model performance on datasets with known arithmetic interaction patterns, and show that \textbf{(1)} tree-based models severely underperform on such datasets, and \textbf{(2)} even NNs and tabular foundation models (TFMs) improve from explicit feature engineering of such patterns. 

\textbf{Ordered Arithmetic Feature Expansion.} \hspace{1.5mm} To target improvements on datasets with algebraic structure, we propose Ordered Arithmetic Feature Expansion (OAFE)\footnote{We provide a detailed generator description in Appendix \ref{appendix:fe-method-details} and focus on the major design decisions in the main paper}.
The algorithm maps numerical base features via arithmetic feature transformations under a budget of maximum $M$ generated features and a maximum order $r$. To improve the odds of generating informative features, we employ a set of heuristics:
\begin{enumerate*}[label=(\textbf{\arabic*})]
    \item Apply lightweight filtering to remove low-cardinality, near-constant, or predominantly missing features.
    \item  Limit the set of base feature candidates to the ones with high variation measured by mode frequency.
    \item Impose a fixed order over operators, starting with division $(/, \times, -, +)$, reflecting differing levels of learnability: Addition and subtraction can be approximated with linear models, whereas multiplication and division introduce increased nonlinearity. Division additionally requires the implicit approximation of a reciprocal and exhibits singular behavior as the denominator approaches zero.
    \item Do not include univariate operations as they severely increase the search space while being less challenging for modern models (e.g., tree-based models are almost unaffected by monotonic transformations, like log-transform).
\end{enumerate*} \\
OAFE generates features without the guarantee that they result from true target subfunctions $f^\star_\ell$.
Instead, $\mathbf{X}_{\Phi}$ contains candidates for low-order patterns, reducing the learning problem to feature selection, rather than requiring the model to discover such structure from scratch. This simplification allows for feasible integration into benchmarking practice.
\subsection{The CatPrep Pipeline for Group-Conditional Target Structure} \label{ssec:catprep}
\textbf{Data Structure.} \hspace{1.5mm} Many tabular datasets contain categorical features, which do not allow a meaningful notion of smoothness or similarity in the raw input space. 
Instead, we argue that they indicate potential structural heterogeneity, where the target or other features may vary systematically across category-defined subsets of data points.
Drawing on concepts from mixed-effects modeling \cite{zuur2009mixed,bates2010lme4,galecki2012linear}, we focus on three forms of structural patterns where categorical features induce distinct effects on the target:
\begin{center}
\begin{minipage}[t]{0.31\linewidth}
\centering
\textbf{(1) Direct Categorical}\\[0.25em]
$
y 
= f^\star_{\ell}(\mathbf{c}_{k})
\!\left(h^\star(\mathbf{x}_{\mathrm{num}})\right)
$\\[0.35em]
where $h^\star$ captures category-invariant numerical effects, and $f^\star_{\ell}(\mathbf{c}_{k})$ denotes a category-conditional transformation of the effect, e.g. offset, or scaling.
\end{minipage}
\hfill
\begin{minipage}[t]{0.31\linewidth}
\centering
\textbf{(2) Cross-categorical}\\[0.25em]
$
y 
= f^\star_{\ell}(c_{S_\ell})
\!\left(h^\star(\mathbf{x}_{\mathrm{num}})\right)
$\\[0.35em]
where $f^\star_{\ell}$ is defined over the Cartesian product on a subset of categorical features $S_\ell = (\mathbf{c}_{1}, \dots, \mathbf{c}_{k})$, capturing joint target effects. 
\end{minipage}
\hfill
\begin{minipage}[t]{0.31\linewidth}
\centering
\textbf{(3) Cat-conditional Numeric}\\[0.25em]
$
y 
= h^\star\!\left(
\mathbf{x}_{\mathrm{num}},
f^\star_{\ell}(\mathbf{x}_{S_L}\mid \mathbf{c}_k)
\right)
$\\[0.35em]
where $f^\star_{\ell}$ transforms a subset of numeric features conditionally on $\mathbf{c}_k$ prior to the shared mapping $h^\star$, e.g. as offsets, rescaling, or other functions. 
\end{minipage}
\end{center}

For notational simplicity, we illustrate a single categorical feature for each effect, though in practice multiple categorical features may affect the target.
Furthermore, high-cardinality categorical features are especially challenging \cite{pargent2022regularized}: 
\textbf{(1)} If group-conditional effects are strong and stable, they can be hard to learn if only a limited number of observations per group were observed.
\textbf{(2)} If a high-cardinality categorical feature is irrelevant, this can be hard to learn, since traditional encoding schemes like one-hot encoding allocate a high number of parameters.

\textbf{Inductive Bias Alignment.} \hspace{1.5mm}
All tabular models use numeric encodings, whereas group-conditional patterns occur in a symbolic space. Hence, all models have an inductive bias defined by the encoding strategy. In Appendix \ref{appendix:ind_bias_categorical}, we analyze the behavior of models from different families w.r.t. each of the three types of group-conditional effects, and find that 
\textbf{(1)} TabPFN-2.5 \cite{grinsztajn2025tabpfn} fails to learn all three patterns reliably if the categorical features are high-cardinality, likely due to such features being absent from known tabular foundation model priors;
\textbf{(2)} Similarly, LightGBM's \cite{ke2017lightgbm} categorical feature handling does not capture any of the three patterns;
\textbf{(3)} TabM \cite{gorishniy2024tabm} performs well in the presence of impactful high-cardinality categorical features, but can overfit if these features are less important;
\textbf{(4)} Only CatBoost \cite{prokhorenkova2018catboost} can learn cross-categorical target modulation effects;
\textbf{(5)} No model learns group-conditional shifts in numerical feature effects without explicit feature engineering.


\textbf{CatPrep: Three Generators for Categorical Feature Effects.} \hspace{1.5mm} 
No single encoding strategy for categorical data is flexible enough to explicitly cover all three kinds of categorical feature effects.
Therefore, we propose a generator pipeline (\textbf{CatPrep}) of three complementary techniques, each targeted towards one type of group-conditional effects.

\textbf{Out-of-Fold Target Encoding.} \hspace{1.5mm}
Target encoding \cite{micci2001preprocessing} is already implemented in standard ML toolkits (scikit-learn, category\_encoders), and its leave-one-out variant is used in some deep learning models \cite{popov-arxiv19a, marton2023grande}. 
However, both target and leave-one-out-encodings have been shown to lead to overfitting, while cross-validated variations are often superior \cite{pargent2022regularized}.
Therefore, we use Out-of-Fold Target Encoding (OOF-TE), which replaces each categorical value with an estimate of its expected target value computed on the training folds and applied to the corresponding held-out fold rows in a cross-validation procedure \cite{pargent2022regularized}. 
For multi-class targets, we perform a one-vs-rest mapping for each class.
Note that this approach captures \emph{unadjusted} group-level associations and not true direct category effects. 

\textbf{Target-Encoded Combinations.} \hspace{1.5mm} Some feature engineering libraries include Cartesian products of categorical features \cite{zhang2023openfe}. However, combinations of categorical features result in high-cardinality categorical features, increasing model training times and the risk of overfitting.
Consequently, we do not include combinations of categorical features directly, but always encode them using OOF-TE, such that the resulting feature representation is an unadjusted, regularized estimation of the joint target modulation effects.
As for arithmetic interactions, we implement budgeted feature generation and generate up to $M$ random target encoded combinations up to order $r$. 



\textbf{Relative GroupBy Interactions.}  \hspace{1.5mm}
To cover interactions between numerical and categorical features, feature engineering libraries such as OpenFE \cite{zhang2023openfe} include GroupBy operations, which compute summary statistics (i.e., mean, max, std, ...) of a numerical feature per category of a categorical feature.
This can lead to a large candidate space of highly similar features with samples of the same category sharing the same value, thereby re-expressing information about differences \emph{between} categories.
We argue that to uncover group-conditional effects on numeric features, \emph{intra-group differences} are more expressive information.
A subfunction $f^\star_{\ell}\!\left(\mathbf{x}_{j} \mid \mathbf{c}_k\right)$ that is challenging to learn, would not only collapse numerical features into single category-specific constants, but produces values that vary within each category.
Therefore, we solely rely on relative GroupBy transformations, like: 
$
\mathbf{x}_{\mathrm{g_1}}
=
\mathbf{x}_j - \mathbb{E}\!\left[\, \mathbf{x_j} \mid \mathbf{c}_k \,\right],
$ and
$
\mathbf{x}_{\mathrm{g_2}}
=
\frac{\mathbf{x_j}}{\mathbb{E}\!\left[\, \mathbf{x_j} \mid \mathbf{c}_k \,\right]},
$
where $\mathbf{x}_j$ denotes a numerical feature, $\mathbf{c}_k$ denotes
a categorical feature, and $\mathbb{E}$ is the expected value estimated from the training data.
Therefore, numerical features are reinterpreted relative to group-specific reference levels. 
In addition, we use the normalized rank of a numerical feature within a category, as in related work \cite{zhang2023openfe}.
To operationalize GroupBy feature engineering, we again rely on ordered budgeted feature generation and sample interactions up to a user-defined amount. In addition, we implement heuristics to maximize the odds of generating informative features (see Appendix \ref{sec:groupby-preprocessor}). 

\subsection{Random Subset Feature Compression for Pseudo-Categorical Interaction Structure}
\textbf{Data Structure.} \hspace{1.5mm} Input features that are categorical are typically explicitly specified and treated differently by models. However, numerical features may relate to the target like a categorical feature, that is, by value instead of range. 
For example, ordinal, date-extracted, or spatial features are often represented numerically, even though proximity in their numeric representation may not imply similarity in target behavior, and individual values may induce distinct, group-specific effects.
This is especially challenging if these patterns only show as interactions with other features.
We formalize this data structure by assuming that a subset $S_\ell$ of one or more numerical features depends on a latent grouping variable.
Specifically, for each sample $i = 1,\dots,n$, we assume
\[
\mathbf{x}_{i,S_\ell}
=
h_{S_\ell}\!\left(g_{i,k}\right)
+
\xi_{k},
\]
where $g_{i,k} \in \mathcal{G}_k$ is a latent grouping variable that takes values in a finite
(possibly large) set $\mathcal{G}_k$, $h_{S_\ell} : \mathcal{G}_k \to \mathbb{R}^{S_\ell}$ is an arbitrary
numeric encoding of group identities, which may result in multiple numeric features, and $\xi_{i,k}$ denotes noise introduced in the
process of representing discrete group labels as numerical values. As a consequence,
proximity in the observed numerical features $x_{i,{S_\ell}}$ does not imply semantic similarity
of the underlying latent groups.
In the extreme case, $g_k$ may correspond to a nominal feature, and $h_{S_\ell}$ to an arbitrary function producing multiple new floating-point features. In a less extreme case, $g_k$ may be an ordered categorical feature with additive noise.

\textbf{Inductive Bias Alignment.}  \hspace{1.5mm} Categorical features, if treated as numeric, induce a high-frequency dependence of $y$, which is a known challenge for neural networks, due to an inductive bias towards low-frequency functions \cite{grinsztajn2022tree}. In the worst case, pseudo-categorical numeric features can be as suboptimal as ordinally-encoded categorical features, which with sufficiently high cardinality is a challenge for any model. But even for low-cardinality features, pseudo-categorical interactions stretching over multiple features are hard to learn. In Appendix \ref{appendix:ind_bias_pseudocat}, we show that all studied models, including LightGBM, CatBoost, TabM, and TabPFN-2.5, can benefit from feature engineering targeting pseudo-categorical numeric interactions. 

\textbf{Random Subset Feature Compression.}  \hspace{1.5mm}
Naively converting numerical features into categorical ones carries a high risk of overfitting. 
Therefore, we \textbf{(1)} never consider single features in isolation, but whether they interact with other features in the way categorical features do, and \textbf{(2)} we never include numerical features as categorical directly, but always with OOF-TE.
To enable a low-risk handling of pseudo-categorical feature interactions, we propose \emph{Random Subset Feature Compression (RSFC)}. The method proceeds as follows:
\begin{enumerate*}[label=(\textbf{\arabic*})]
    \item Discretize numerical features by rounding to a predefined resolution.
    \item Randomly sample subsets of features from the set of discretized numerical and true categorical features within a specified range of
    interaction orders $[r_{\mathrm{min}},r_{\mathrm{max}}]$.
    \item Combine each sampled subset into a single categorical feature. 
    \item Apply out-of-fold target encoding (OOF-TE) to the resulting combined features.
    \item Stop once $M$ total features have been generated.
\end{enumerate*}
The resulting features serve as compressed summaries of potentially complex value-wise interactions.
In the presence of many low-cardinality features, we use higher-order random feature combinations and let the first generated feature be a combination of all features, covering sample regions with high density in the training data.
Due to this design, RSFC helps not only with pseudo-categorical features, but also with higher-order interactions in low-cardinality features in general.

\subsection{TabPrep: A Tabular Preprocessing Pipeline for Benchmarking Feature Engineering}
To enable large-scale benchmarking, we combine the generators into a preprocessing pipeline, termed \textbf{TabPrep}. In summary, efficiency compared to related autoFE frameworks is achieved as follows: 
\begin{enumerate*}[label=(\textbf{\arabic*})]
\item We reduce the search space to a few generators for specific structural patterns instead of single operations.
\item We rely on predefined feature budgets per generator to avoid the extreme computational costs of feature selection over a large set of prior-generated features. 
\item All generators use heuristics to produce informative features by default s.t. expensive hyperparameter optimization of the preprocessors is not necessary. The exact hyperparameters for the TabPrep default pipeline are in \autoref{appendix:fe-method-details}.
\item We account for the fact that feature engineering techniques cannot be universally beneficial and treat feature expansion as a hyperparameter in the modeling pipeline. However, to demonstrate the tuning simplicity of TabPrep, we solely tune whether specific preprocessors are used along with the random selection of features.
\end{enumerate*}
Note that whenever we use target information in any way, we nest it inside the inner folds solely using the training data to prevent any leakage.

\section{Experimental Results} \label{sec:tabarena_results}

\textbf{Experimental Setup.} \hspace{1.5mm} For our evaluation, we use TabArena \cite{erickson2025tabarena}, a recent large-scale benchmark for evaluating supervised models on tabular data. 
We adopt the TabArena evaluation protocol without modification, where each model is evaluated in three setups: (1) default hyperparameters, (2) hyperparameter tuning, and (3) an ensemble constructed over all configurations explored during tuning. We evaluate TabPrep in combination with four models from different families: (1) a linear model, (2) LightGBM \cite{ke2017lightgbm}, (3) TabM \cite{gorishniy2024tabm}, and (4) TabPFN-2.5 \cite{grinsztajn2025tabpfn}. \\
To evaluate the potential of TabPrep to augment the models, we use TabPrep as follows: convert existing categorical features using OOF-TE and add up to the following numbers of features: 
2000 arithmetic features generated with OAFE, 
100 target encoded categorical combination features, 
500 relative GroupBy features, and 
50 random-subset-compressed features. 
To demonstrate that TabPrep is indeed a simple baseline that does not require extensive tuning in order to be useful, we only tune the number of generated features and their random sampling.
To ensure evaluation under the same conditions and adhere to the feature size limits of TabPFN-2.5, we limit the number of features passed to the model to 2000. \\
For models augmented with TabPrep, we follow TabArena and tune 200 configurations using the original TabArena model search space and add TabPrep to 25\% of the configurations.
For LightGBM and the linear model, we use 8 vCPUs, and for TabPFN-2.5 and TabM we use NVIDIA H100 GPUs.
Although we benchmark our pipeline for four models, we compare to every model currently in the TabArena benchmark, which at the time of writing are 27 models (including TabPFN-2.6 \cite{priorlabs_tabpfn_2_6} and TabICLv2 \cite{qu2026tabiclv2}), where most models are evaluated with 200 configurations for hyperparameter optimization.
The development of TabPrep was finished before the addition of TabPFN-2.6 and TabICLv2 to TabArena, which allows us to compare the data-centric improvements through TabPrep directly to the two most recent model-centric improvements.
Following TabArena, we use the ELO metric as the main aggregation metric.
Full details on the experimental design are in \autoref{appendix:experiments}.



\subsection{Peak Performance Analysis on Benchmark and Dataset Level}
\begin{figure}[t]
    \centering
    \begin{subfigure}[t]{0.5\columnwidth}
        \centering
        \includegraphics[width=\linewidth]{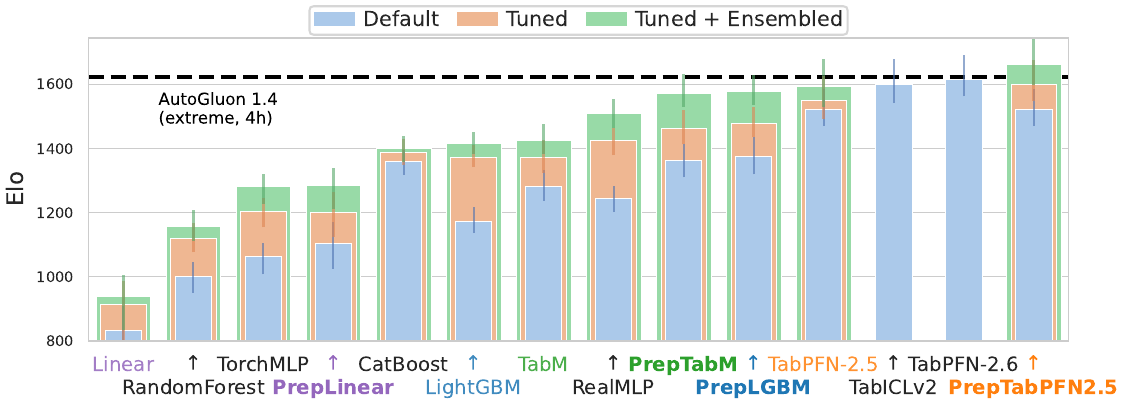}
        \label{fig:tabarena_all_models}
    \end{subfigure}%
    \begin{subfigure}[t]{0.5\columnwidth}
        \centering
        \includegraphics[width=\linewidth]{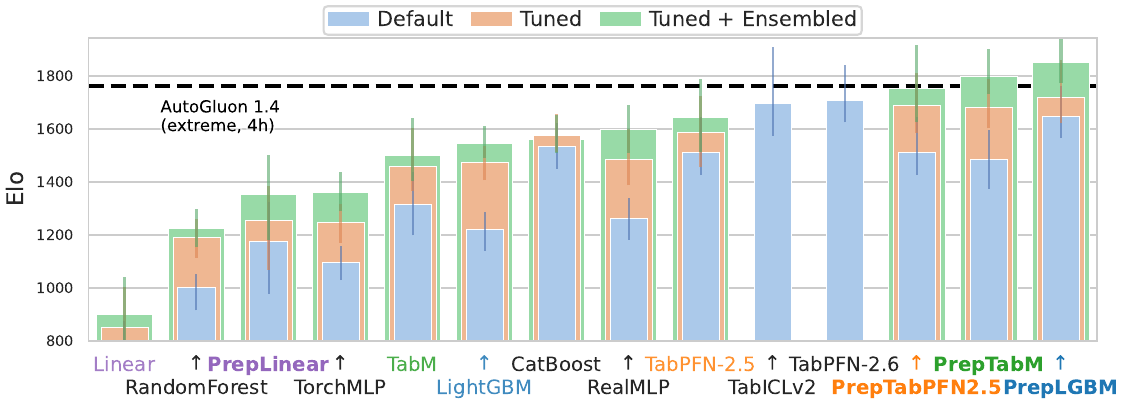}
        \label{fig:medium}
    \end{subfigure}

    \caption{\textbf{Performance of models augmented with TabPrep on the TabArena benchmark.} Elo scores (higher is better) are averaged over datasets. Models labeled with \textit{Prep} use TabPrep for data preprocessing, while all other results are taken from the official TabArena benchmark \cite{erickson2025tabarena}. \textbf{Left:} Results for the full 51 TabArena datasets; \textbf{Right:} Results for the 15 largest datasets in the benchmark, between 10,000 and 100,000 training samples. }
    \label{fig:elo_full_medium}
\end{figure}

\textbf{TabPrep Improves Benchmark Performance Across Models.} \hspace{1.5mm}
\autoref{fig:elo_full_medium} shows that TabPrep improves all augmented models on average across datasets: 
\begin{enumerate*}[label=(\textbf{\arabic*})]
    \item When augmented with TabPrep, even the linear model is competitive with minimally preprocessed versions of more advanced models, such as random forests or MLPs. 
    \item LightGBM and TabM, which previously lagged behind foundation models by a substantial margin, catch up close to the performance of modern foundation models when augmented with TabPrep.
    \item TabPFN-2.5 augmented with TabPrep outperforms all other competitors including the more recent models TabPFN-2.6 and TabICLv2.
\end{enumerate*}
We use the ELO metric because it is the main TabArena metric, however, TabPrep's improvements are even more substantial when looking at other aggregation metrics, especially those rewarding new peak performance improvements, such as harmonic mean rank (see Appendix \ref{app:alternative_metrics}).



\textbf{TabPrep Improves Peak Performance on Dataset-Level.} \hspace{1.5mm}
Crucially, TabPrep not only improves average performance across datasets, but also achieves substantial dataset-level gains establishing new peak performances.
In \autoref{fig:performance_across_datasets}, TabPrep-augmented models outperform the previously best-performing of all TabArena models on 24/51 TabArena datasets (19 with TabPFN-2.6 and TabICLv2).
Moreover, it can be seen that different models achieve peak performance on different datasets, indicating that mitigating the bias of different models through feature engineering can unlock distinct potential for improvements across models.
With the right feature engineering, different models (sometimes even a linear model) can outperform all other TabArena models on several datasets.
Finally, the performance gains observed across a large number of datasets confirm that the data structures identified in \autoref{sec:latent_structure} generalize to a broad range of real-world datasets.
In addition, \autoref{fig:isolate_comparison} (and \autoref{app:loo_contrib}) confirm that our implemented generators are complementary, and that to minimize the risk of performance degradation on individual datasets, hyperparameter optimization is required.

\textbf{Analyzing the Feature Engineering Gap.} \hspace{1.5mm}
Figure~\ref{fig:performance_across_datasets} visualizes recent model-centric advances along with data-centric gains from augmenting models with TabPrep. Both directions, new models and TabPrep improve the benchmark frontier on several datasets. In comparison, TabPrep improves on more datasets and often by larger margins. This shows that current tabular benchmarks leave considerable performance gains unaccounted for when feature engineering is omitted. The largest TabPrep gains often occur where even recent model additions remain below the TabPrep-augmented frontier, while some datasets with little or no TabPrep benefit are improved by newer models, suggesting that data-centric and model-centric progress address partially distinct sources of performance. On datasets where new models and TabPrep both help, TabPrep typically yields larger relative gains, indicating that explicitly exposing structural patterns can be more effective than changing the model architecture alone. Overall, feature engineering captures complementary modes of insufficient model biases that remain invisible in purely model-centric evaluations. TabPrep closes this gap for the three studied structural patterns.


\subsection{Comparison to AutoFE Libraries}
\begin{figure}[t]
    \centering

    \begin{subfigure}[t]{0.57\textwidth}
        \centering
        \includegraphics[width=\linewidth]{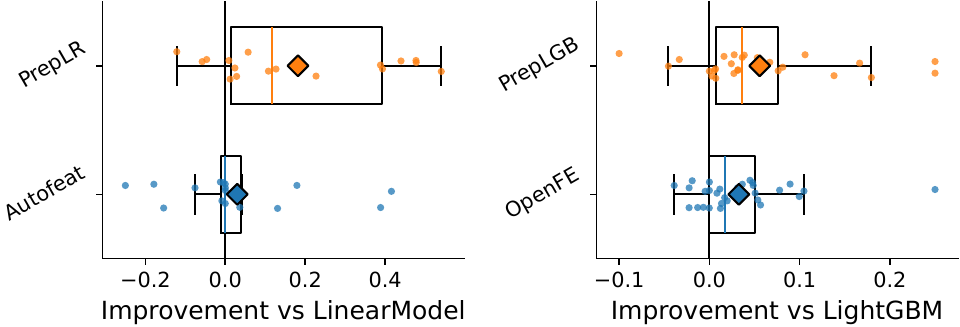}
        \caption{\textbf{Comparison of TabPrep vs. autoFE libraries.}
        }
        \label{fig:autofe_main}
    \end{subfigure}
    \hfill
    \begin{subfigure}[t]{0.42\textwidth}
        \centering
        \includegraphics[width=\linewidth]{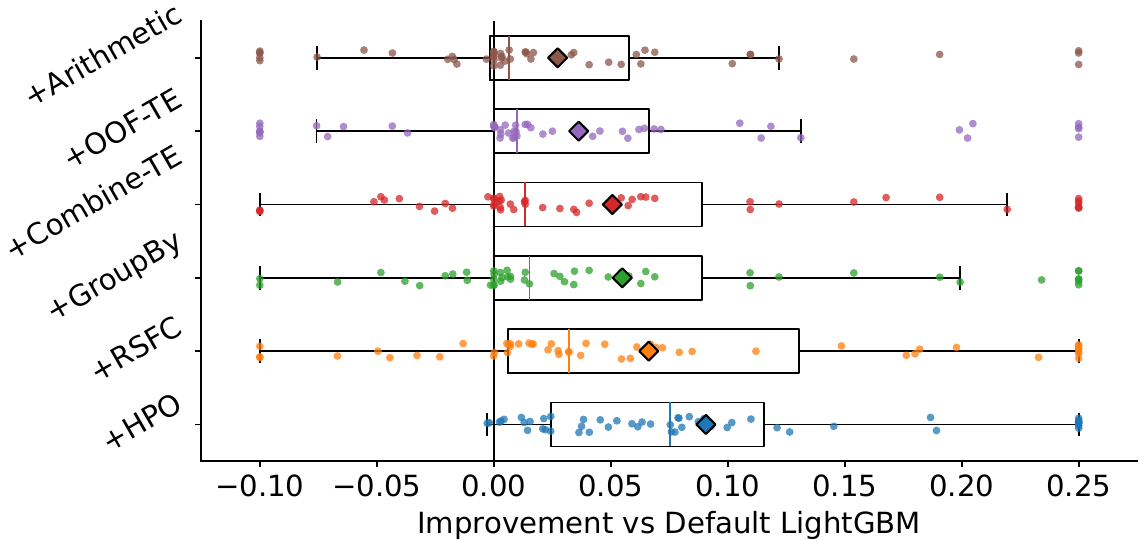}
        \caption{\textbf{Analysis of contribution to performance.}}
        \label{fig:isolate_comparison}
    \end{subfigure}

    \caption{\textbf{Left}: Relative improvement of TabPrep compared to autoFE libraries. \textbf{Right}: Relative improvement of each generator when sequentially added to a default LightGBM model.
    }
    \label{fig:autofe_and_isolate}
\end{figure}

We compare TabPrep against two established autoFE libraries, autofeat~\cite{horn2019autofeat} and OpenFE~\cite{zhang2023openfe}. Both follow the expand-and-reduce paradigm and rely on supervised selection using linear models for autofeat and LightGBM for OpenFE. To benchmark both libraries at their best, we evaluate autofeat against PrepLinearModel and OpenFE against PrepLightGBM. \autoref{fig:autofe_main} shows that TabPrep outperforms both approaches despite avoiding exhaustive search and supervised feature selection, demonstrating that informed feature generation can be a stronger baseline than expand-and-reduce autoFE. Note that we only compare performance on datasets where the libraries ran successfully to distinguish implementation quality from methodological quality. autofeat and OpenFE fail due to memory or time limits on $33$ and $22$ datasets, respectively, whereas TabPrep runs on all datasets. Even on successful runs, the median single-model training-time increase is $180\times$ for autofeat and $19\times$ for OpenFE, compared to only $6\times$ for TabPrep with LightGBM. Thus, existing autoFE libraries remain impractical for HPO-based benchmarking, whereas TabPrep provides a scalable feature-engineering baseline with manageable overhead justified by improved peak performance. For more details on the comparison, see Appendix \ref{appendix:autofe_time}.

\subsection{Practicality Analysis on TabPrep for Benchmarking}

\textbf{TabPrep Excels on Medium-Sized Datasets.} \hspace{1.5mm}
A potential concern with feature expansion is that it may become impractical on larger datasets. Counterintuitively, \autoref{fig:elo_full_medium}~(right) shows that all studied non-linear models augmented with TabPrep benefit even more on medium-sized datasets with 10{,}000--100{,}000 training samples. While the full benchmark is skewed toward smaller datasets where foundation models are already strong, TabPrep yields particularly large gains in the medium-data regime, enabling even LightGBM to outperform AutoGluon's multi-model ensemble~\cite{erickson-arxiv20a}. This can be attributed to the fact that with more samples latent structure is better supported by evidence and the risk of overfitting decreases, making bias reduction through explicit feature construction more valuable. Although TabPFN-2.5 benefits considerably and the model is applicable up to 100{,}000 samples, the attention-based architecture faces practical scaling constraints in both sample size and dimensionality, leading to LightGBM and TabM performing better. Overall, these results indicate that feature-engineered tree-based models remain highly competitive in the medium-data regime. 

\textbf{Compute--Performance Trade-offs.} \hspace{1.5mm} All autoFE methods inevitably increase compute cost. 
\begin{wrapfigure}[23]{r}{0.445\textwidth}
    \centering
    \includegraphics[width=\linewidth]{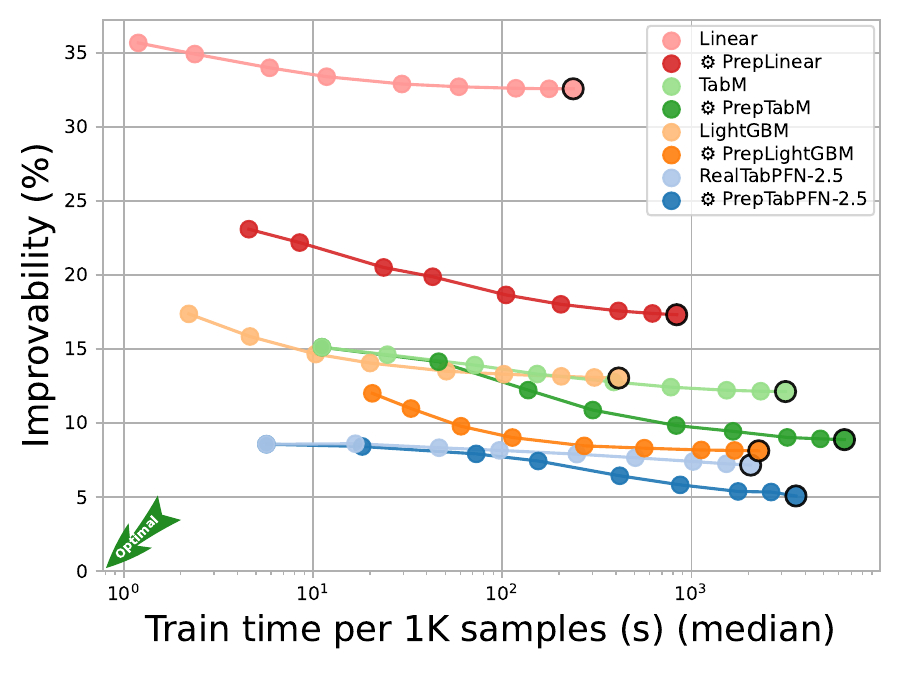}
    \caption{\textbf{Improvability tuning trajectories.} Points from left to right mark ensembles of increasing numbers of random configurations (1, 2, 5, 10, 25, 50, 100, 150, 201). The trajectories are sampled 20 times from all trials and averaged. Improvability (lower is better) measures how many percent lower the error of the best method is than the current method on a dataset, averaged over datasets.}
    \label{fig:parteo_tuning_over_time}
\end{wrapfigure}
Nevertheless, our results show that, while the cost of training a single TabPrep-augmented model increases, TabPrep improves compute--performance trade-offs at the benchmark level. 
Comparing TabPrep-augmented models to their non-augmented counterparts in \autoref{fig:parteo_tuning_over_time} reveals that during training, performance gains already appear using the default setting or with small tuning budgets, leading to stronger performance materializing faster along the tuning trajectory. Furthermore, adding TabPrep does not materially slow HPO convergence. 
In addition, TabPrep narrows the gap between lightweight CPU-based models on cheap hardware and GPU-dependent methods, enabling similar performance on commodity hardware without expensive GPUs. 
Finally, while the overhead is naturally not zero, the results suggest that TabPrep is often a practical way to reach stronger performance more efficiently, making it a reasonable addition whenever predictive accuracy is the main objective. 
In the Appendix, we also show that \\
\begin{enumerate*}[label=(\textbf{\arabic*})]
    \item Appendix \ref{app:additional_tradeoffs}:  TabPrep models lie on the Pareto frontier of improvability versus inference cost and the trade-offs are particularly favorable on larger datasets; 
    \item Appendix \ref{app:no_feats_sensitivity}: performance improvements can already materialize with fewer features added (e.g. 100); 
    \item Appendix \ref{app:seed_sensitivity}: the seed sensitivity in random feature generation is negligible on most datasets;
    \item Appendix \ref{app:tabprep_as_hp}: using TabPrep as a hyperparameter effectively prevents performance degradation.
\end{enumerate*}

\section{Discussion and Conclusion}
\label{sec:discussion_conclusion}

\textbf{Impact on Evaluation and Benchmarking.} \hspace{1.5mm}
We proposed TabPrep, a feature engineering baseline targeting three recurring structural patterns in tabular data, and showed that even a simple pipeline can establish new peak performance on general tabular learning benchmarks. TabPrep changes the interpretation of benchmark results by making feature engineering an explicit part of the evaluation protocol rather than an omitted implementation detail. Our results show that purely model-centric comparisons can underestimate peak performance and obscure systematic model blind spots. TabPrep therefore enables future evaluations to test whether new tabular methods are able to learn the studied recurring structural patterns. In this role, TabPrep provides a baseline for measuring whether future models close the feature engineering gap, or whether explicit feature engineering continues to provide complementary gains. At the same time, TabPrep can correct what is considered peak performance in tabular benchmarks.
Moreover, our work can guide future tabular foundation models in their development to effectively (1) learn arithmetic structure, (2) handle (high-cardinality) categorical features, or (3) deal with pseudo-categorical numerical features.

\textbf{Limitations.} \hspace{1.5mm}
\textbf{(1)} TabPrep is designed as a benchmark baseline and does not aim to find optimal features. Consequently, the reported gains should be interpreted as evidence that even lightweight feature engineering impacts benchmark conclusions, rather than as an upper bound on the potential of feature engineering.
\textbf{(2)} We evaluate TabPrep with four representative model families. While this covers major classes of tabular learners and we compare against the broader TabArena model zoo, additional model-specific interactions may remain unexplored.
\textbf{(3)} Feature expansion increases training and inference cost. However, under the TabArena-style benchmark setting, where models are repeatedly trained across datasets, folds, and HPO configurations, our tuning trajectories in \autoref{fig:parteo_tuning_over_time} show that TabPrep can yield better performance at a given compute budget than spending the same compute on HPO without TabPrep.
\textbf{(4)} We focus on three data structures identified from previous benchmarking studies \cite{tschalzev2024data,mcelfresh2023neural}, and therefore provide a selective view. Identifying further structures that can be exposed through feature engineering remains an important direction for future work.
\textbf{(5)} Our evaluation follows TabArena, which focuses on IID tabular prediction tasks and excludes settings such as non-IID data, relational data, or datasets with text fields. Extending to these settings is outside the scope of this work, but we discuss extensions in \autoref{app:beyond_ta_scope}.

\textbf{Conclusion.} \hspace{1.5mm}
Our results show that the feature engineering gap remains a substantial source of missed performance in tabular benchmarks. In the future, TabPrep can provide a simple way to expose this gap and a practical baseline for assessing whether future methods close it.

\bibliography{neurips_2026}
\bibliographystyle{unsrtnat}

\clearpage
\appendix

\section*{Appendices}
\addcontentsline{toc}{section}{Appendix}

\startcontents[appendix]
\printcontents[appendix]{}{1}{\setcounter{tocdepth}{2}}

\clearpage


\section{Experimental Details} \label{appendix:experiments}
\newcommand{\err}{\operatorname{err}}
\newcommand{\besterr}{\operatorname{best\_err}}

\textbf{TabArena Setup.} \hspace{1.5mm}
We adopt the evaluation design of the TabArena benchmark \cite{erickson2025tabarena}. 
We evaluate models on the set of 51 real-world tabular datasets curated in TabArena. These datasets were manually selected from an initial pool of 1{,}053 datasets to represent classification and regression tasks encountered in practice under rigorous selection criteria. \\
TabArena uses a nested cross-validation evaluation protocol to reduce bias and variance in performance estimates. Hyperparameter selection is performed using 8-fold inner cross-validation, while final performance is estimated using repeated outer cross-validation.
Classification tasks use class-stratified splits. 
We use three outer folds to improve on the TabArena-Lite benchmark recommendation to use one outer fold. Therefore, we guard against randomness in the results by evaluating in an $8\times3$ nested cross-validation setup training 24 models per configuration, selecting the best model on 8 inner folds and averaging test results over three repetitions.
Consistent with TabArena, we evaluate one default configuration and a fixed budget of hyperparameter configurations for each tunable model.
Hyperparameter selection is based on inner cross-validation performance. \\
To assess peak achievable performance of each model, TabArena employs post-hoc weighted ensembling over multiple hyperparameter configurations. 
Therefore, each model is evaluated in three setups: (1) default, (2) tuned, and (3) an ensemble over all configurations of a model's tuning process. \\
We compare against the published TabArena baselines, which at the time of writing includes the following $27$ models:
\begin{itemize}
    \item $7$ Tree-based methods: Random Forest \cite{breiman-mlj01a}, XGBoost \cite{chen2016xgboost}, LightGBM \cite{ke2017lightgbm}, CatBoost \cite{prokhorenkova2018catboost}, Extra Trees \cite{geurts-ml06a}, PerpetualBooster \cite{perpetual_ml_perpetual}, and Explainable Boosting Machine \cite{lou2013accurate}.
    \item $5$ Neural network architectures: TabM \cite{gorishniy2024tabm}, RealMLP \cite{holzmuller2024better}, ModernNCA \cite{ye2024modern}, TorchMLP \cite{erickson-arxiv20a}, and FastAIMLP \cite{erickson-arxiv20a}.
    \item $12$ Tabular Foundation Models: TabPFNv2 \cite{hollmann-nature25a}, TabPFN-2.5 \cite{grinsztajn2025tabpfn}, TabPFN-2.6 \cite{priorlabs_tabpfn_2_6}, TabICL \cite{qu2025tabicl}, TabICLv2 \cite{qu2026tabiclv2}, TabDPT \cite{ma2024tabdpt}, TabSTAR \cite{arazi2025tabstar}, SAP-RPT-OSS \cite{spinaci2025contexttab}, Mitra \cite{zhang2025mitra}, Limix \cite{zhang2025limix}, TabFlex \cite{zeng2024tabflex}, and BetaTabPFN \cite{liu2025tabpfn}
    \item $2$ baselines: linear model from AutoGluon \cite{erickson-arxiv20a}, and kNN from AutoGluon \cite{erickson-arxiv20a}
    \item $1$ other: xRFM \cite{beaglehole2025xrfm} 
\end{itemize}
Because not all models are competitive on the benchmark, we only report results for the best models along with some baselines as a reference. 
All baseline results are taken from TabArena. We use the same data splits, validation strategy, and computational constraints to ensure a fair comparison. The TabArena baseline results can be accessed at the \href{official public leaderboard}{https://huggingface.co/spaces/TabArena/leaderboard}.

\textbf{Metrics} \hspace{1.5mm}
Following TabArena, we use the Elo rating as our primary aggregation metric. Elo scores are computed from pairwise model comparisons and assign equal weight to each dataset, avoiding bias toward specific domains or dataset sizes. A 95\% confidence interval is obtained via bootstrapping. Task-specific base metrics are ROC AUC for binary classification, log-loss for multiclass classification, and RMSE for regression. 
Following, the TabArena setup, we visualize the improvability aggregation metrics when plotting training/inference time against performance. We additionally report relative improvement over baseline methods per dataset. Assume that errors $\err_i$ were measured for each dataset $i$ by averaging error metrics (1-AUROC for binary, logloss for multiclass, and RMSE for regression) over all outer folds. The following aggregation metrics are utilized: 
\begin{itemize}[topsep=2pt,itemsep=2pt,parsep=0pt]
    \item \textbf{Elo}: The metric relies on a pairwise comparison-based rating system where each model's rating predicts its expected win probability against others, 
    with a 400-point Elo gap corresponding to a 10 to 1 (91\%) expected win rate. TabArena calibrates 1000 Elo to the performance of the default random forest configuration, and performs 200 rounds of bootstrapping to obtain 95\% confidence intervals. Higher is better.
    \item \textbf{Normalized score}: For each dataset, we linearly rescale the error such that the best method has a normalized score of one, and a worse performing reference method (median, if not stated otherwise) has a normalized score of 0. Scores below zero are clipped to zero. When aggregating over datasets, these scores are averaged across datasets. Higher is better.
    \item \textbf{Relative Improvement}: We report relative improvement of a competitor model $c$ on a dataset $i$ as $\frac{e_i - e_{i,c}}{|e_i|}$, where \(e_i\) is the baseline model's error and \(e_{i,c}\) is the competitor's error. Higher is better.
    \item \textbf{Improvability}: The metric measures ($[0\%,100\%]$) how many percent lower the error of the best method is than the current method on a dataset. This is then averaged over datasets. Formally, for a single dataset,
    \begin{equation*}
        \operatorname{Improvability} := \frac{\err_i - \besterr_i}{\err_i} \cdot 100\%~.
    \end{equation*}
    Lower is better.    
\end{itemize}

\textbf{TabPrep Experiments.}  \hspace{1.5mm}
All experiments are conducted using the fixed dataset versions, splits, and evaluation scripts defined by TabArena, and we execute the TabArena codebase to obtain results. 
We evaluate our proposed TabPrep pipeline in combination with four models: (1) a linear model from AutoGluon \cite{erickson-arxiv20a}, (2) LightGBM \cite{ke2017lightgbm}, (3) TabM \cite{gorishniy2024tabm}, and (4) RealTabPFN-2.5 \cite{grinsztajn2025tabpfn}. Note that while we sometimes use abbreviations like TabPFN-2.5, TabPFN, or PFN, we used RealTabPFN-2.5 for all experiments where we combine with TabPrep.
We apply TabPrep separately to the training data of each inner fold and exclude the validation data of each inner cross-validation fold for feature engineering. Therefore, just as the test data, the validation data is treated as unseen to obtain unbiased validation estimates and avoid data leakage.  \\
Whenever the number of new features exceeds 2000 after applying TabPrep, we retain the 2000 features with the highest absolute Spearman correlation with the target, computed only on the training samples of the corresponding fold. This filtering step is not part of the proposed TabPrep pipeline, but is used solely for evaluation to avoid out-of-memory failures and to ensure comparability with TabPFN-2.5, which is limited to 2000 features. Since the filtering is applied only after feature generation, only on training data, and only when the feature cap is exceeded, it does not affect the leakage-free evaluation protocol. In practice, we recommend using as many generated features as feasible under the available hardware and time constraints, and pruning features based on feature importance when deployment-optimized models are the goal.
Although we benchmark our pipeline for four models, we compare to every model currently in the TabArena benchmark, which at the time of writing are $27$ models, trained with up to $200$ configurations.
Whenever we compare results to models without our feature engineering, we use the publicly available, precomputed results released with the benchmark, where each model was evaluated with 200 configurations for hyperparameter optimization. 
For the ablations we evaluate models with default hyperparameters while for the plots comparing performance on benchmark and dataset level, we use the tuned and post-hoc ensembled versions. 

\textbf{Hyperparameter Search Spaces}  \hspace{1.5mm}
Tables \ref{tab:space:tabprep} to \ref{tab:space:tabm} show the used hyperparameter search spaces for TabPrep and the models. We evaluate the TabPrep pipeline in the following hyperparameter search setup: 
TabPrep is treated as a hyperparameter sampling 25\% out of 200 configurations where TabPrep is enabled, such that 75\% of the configurations correspond to the non-augmented models.
Note that we reuse the exact same hyperparameter search for model parameters as was used in TabArena. Therefore, all configurations not augmented with TabPrep exactly match the TabArena version, allowing to reuse the results published by the TabArena authors.
For TabPrep, we do not tune hyperparameters exhaustively, to emphasize the general applicability of our baselines. The search space is in \autoref{tab:space:tabprep}.
We only tune the number of generated features and the random order in which features are selected. For OAFE, categorical combinations, and RSFC this means that different feature pairs are sampled. For OOF-TE this means varying the distribution of samples to train/valid for the internal cross-validation. Note that tuning specific random generation hyperparameters is common practice in ML. For example, LightGBM offers the possibility to tune the bagging seed or the extra tree seed; TabPFN allows varying the feature permutation order; and for neural networks varying the sample batch order or initialization seed is possible. In our case, we ablate our tuning decisions in Appendix \ref{app:no_feats_sensitivity} and \ref{app:seed_sensitivity}.

\begin{table}[!h]
    \caption{Hyperparameter search space for TabPrep. The first entry in each set corresponds to the default.} \label{tab:space:tabprep}
    \centering
    \begin{tabular}{ll}
    \toprule
    Hyperparameter & Space \\
    \midrule
    oafe\_max\_feats & Choice([2000, 1000]) \\
    oafe\_generation\_seed & Choice([42, 84, 168, 336, 672]) \\    
    cat\_combine\_max\_feats & Choice([100, 500]) \\
    cat\_combine\_generation\_seed & Choice([42, 84, 168, 336, 672]) \\
    rsfc\_n\_subsets & Choice([50, 1, 100]) \\
    rsfc\_generation\_seed & Choice([42, 84, 168, 336, 672]) \\
    oofte\_generation\_seed & Choice([42, 84, 168, 336, 672]) \\
    groupby\_max\_feats & Choice([500, 100, 1000]) \\    
    \bottomrule
    \end{tabular}
\end{table}

\begin{table}[!h]
    \caption{Hyperparameter search space for LightGBM.} \label{tab:space:lightgbm}
    \centering
    \begin{tabular}{ll}
    \toprule
    Hyperparameter & Space \\
    \midrule
    \texttt{learning\_rate} & LogUniform([0.005, 0.1]) \\
    \texttt{feature\_fraction} & Uniform([0.4, 1.0]) \\
    \texttt{bagging\_fraction} & Uniform([0.7, 1.0]) \\
    \texttt{bagging\_freq} & 1 \\
    \texttt{num\_leaves} & LogUniformInt([2, 200]) \\
    \texttt{min\_data\_in\_leaf} & LogUniformInt([1, 64]) \\
    \texttt{extra\_trees} & Choice([False, True]) \\
    \texttt{min\_data\_per\_group} & LogUniformInt([2, 100]) \\
    \texttt{cat\_l2} & LogUniform([0.005, 2]) \\
    \texttt{cat\_smooth} & LogUniform([0.001, 100]) \\
    \texttt{max\_cat\_to\_onehot} & LogUniformInt([8, 100]) \\
    \texttt{lambda\_l1} & Uniform([1e-4, 1.0]) \\
    \texttt{lambda\_l2} & Uniform([1e-4, 2.0]) \\
    \bottomrule
    \end{tabular}
\end{table}

\begin{table}[!h]
    \caption{Hyperparameter search space for TabPFN-2.5.} \label{tab:space:tabpfn}
    \centering
    \begin{tabular}{p{0.45\linewidth} p{0.5\linewidth}}
    \hline
    \textbf{Parameter} & \textbf{Values} \\
    \hline
    \texttt{model\_used} &
    Choice([One of 8 models for classification and \\ & 6 for regression]) \\
    
    \texttt{softmax\_temperature} &
    Choice([0.25, 0.5, 0.6, 0.7, 0.8, 0.9, 1.0, 1.25, 1.5]) \\
    
    \texttt{balance\_probabilities} &
    Choice([True, False]) \\
    
    \texttt{outlier\_removal\_std} &
    Choice([3, 6, 12]) \\
    
    \texttt{polynomial\_features} &
    Choice(["no", 25]) \\
    
    \texttt{regression\_y\_preprocess\_transforms} &
    Choice([None, (None, "safepower"), "safepower", "quantile\_uni"]) \\
    
    \texttt{numerical\_scaling} &
    Choice([None, "quantile\_uni\_coarse", "quantile\_norm\_coarse", "safepower", "quantile\_uni", (None, "quantile\_uni\_coarse"), ("squashing\_scaler\_default", "quantile\_uni\_coarse"), "squashing\_scaler\_default"]) \\
    
    \texttt{categorical\_encoding} &
    Choice(["numeric", "onehot", None]) \\
    
    \texttt{preprocessing\_append\_original} &
    Choice([True, False]) \\
    
    \texttt{preprocessing\_global} &
    Choice([None, "svd", "svd\_quarter\_components"]) \\
    
    \hline
    \end{tabular}
\end{table}

\begin{table}[!h]
    \caption{Hyperparameter search space for the linear model. The parameter \texttt{C} is set deterministically depending on the number of features $d$: 0.001 if $d>1000$, $0.01$ if $d>500$, 0.1 if $d>100$, and 1 otherwise. The parameter \texttt{C\_scale} is used to scale the deterministically set constant in hyperparameter optimization.} \label{tab:space:linear}
    \centering
    \begin{tabular}{ll}
    \toprule
    Hyperparameter & Space \\
    \midrule
    \texttt{C} & "auto" \\
    \texttt{C\_scale} & Choice([0.001, 0.005, 0.01, 0.05, 0.1, 0.2, 0.3, 0.4, 0.5, 0.6, 0.7, 0.8, 0.9, \\ &  1, 2, 4, 5, 6, 8, 10]) \\
    \texttt{proc.impute\_strategy} & Choice(["median", "mean"]) \\
    \texttt{penalty} & Choice(["L2", "L1"]) \\
    \texttt{scaler} & Choice(["standard", "squashing", "quantile-normal"]) \\
    \bottomrule
    \end{tabular}
\end{table}

\begin{table}[!h]
    \caption{Hyperparameter search space for CatBoost.} \label{tab:space:catboost}
    \centering
    \begin{tabular}{ll}
    \toprule
    Hyperparameter & Space \\
    \midrule
    \texttt{learning\_rate} & LogUniform([0.005, 0.1]) \\
    \texttt{bootstrap\_type} & Bernoulli \\
    \texttt{subsample} & Uniform([0.7, 1.0]) \\
    \texttt{grow\_policy} & Choice(["SymmetricTree", "Depthwise"]) \\
    \texttt{depth} & UniformInt([4, 8]) \\
    \texttt{colsample\_bylevel} & Uniform([0.85, 1.0]) \\
    \texttt{l2\_leaf\_reg} & LogUniform([1e-4, 5]) \\
    \texttt{leaf\_estimation\_iterations} & LogUniformInt([1, 20]) \\
    \texttt{one\_hot\_max\_size} & LogUniformInt([8, 100]) \\
    \texttt{model\_size\_reg} & LogUniform([0.1, 1.5]) \\
    \texttt{max\_ctr\_complexity} & UniformInt([2, 5]) \\
    \texttt{boosting\_type} & Plain \\
    \texttt{max\_bin} & 254 \\
    \bottomrule
    \end{tabular}

\end{table}

\begin{table}[!h]
    \caption{Hyperparameter search space for TabM.} \label{tab:space:tabm}
    \centering
    \begin{tabular}{ll}
    \toprule
    Hyperparameter & Space \\
    \midrule
    \texttt{batch\_size} & auto \\
    \texttt{patience} & 16 \\
    \texttt{amp} & False \\
    \texttt{arch\_type} & tabm-mini \\
    \texttt{tabm\_k} & 32 \\
    \texttt{gradient\_clipping\_norm} & 1.0 \\
    \texttt{share\_training\_batches} & False \\
    \texttt{lr} & LogUniform([1e-4, 3e-3]) \\
    \texttt{weight\_decay} & Choice([0.0, LogUniform([1e-4, 1e-1])]) \\
    \texttt{n\_blocks} & UniformInt([2, 5]) \\
    \texttt{d\_block} & Choice([128, 144, 160, \ldots, 1008, 1024]) \\
    \texttt{dropout} & Choice([0.0, Uniform([0.0, 0.5])]) \\
    \texttt{num\_emb\_type} & pwl \\
    \texttt{d\_embedding} & Choice([8, 12, 16, 20, 24, 28, 32]) \\
    \texttt{num\_emb\_n\_bins} & UniformInt([2, 128]) \\
    \bottomrule
    \end{tabular}

\end{table}

\clearpage

\section{Additional Results}

\subsection{TabArena Results on Different Data Subsets.} \label{app:additional_results_subsets}
\begin{figure}[!h]
    \centering
    \includegraphics[width=\linewidth]{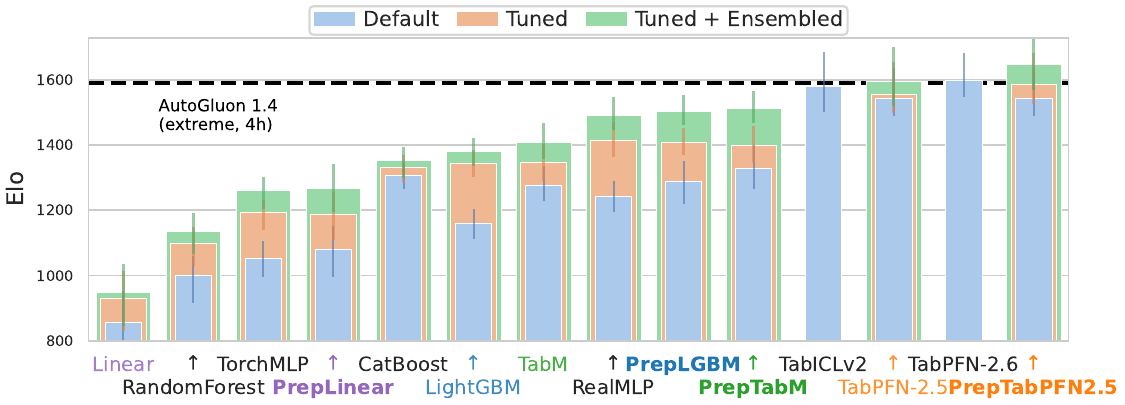}
    \caption{\textbf{Performance of models augmented with TabPrep on the small datasets of the TabArena benchmark.} Elo scores (higher is better) are averaged over datasets comparing default models, tuned models, and tuned+ensembled variants. Models labeled with \textit{Prep} use TabPrep for data preprocessing, while all other results are taken from the official TabArena benchmark \cite{erickson2025tabarena}. Results on the 35 small TabArena datasets, below 10,000 training samples.}
    \label{fig:elo_full_small}
\end{figure}

\begin{figure}[!h]
    \centering
    \includegraphics[width=\linewidth]{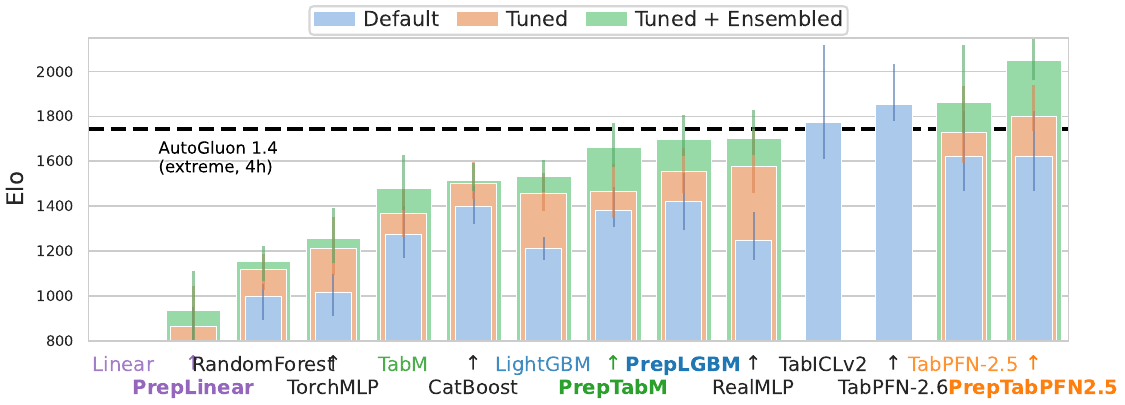}
    \caption{\textbf{Performance of models augmented with TabPrep on the regression datasets of the TabArena benchmark.} Elo scores (higher is better) are averaged over datasets comparing default models, tuned models, and tuned+ensembled variants. Models labeled with \textit{Prep} use TabPrep for data preprocessing, while all other results are taken from the official TabArena benchmark \cite{erickson2025tabarena}. }
    \label{fig:elo_full_reg}
\end{figure}

\begin{figure}[!h]
    \centering
    \includegraphics[width=\linewidth]{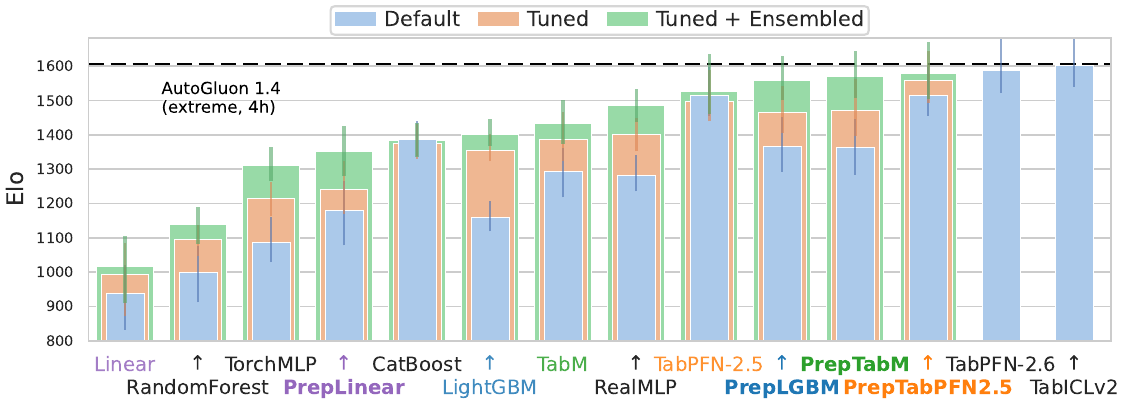}
    \caption{\textbf{Performance of models augmented with TabPrep on the binary datasets of the TabArena benchmark.} Elo scores (higher is better) are averaged over datasets comparing default models, tuned models, and tuned+ensembled variants. Models labeled with \textit{Prep} use TabPrep for data preprocessing, while all other results are taken from the official TabArena benchmark \cite{erickson2025tabarena}.}
    \label{fig:elo_full_bin}
\end{figure}

\begin{figure}[!h]
    \centering
    \includegraphics[width=\linewidth]{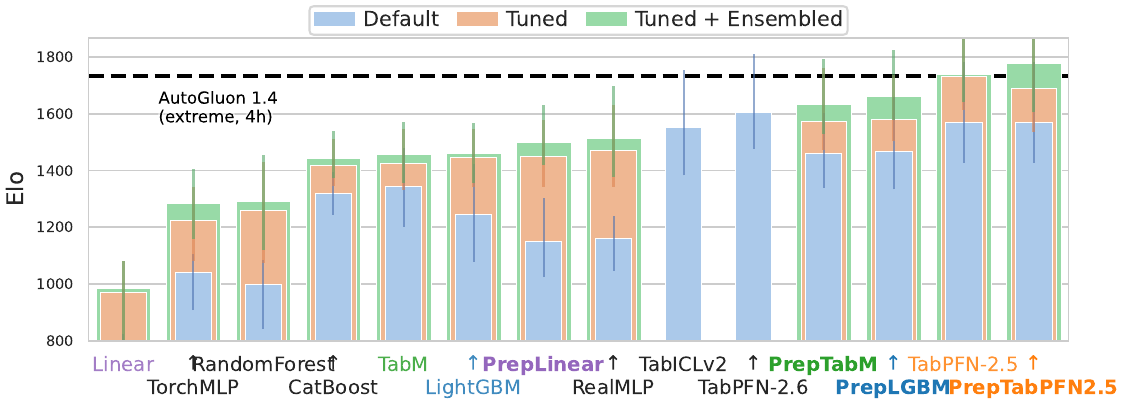}
    \caption{\textbf{Performance of models augmented with TabPrep on the multi-class datasets of the TabArena benchmark.} Elo scores (higher is better) are averaged over datasets comparing default models, tuned models, and tuned+ensembled variants. Models labeled with \textit{Prep} use TabPrep for data preprocessing, while all other results are taken from the official TabArena benchmark \cite{erickson2025tabarena}.}
    \label{fig:elo_full_multi}
\end{figure}

Figures~\ref{fig:elo_full_small}--\ref{fig:elo_full_multi} break down the TabArena results by dataset size and task type. 
Across all subsets, the overall conclusion is consistent with the full benchmark results: augmenting models with TabPrep improves performance across model families. 
On the 35 small datasets, PrepTabPFN-2.5 remains the strongest overall method, while the TabPrep-augmented variants of LightGBM and TabM substantially reduce the gap to the foundation-model baselines. 
On the regression datasets, TabPrep is particularly beneficial for PrepTabPFN-2.5. 
The binary and multi-class classification subsets show the same qualitative pattern, although the relative advantage of TabPrep varies across model families and is smaller for models that are already highly competitive without additional feature engineering.
Overall, these subset results indicate that the gains from TabPrep are not driven by a single task type or dataset-size regime, but reflect a consistent improvement across heterogeneous parts of TabArena.

\subsection{Additional Results on Performance-Efficiency Trade-Offs.} \label{app:additional_tradeoffs}
\begin{figure}[!h]
    \centering
    \includegraphics[width=0.6\linewidth]{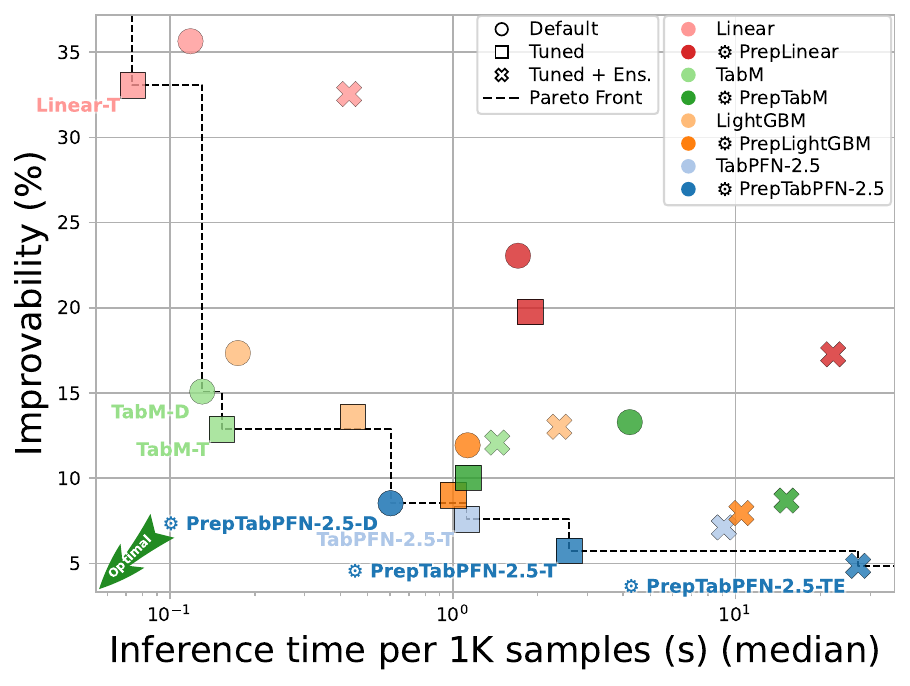}
    \caption{\textbf{Pareto frontier of improvability and inference time}. Following the TabArena benchmark, we report the median inference time per 1000 samples across all datasets.}
    \label{fig:pareto_infer_improv}
\end{figure}

\begin{figure}[!h]
    \centering

    \begin{subfigure}{0.49\linewidth}
        \centering
        \includegraphics[width=\linewidth]{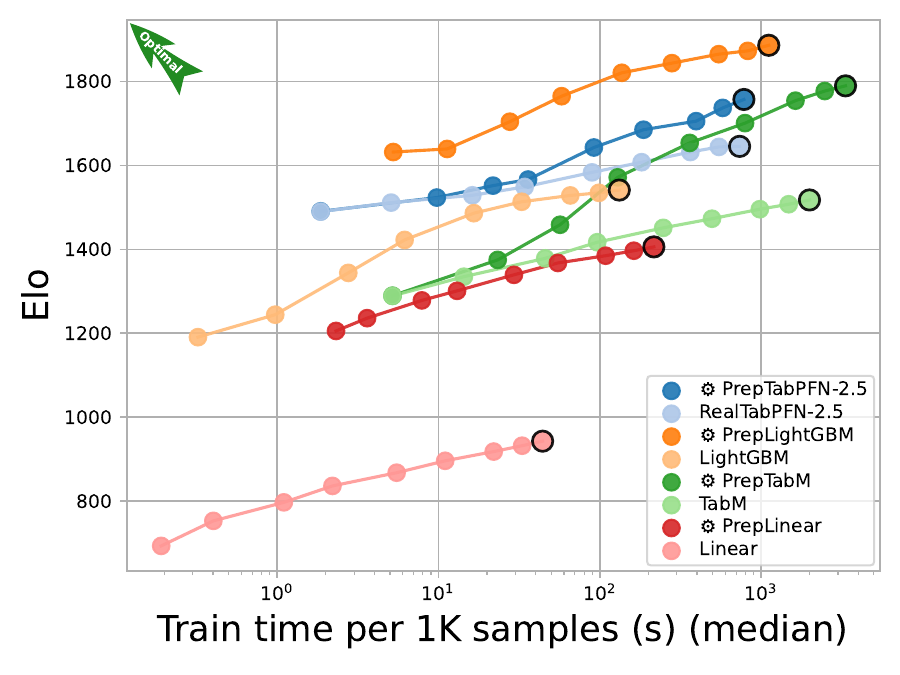}
        \label{fig:pareto-n-configs-elo-train}
    \end{subfigure}
    \hfill
    \begin{subfigure}{0.49\linewidth}
        \centering
        \includegraphics[width=\linewidth]{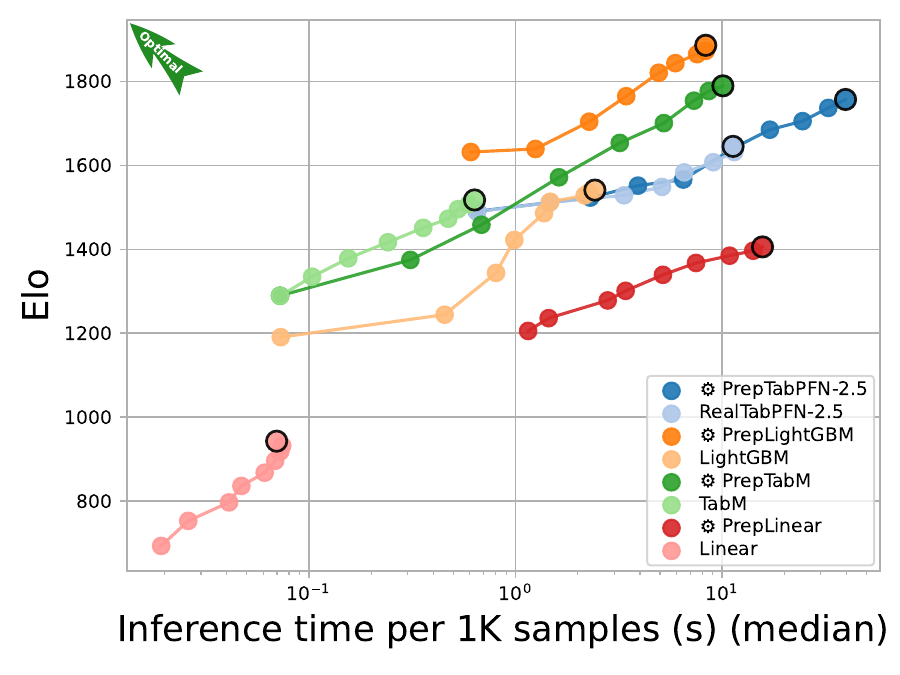}
        \label{fig:pareto-n-configs-elo-infer}
    \end{subfigure}

    \caption{\textbf{Elo tuning trajectories for the 15 largest datasets in TabArena.} Left: training-time tuning trajectories. Right: inference-time tuning trajectories. Points from left to right mark ensembles of increasing numbers of random configurations (1, 2, 5, 10, 25, 50, 100, 150, 201), sampled 20 times from all trials and averaged. The right-most highlighted points use all configurations. Higher is better for the Elo metric.}
    \label{fig:pareto-n-configs-elo}
\end{figure}

\autoref{fig:pareto_infer_improv} provides a complementary view of the performance–efficiency trade-offs focusing on inference time. It can be seen that strong performance gains can be achieved using TabPrep without excessive inference-time overhead, highlighting methods that lie on the Pareto frontier of improvability versus inference cost. E.g., tuned PrepTabPFN-2.5 model has a median throughput of $~30,000$ samples/minute, making it a valid solution for most tabular data tasks. Note that this sample size already covers >40 of all datasets in TabArena -- in one minute. 
Furthermore, it can be seen that LightGBM, although not exactly on the frontier, is close to the non-augmented TabPFN-2.5 and therefore offers the possibility to achieve similar performance with \textit{much} cheaper hardware.
All in all, even if inference time increases, it stays within a reasonable amount offering favorable trade-offs.
Furthermore, \autoref{fig:pareto-n-configs-elo} shows that on medium-sized datasets, LightGBM Pareto dominates all other models at its given compute time (both training and inference) appearing on the Pareto frontier at any tuning budget.

\subsection{Win-Rate Analysis} \label{app:win_rates}
\begin{figure}[!h]
    \centering
    \includegraphics[width=\columnwidth]{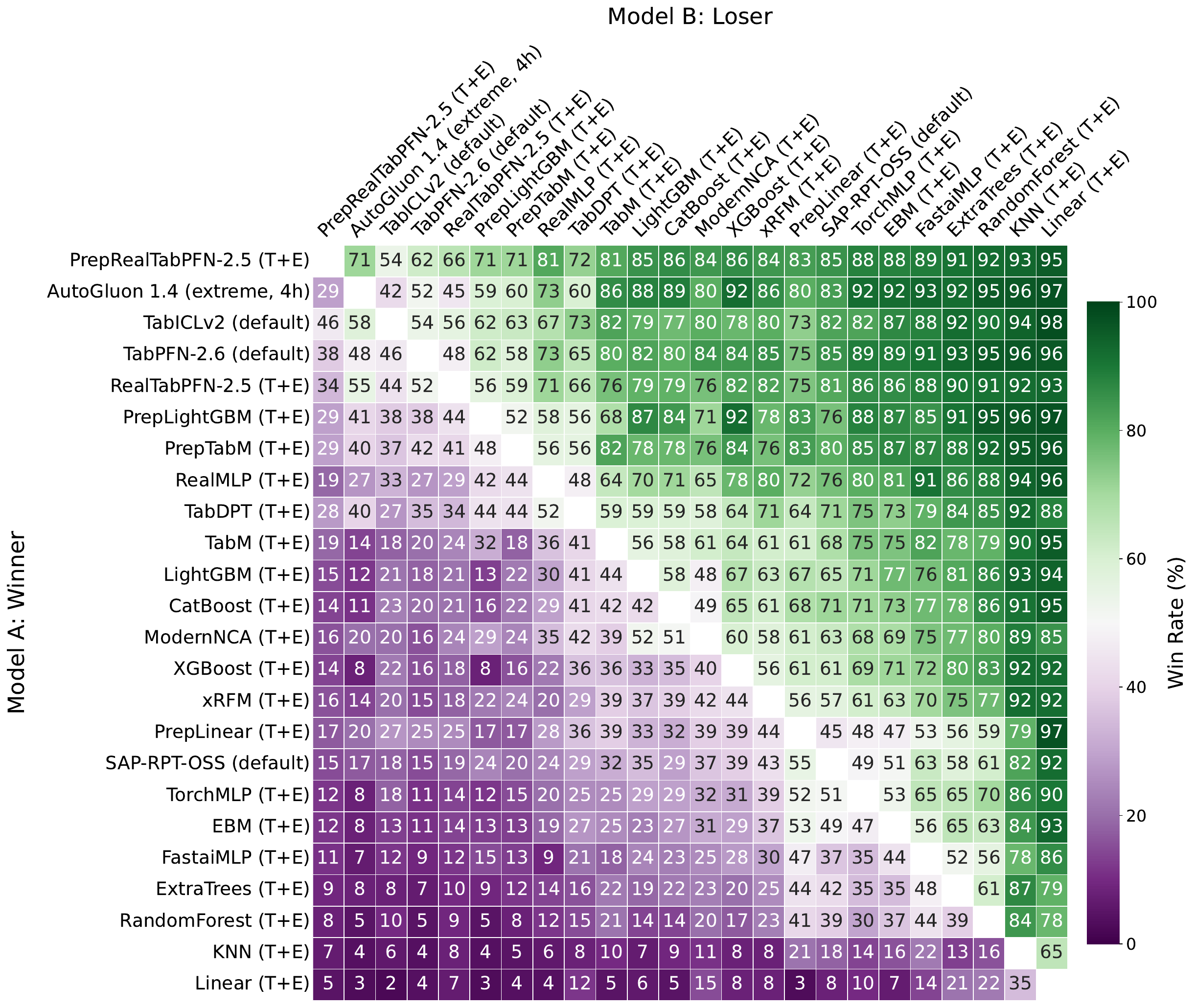}
    \caption{\textbf{Pairwise win rate comparison for all datasets.} Higher numbers correspond to a better win rate for the model on the y-axis.}
    \label{fig:winrate_all}
\end{figure}

\autoref{fig:winrate_all} presents a global pairwise comparison across all 51 TabArena datasets. The win-rate matrix confirms the relative ranking suggested by the Pareto and tuning analyses, showing consistent advantages for the strongest methods across a wide range of dataset pairs, rather than isolated wins on a small subset of tasks. In particular, PrepLinear has a 97\% winrate vs. Linear, PrepLightGBM has an 85\% winrate vs. LightGBM, PrepTabM has an 82\% winrate vs. TabM, and PrepRealTabPFN-2.5 has a 66\% winrate vs. RealTabPFN-2.5. Moreover, note that while PrepTabPFN-2.5 only wins over TabICLv2 in 54\% of cases, \autoref{fig:performance_across_datasets} reveals that both approaches improve the state-of-the-art on distinct datasets, making both valuable in the current portfolio of models.

\subsection{Sensitivity to the Number of Generated Features} \label{app:no_feats_sensitivity}
\begin{figure}[!h]
    \centering
    \includegraphics[width=0.75\columnwidth]{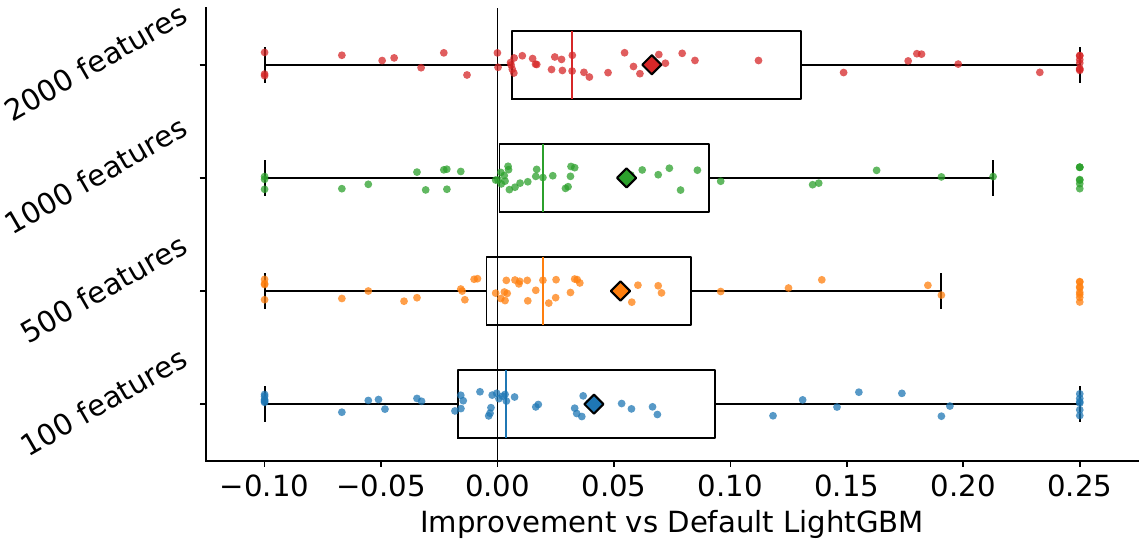}
    \caption{\textbf{Sensitivity analysis of TabPrep improvements with varying the number of added features.}}
    \label{fig:feature_size}
\end{figure}

\autoref{fig:feature_size} demonstrates that TabPrep shows favorable trade-offs with respect to feature size. Even 100 generated features are already enough to benefit from TabPrep. Therefore, users can decide how many features they can afford in their given setup.

\subsection{Sensitivity to the Feature Generation Seed} \label{app:seed_sensitivity}
\begin{figure}[h]
    \centering
    \includegraphics[width=0.99\columnwidth]{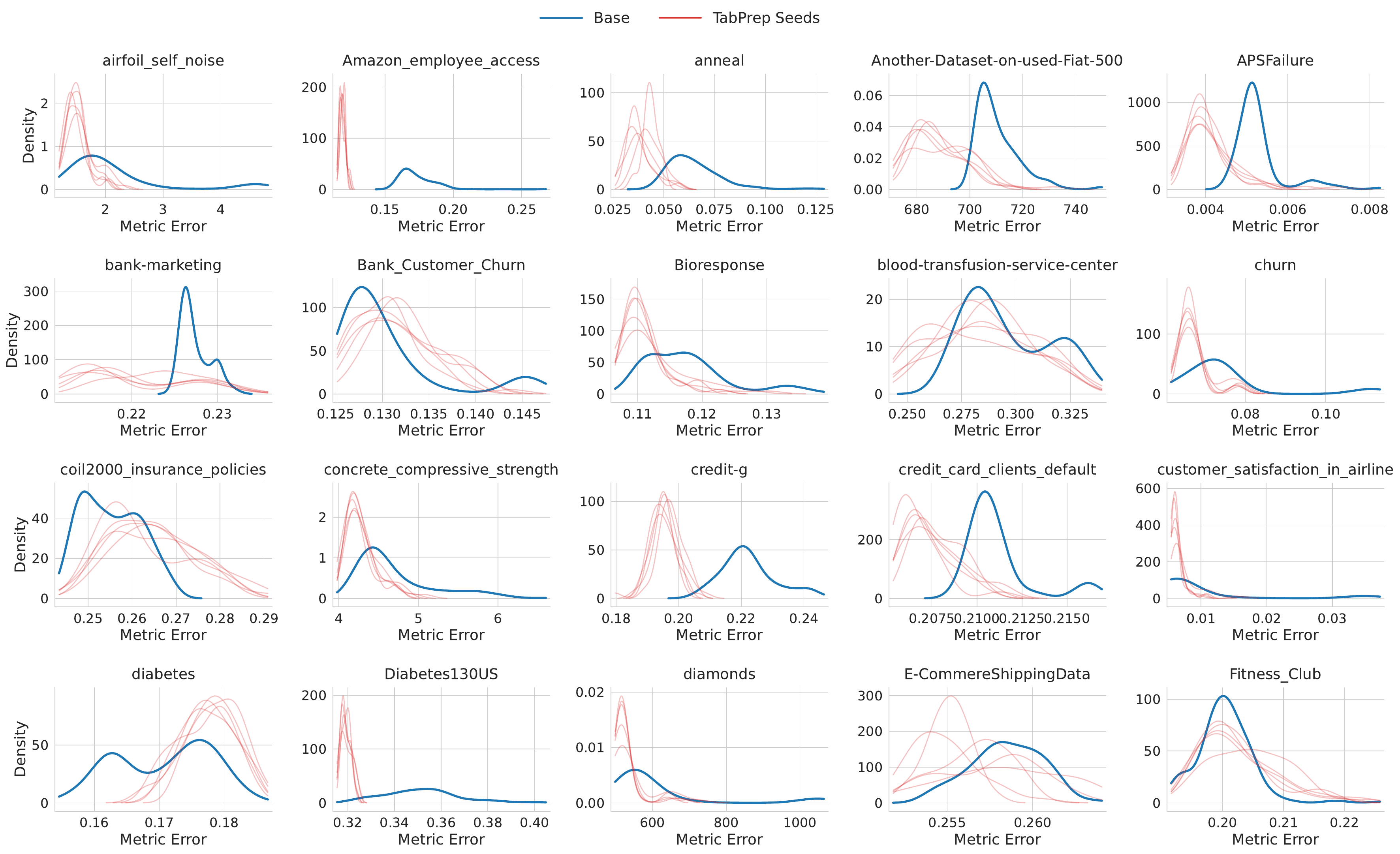}
    \caption{\textbf{Analysis of sensitivity to the random seed in the feature generation of TabPrep.} We examine seed sensitivity for a random batch of 20 TabArena datasets (first 20 when sorting names alphabetically). The blue curves correspond to the error distribution estimated from 200 LightGBM configurations without TabPrep. The red curves are the error distributions obtained with TabPrep under 5 different random seeds for feature generation. Curves with more density on the left side indicate better performance. High separation of the red curves would indicate sensitivity to the seed choice, with some seeds giving clearly different performance than others. TabPrep distributions being more similar to each other than to the non-augmented blue distribution, indicates low sensitivity to the random seed.}
    \label{fig:seed_sensitivity}
\end{figure}

\autoref{fig:seed_sensitivity} demonstrates that TabPrep shows negligible sensitivity to tuning the feature generation seed. Overall, it can be seen that while in some cases the performance distributions may vary depending on the seed, in most cases it does not change the picture that feature engineering is beneficial. Note that this plot also reveals how the feature engineering used in TabPrep generally improves all tested configurations when the structural patterns it is designed for are actually present.

\subsection{Leave-One-Out Contributions for the TabPrep Generators} \label{app:loo_contrib}
\begin{figure}[h]
    \centering
    \includegraphics[width=0.75\columnwidth]{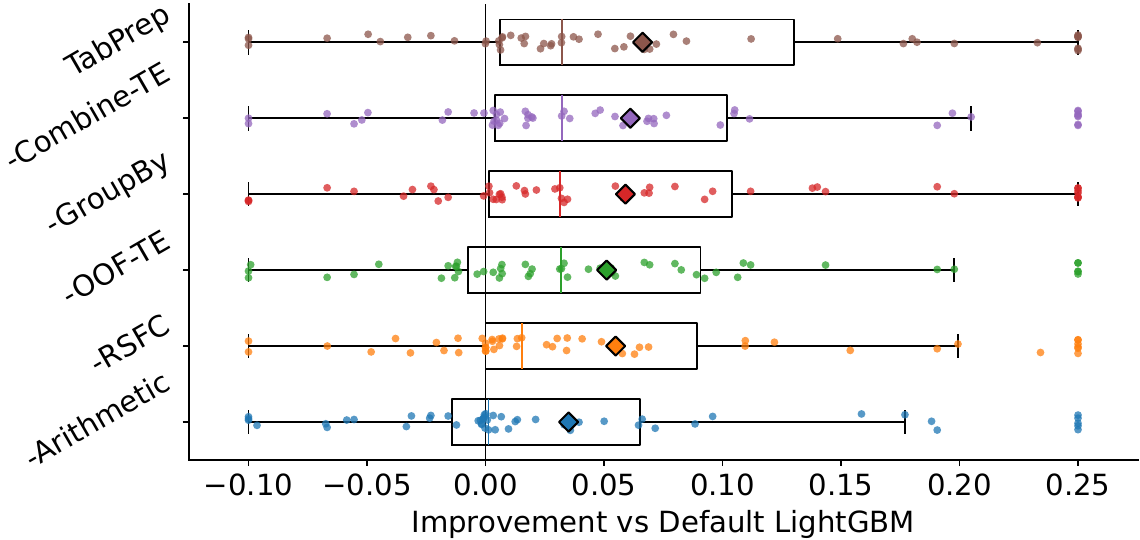}
    \caption{\textbf{Leave-one-out contribution of each preprocessor to performance.}}
    \label{fig:loo_contribution}
\end{figure}

\autoref{fig:loo_contribution} presents additional evidence for the usefulness of each preprocessor. We provide a leave-one-out version of Figure \ref{fig:isolate_comparison}. The experiment confirms that each component contributes to performance, while arithmetic interactions are the most important technique for performance.

\subsection{TabPrep as a Hyperparameter vs. Always On} \label{app:tabprep_as_hp}
\begin{figure}[!t]
    \centering
    \includegraphics[width=0.75\columnwidth]{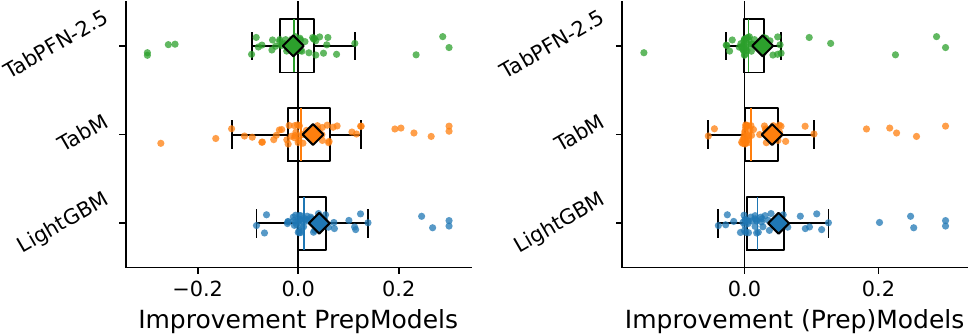}
    \caption{\textbf{Relative improvements of always using TabPrep vs. treating it as a hyperparameter in the search space.} \textit{Left}: PrepModels denotes results from a search space where TabPrep is used in each configuration. \textit{Right}: \textit{(Prep)Models} denotes results from a search space where non-augmented configurations are included. Both are compared to the base models in the tuned + ensembled regime (200 TabArena tuning configs).}
    \label{fig:combined_trials}
\end{figure}

We study how models differ in their ability to benefit from TabPrep.
\autoref{fig:combined_trials} (right) shows that when TabPrep is treated as a hyperparameter, all models benefit from additional trials with advanced preprocessing.
However, for TabM and TabPFN-2.5, just relying on TabPrep is not always better than the baseline.
In contrast, for LightGBM, always using the feature engineering pipeline seems almost universally beneficial, such that PrepLightGBM can be used as a standalone model.
This difference in model behavior can be explained by LightGBM having better feature selection capabilities, as has been shown for tree-based models in general \cite{grinsztajn2022tree}.
This indicates that TabPFN-2.5 shares the inductive bias of limited feature selection abilities with other neural networks.
Moreover, the results confirm our intuition that feature engineering techniques should be treated as a hyperparameter rather than as a universal solution.

\subsection{TabArena-full vs. Our experiments} \label{app:3folds_vs_full}
\begin{figure}[!t]
    \centering
    \begin{subfigure}[t]{0.5\columnwidth}
        \centering
        \includegraphics[width=\linewidth]{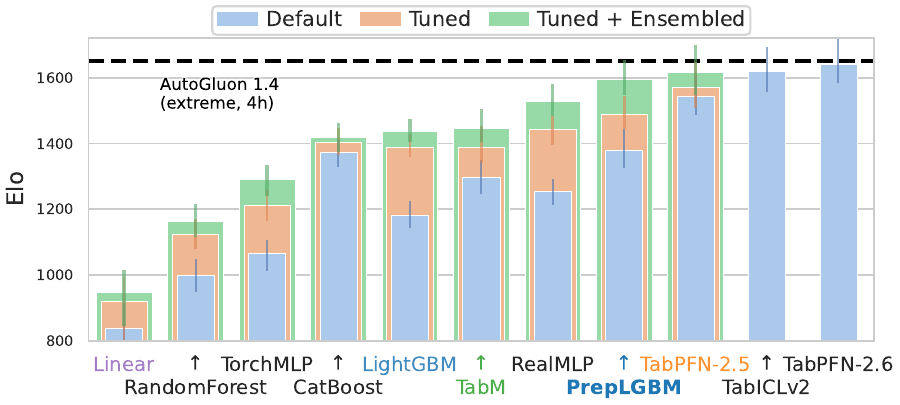}
        \label{fig:tabarena_3folds}
    \end{subfigure}%
    \begin{subfigure}[t]{0.5\columnwidth}
        \centering
        \includegraphics[width=\linewidth]{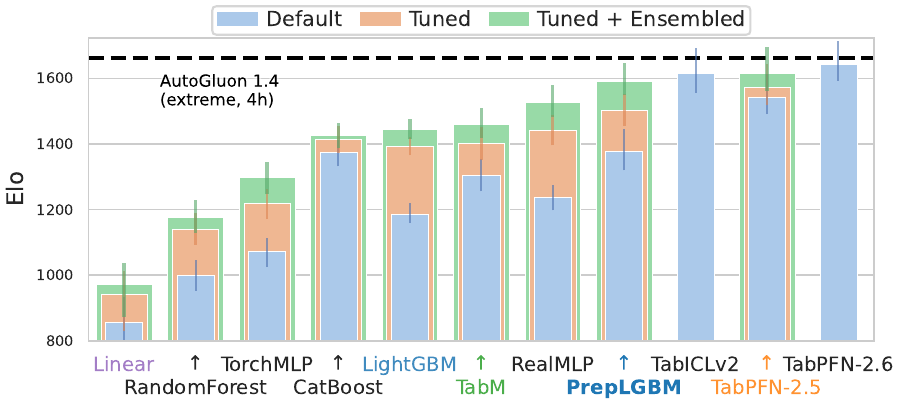}
        \label{fig:tabarena_all_models_2}
    \end{subfigure}

    \caption{\textbf{Performance of LightGBM augmented with TabPrep on the TabArena benchmark.} \textbf{Left:} Results on the first three outer splits of the TabArena benchmark. \textbf{Right:} Results on the full benchmark with 30 outer folds for small, and 9 for medium-sized datasets. Elo scores (higher is better) are averaged over datasets. Models labeled with \textit{Prep} use TabPrep for data preprocessing, while all other results are taken from the official TabArena benchmark \cite{erickson2025tabarena}. }
    \label{fig:elo_3folds_full}
\end{figure}

In our evaluation, we used the first three outer folds of the TabArena benchmark, corresponding to a 3-fold cross-validation setup. The TabArena benchmark contains results for repeated 3-fold cross-validation with evaluation on 30 outer folds for small and 9 for medium-sized datasets. To ensure that our setup with three outer folds is sufficient, we run the LightGBM model on the full TabArena benchmark and compare the results. \autoref{fig:elo_3folds_full} shows that the difference between the two evaluation setups is minor and does not affect our conclusions, confirming the validity of our evaluation setup. 

\subsection{Alternative Aggregation Metrics} \label{app:alternative_metrics}
\autoref{tab:leaderboard} shows that PrepTabPFN leads the leaderboard across all aggregation metrics. In the main paper, we restricted the evaluation to ELO scores, since this is the default TabArena metric. However, the performance improvements of TabPrep are even larger when considering other aggregation metrics. With the improved performance, the runtime of TabPrep-augmented individual models also clearly increases. However, note that the leaderboard only shows results for 200 tuning configurations, while only a few configurations would already be sufficient to realize the performance gains from TabPrep. Finally, the large number of wins and the low harmonic mean rank demonstrate TabPrep's ability to improve peak performance on individual datasets, which is the main objective of TabPrep.


\begin{table}[h!]
\centering
\caption{\textbf{Results on the TabArena Leaderboard.} We show default (D), tuned (T), and tuned + ensembled (T+E) performances. The best three values in columns are highlighted with \textcolor{gold}{gold}, \textcolor{silver}{silver}, and \textcolor{bronze}{bronze} colors. For Elo values, we also indicate their approximate 95\% confidence intervals obtained through bootstrapping.} 
\label{tab:leaderboard}
\resizebox{\textwidth}{!}{
    \addtolength{\tabcolsep}{-0.4em}

\begin{tabular}{llcccccrr}
\toprule
\small
\textbf{Model} & \textbf{Elo ($\uparrow$)} & \textbf{Norm.} & \textbf{Avg.} & \textbf{Harm.} & \textbf{\#wins ($\uparrow$)} & \textbf{Improva-} & \textbf{Train time} & \textbf{Predict time} \\
 &  & \textbf{score ($\uparrow$)} & \textbf{rank ($\downarrow$)} & \textbf{mean} &  & \textbf{bility ($\downarrow$)} & \textbf{per 1K [s]} & \textbf{per 1K [s]} \\
 &  &  &  & \textbf{rank ($\downarrow$)} &  &  &  &  \\
\midrule
PrepTabPFN-2.5 (T+E) & \textcolor{gold}{\textbf{1663${}_{-65,+98}$}} & \textcolor{gold}{\textbf{0.719}} & \textcolor{gold}{\textbf{10.4}} & \textcolor{gold}{\textbf{3.0}} & \textcolor{gold}{\textbf{10.0}} & \textcolor{gold}{\textbf{5.5\%}} & 3626.44 & 27.14 \\
AutoGluon 1.4  & \textcolor{silver}{\textbf{1624${}_{-50,+50}$}} & 0.614 & \textcolor{silver}{\textbf{12.1}} & 5.9 & 2.33 & 8.3\% & 570.73 & 6.16 \\
TabPFN-2.6 (D) & \textcolor{bronze}{\textbf{1616${}_{-54,+75}$}} & \textcolor{bronze}{\textbf{0.626}} & \textcolor{bronze}{\textbf{12.4}} & 5.5 & \textcolor{bronze}{\textbf{3.0}} & 8.4\% & 5.49 & 0.58 \\
PrepTabPFN-2.5 (T) & 1602${}_{-53,+74}$ & 0.615 & 13.2 & 5.5 & 1.83 & \textcolor{silver}{\textbf{6.4\%}} & 3626.44 & 2.58 \\
TabICLv2 (D) & 1600${}_{-59,+79}$ & \textcolor{silver}{\textbf{0.666}} & 13.2 & \textcolor{silver}{\textbf{3.9}} & \textcolor{silver}{\textbf{6.66}} & \textcolor{bronze}{\textbf{7.6\%}} & 4.16 & 0.36 \\
RealTabPFN-2.5 (T+E) & 1596${}_{-65,+83}$ & 0.611 & 13.5 & \textcolor{bronze}{\textbf{5.4}} & 1.66 & 7.8\% & 2058.44 & 9.10 \\
PrepLightGBM (T+E) & 1578${}_{-43,+53}$ & 0.524 & 14.4 & 5.9 & 2.33 & 8.4\% & 2407.89 & 10.49 \\
PrepTabM (T+E) & 1574${}_{-45,+59}$ & 0.529 & 14.6 & 7.3 & 0.99 & 9.2\% & 6356.41 & 15.14 \\
RealTabPFN-2.5 (T) & 1552${}_{-59,+75}$ & 0.546 & 15.8 & 6.8 & 1.833 & 8.3\% & 2058.44 & 1.12 \\
RealTabPFN-2.5 (D) & 1521${}_{-52,+66}$ & 0.509 & 17.5 & 9.2 & 0.66 & 9.2\% & 5.68 & 0.60 \\
RealMLP (T+E) & 1509${}_{-45,+45}$ & 0.431 & 18.2 & 11.4 & 0.66 & 10.8\% & 2958.86 & 11.40 \\
PrepLightGBM (T) & 1479${}_{-44,+51}$ & 0.381 & 20.1 & 8.2 & 1.333 & 9.5\% & 2407.89 & 1.00 \\
TabDPT (T+E) & 1467${}_{-53,+63}$ & 0.417 & 20.9 & 6.5 & 2.66 & 11.1\% & 5467.20 & 316.26 \\
PrepTabM (T) & 1463${}_{-51,+57}$ & 0.388 & 21.1 & 11.5 & 0.333 & 10.5\% & 6356.41 & 1.13 \\
RealMLP (T) & 1427${}_{-48,+38}$ & 0.316 & 23.5 & 13.2 & 0.0 & 12.0\% & 2958.86 & 0.52 \\
TabM (T+E) & 1425${}_{-42,+53}$ & 0.302 & 23.6 & 14.3 & 0.333 & 12.6\% & 3145.55 & 1.44 \\
TabDPT (T) & 1419${}_{-60,+61}$ & 0.365 & 24.0 & 8.5 & 1.0 & 12.5\% & 5467.20 & 39.74 \\
LightGBM (T+E) & 1417${}_{-34,+36}$ & 0.231 & 24.2 & 18.6 & 0.0 & 13.5\% & 413.11 & 2.38 \\
CatBoost (T+E) & 1401${}_{-42,+39}$ & 0.240 & 25.3 & 17.2 & 0.333 & 13.1\% & 1659.69 & 0.65 \\
ModernNCA (T+E) & 1393${}_{-53,+69}$ & 0.292 & 25.8 & 13.5 & 0.66 & 13.4\% & 4538.96 & 8.29 \\
CatBoost (T) & 1389${}_{-41,+40}$ & 0.221 & 26.1 & 17.5 & 0.333 & 13.3\% & 1659.69 & 0.07 \\
PrepLightGBM (D) & 1375${}_{-56,+61}$ & 0.280 & 27.1 & 11.3 & 1.0 & 12.4\% & 20.61 & 1.13 \\
LightGBM (T) & 1372${}_{-31,+38}$ & 0.169 & 27.3 & 20.8 & 0.0 & 14.1\% & 413.11 & 0.44 \\
TabM (T) & 1371${}_{-47,+57}$ & 0.249 & 27.3 & 18.4 & 0.0 & 13.4\% & 3145.55 & 0.15 \\
PrepTabM (D) & 1363${}_{-54,+50}$ & 0.264 & 27.9 & 14.4 & 0.66 & 13.7\% & 101.94 & 4.23 \\
XGBoost (T+E) & 1360${}_{-34,+33}$ & 0.171 & 28.1 & 21.0 & 0.0 & 14.2\% & 687.26 & 1.43 \\
CatBoost (D) & 1360${}_{-43,+35}$ & 0.188 & 28.1 & 17.3 & 0.333 & 13.9\% & 6.41 & 0.10 \\
XGBoost (T) & 1341${}_{-29,+27}$ & 0.134 & 29.4 & 25.2 & 0.0 & 14.5\% & 687.26 & 0.32 \\
xRFM (T+E) & 1341${}_{-42,+51}$ & 0.201 & 29.4 & 19.0 & 0.333 & 14.6\% & 847.29 & 2.12 \\
ModernNCA (T) & 1337${}_{-42,+43}$ & 0.163 & 29.7 & 16.0 & 0.66 & 14.1\% & 4538.96 & 0.44 \\
TabDPT (D) & 1317${}_{-65,+66}$ & 0.267 & 31.1 & 10.3 & 1.0 & 15.1\% & 48.57 & 39.36 \\
xRFM (T) & 1317${}_{-43,+47}$ & 0.169 & 31.2 & 19.3 & 0.333 & 15.4\% & 847.29 & 0.11 \\
PrepLinear (T+E) & 1285${}_{-76,+56}$ & 0.218 & 33.4 & 14.6 & 0.66 & 17.7\% & 853.05 & 22.15 \\
TabM (D) & 1282${}_{-47,+51}$ & 0.147 & 33.6 & 21.0 & 0.333 & 15.5\% & 11.16 & 0.13 \\
TorchMLP (T+E) & 1282${}_{-54,+41}$ & 0.127 & 33.6 & 24.1 & 0.0 & 15.3\% & 2867.82 & 1.86 \\
SAP-RPT-OSS (D) & 1282${}_{-56,+60}$ & 0.201 & 33.6 & 12.9 & 1.333 & 16.7\% & 14.34 & 2.08 \\
EBM (T+E) & 1260${}_{-43,+38}$ & 0.086 & 35.2 & 25.8 & 0.0 & 17.3\% & 3002.79 & 0.42 \\
ModernNCA (D) & 1248${}_{-42,+49}$ & 0.094 & 36.0 & 19.3 & 0.66 & 17.5\% & 14.75 & 0.31 \\
RealMLP (D) & 1245${}_{-44,+38}$ & 0.091 & 36.2 & 26.1 & 0.0 & 16.5\% & 10.69 & 1.78 \\
EBM (T) & 1217${}_{-52,+43}$ & 0.060 & 38.1 & 25.7 & 0.333 & 18.0\% & 3002.79 & 0.06 \\
XGBoost (D) & 1213${}_{-42,+48}$ & 0.060 & 38.4 & 21.2 & 0.66 & 17.2\% & 1.99 & 0.15 \\
TorchMLP (T) & 1204${}_{-51,+43}$ & 0.062 & 39.0 & 31.5 & 0.0 & 17.1\% & 2867.82 & 0.12 \\
PrepLinear (T) & 1201${}_{-77,+63}$ & 0.132 & 39.2 & 25.3 & 0.0 & 20.2\% & 853.05 & 1.88 \\
EBM (D) & 1189${}_{-50,+46}$ & 0.057 & 40.0 & 27.2 & 0.333 & 19.0\% & 7.35 & 0.05 \\
LightGBM (D) & 1174${}_{-38,+44}$ & 0.036 & 41.0 & 36.2 & 0.0 & 17.8\% & 2.21 & 0.17 \\
RandomForest (T+E) & 1158${}_{-47,+52}$ & 0.051 & 42.0 & 30.4 & 0.0 & 19.6\% & 377.13 & 0.76 \\
RandomForest (T) & 1121${}_{-45,+47}$ & 0.026 & 44.3 & 31.6 & 0.166 & 20.4\% & 377.13 & 0.08 \\
PrepLinear (D) & 1105${}_{-80,+66}$ & 0.067 & 45.3 & 24.5 & 0.333 & 23.4\% & 4.59 & 1.70 \\
TorchMLP (D) & 1063${}_{-56,+41}$ & 0.015 & 47.7 & 44.3 & 0.0 & 21.4\% & 10.93 & 0.13 \\
xRFM (D) & 1048${}_{-69,+67}$ & 0.030 & 48.5 & 39.0 & 0.0 & 24.7\% & 3.21 & 0.69 \\
RandomForest (D) & 1000${}_{-51,+47}$ & 0.000 & 50.9 & 44.6 & 0.0 & 24.7\% & 0.45 & 0.05 \\
Linear (T+E) & 937${}_{-104,+68}$ & 0.024 & 53.7 & 30.8 & 0.66 & 32.8\% & 237.45 & 0.43 \\
Linear (T) & 914${}_{-112,+73}$ & 0.020 & 54.6 & 38.1 & 0.0 & 33.4\% & 237.45 & 0.07 \\
Linear (D) & 831${}_{-127,+76}$ & 0.002 & 57.4 & 53.7 & 0.0 & 35.9\% & 1.19 & 0.12 \\
\bottomrule
\end{tabular}
}
\end{table}
\section{Additional Details on TabPrep and Feature Generators} \label{appendix:fe-method-details}
This section provides implementation details for the TabPrep generators. \autoref{fig:system} summarizes the structural patterns and preprocessing methods included in TabPrep. 
The goal of each generator is to produce a bounded set of plausible candidates with low generation and selection overhead. Unlike other autoFE approaches, we do not attempt to exhaustively find optimal operations/transformations, which leads to efficiency and seemless integration into benchmarking workflows and evaluation pipelines in general. 
Accordingly, many design choices below prioritize low, predictable runtime, leakage-safe integration into cross-validation setups, and robustness across heterogeneous tabular datasets.

\subsection{Ordered Arithmetic Feature Expansion}
To fill the gap of a dedicated generator for arithmetic structure missing from existing preprocessing and feature engineering libraries, we propose Ordered Arithmetic Feature Expansion (OAFE). The algorithm relies on lightweight base feature filtering and an ordered informed generation process.

\begin{table}[t]
    \caption{Default hyperparameters for Ordered Arithmetic Feature Expansion.}
    \label{tab:space:arithmetic}
    \centering
    \begin{tabular}{ll}
    \toprule
    Hyperparameter & Default \\
    \midrule
    \multicolumn{2}{c}{\textbf{--------------Generation control--------------}} \\
    interaction\_types & \{/, *, -, +\} \\
    max\_order ($R$) & 3 \\
    feature\_generation\_seed & 42 \\
    cat\_as\_num & False \\
    \multicolumn{2}{c}{\textbf{--------------Efficiency control--------------}} \\
    max\_feats ($M$) & 2000 \\
    max\_base\_feats & 150 \\
    subsample & 100,000 \\
    \multicolumn{2}{c}{\textbf{--------------Lightweight filtering--------------}} \\
    min\_cardinality ($\tau_{\mathrm{card}}$) & 3 \\
    nan\_threshold ($\tau_{miss}$) & 0.95 \\
    mode\_imbalance\_threshold ($\tau_{\mathrm{mode}}$) & 0.9 \\
    \bottomrule
    \end{tabular}
\end{table}

\subsubsection{Lightweight feature filtering.}
Before generating new arithmetic features, we apply a lightweight filtering step to remove features that provide little statistical information or may lead to unstable transformations.
\begin{enumerate}[topsep=2pt,itemsep=2pt,parsep=0pt]

\item Features with very low cardinality are filtered. The default pipeline removes binary features and maintains features with a cardinality $\tau_{\mathrm{card}}>=3$.
\item We remove features that show a missing-rate higher than a predefined threshold. The default pipeline removes features where more than $95\%$ of the values are missing ($\tau_{miss}>0.95$).
\item To improve the likelihood of using informative base features, we remove features whose empirical distributions are highly concentrated on a single value.
Let $\mathcal{V}_j$ denote the set of unique observed values of a feature $\mathbf{x}_j$ and define the mode frequency
\[
f_j
=
\frac{1}{n}
\max_{v \in \mathcal{V}_j}
\left|
\left\{ i \;:\; x_{ij} = v \right\}
\right|.
\]
Features are retained only if $f_j < \tau_{\mathrm{mode}},$
where $\tau_{\mathrm{mode}}$ close to $1$ removes nearly constant features that would produce arithmetic combinations with low additional signal.
The default pipeline removes all features where $95\%$ of the values correspond to the mode ($\tau_{mode}>0.95$).
\item Whenever a dataset contains more than the defined maximum number of base features to use ($\mathrm{max\_base\_feats}$), we sort the features by mode frequency and use the features with the lowest mode frequency. 
\item To ensure efficiency for large datasets, the data is randomly subsampled prior to computing the statistics used for filtering. 
\end{enumerate}
Note that the algorithm identifies numerical features by pandas data type, and optionally can add categorical features converted to numerical as base features. The latter is implemented and mentioned for completeness, but covers corner cases that are not evaluated in this paper.
After applying these filters, the remaining numerical feature matrix
\[
\tilde{\mathbf{X}}_{\mathrm{num}}
\in
\mathbb{R}^{n \times \tilde d_{\mathrm{num}}}
\]
contains the subset of numerical features suitable for arithmetic feature generation and serves as the input to the Ordered Arithmetic Feature Expansion procedure.

\subsubsection{Ordered Arithmetic Feature Generation Procedure}
\textbf{Random Interaction Selection.} \hspace{1.5mm}
We rely on ordered generation of arithmetic features focusing on efficient, budgeted generation, without exhaustive selection using the target variable.  
Instead of enumerating the full combinatorial space of arithmetic expressions, the algorithm samples candidate feature combinations while respecting a global feature budget $M$.  
Feature generation proceeds \emph{in an ordered manner}, first by increasing interaction order and, within each order, by a fixed operator sequence.

Given the filtered numerical feature matrix $\tilde{\mathbf{X}}_{\mathrm{num}}$ with base features 
\[
\mathcal{B} = \{x_1,\dots,x_p\}, \qquad p = |\mathcal{B}| <= \mathrm{max\_base\_feats},
\]
arithmetic expressions are generated using the ordered operator set
\[
\mathcal{O} = (/, \times, -, +),
\]
subject to a global limit of at most $M$ generated features and a maximum interaction order $R$.

\textbf{Ordered Expansion Procedure.} \hspace{1.5mm}
Feature generation proceeds sequentially across interaction orders.  
Let $\mathcal{E}_r$ denote the set of generated expressions of order $r$.

\begin{enumerate}[topsep=2pt,itemsep=2pt,parsep=0pt]

\item \textbf{Second-order interactions.}  
Pairs of base features are randomly sampled from the set
\[
\mathcal{P}_2 = \{(x_i,x_j) \mid 1 \le i < j \le p\}.
\]
For each sampled pair $(x_i,x_j)$, arithmetic expressions are created by applying the operators in the fixed sequence $\mathcal{O}$.
This results in the ordered candidate expressions
\[
x_i/x_j,\quad x_i \times x_j,\quad x_i - x_j,\quad x_i + x_j,\quad x_j/x_i,.
\]
Division is directional and therefore evaluated for both operand orders, while the $j/i$ division has the lowest priority.  
Expressions are appended to the candidate set by iterating over $i/j$ divisions for the sampled pairs. We then proceed sequentially with the other operators until all operators are computed or the global feature budget $M$ is reached.

\item \textbf{Higher-order interactions.}  
For interaction orders $r = 3,\dots,R$, new expressions are constructed by extending previously generated expressions.  
Candidate pairs are sampled from
\[
\mathcal{P}_r = \mathcal{E}_{r-1} \times \mathcal{B}.
\]

For each sampled pair $(e, x_j)$ with $e \in \mathcal{E}_{r-1}$, the same ordered operator sequence is applied:
\[
e/x_j,\quad e \times x_j,\quad e - x_j,\quad e + x_j.
\]

\item \textbf{Redundancy control.} To reduce redundant expressions and improve information gain:
\begin{itemize}
\item Commutative operators ($+$ and $\times$) are canonicalized so that equivalent forms such as $x_i + x_j$ and $x_j + x_i$ appear only once.
\item When extending an expression $e$ with a base feature $x_j$, the expansion is allowed only if
\[
x_j \notin S(e),
\]
where $S(e)$ is the set of base features already used in $e$ (i.e., each base feature may appear at most once per expression).
\end{itemize}

\item \textbf{Budget control.}  
Feature generation continues sequentially across interaction orders and operators until either
\[
|\mathcal{E}_2 \cup \dots \cup \mathcal{E}_R| \ge M
\]
or the maximum order $R$ is reached.
If the budget is exceeded, the list of expressions is truncated to the first $M$ generated features. That means, for very high-dimensional datasets and limited $M$, the generation will be restricted to adding ratios of two features. For datasets with smaller feature counts, more complex interactions are possible.

\end{enumerate}

\subsubsection{Additional Design Decisions}
\textbf{Normalization.} \hspace{1.5mm} 
One design choice in learning arithmetic feature interactions is whether to normalize features prior to computing interactions. Although this could generally be a hyperparameter, we always keep the original feature space for these operations. We aim for a feature engineering approach that targets hidden data structure, not for an approach that targets feature interactions in general. Therefore, assuming that the data is given in a raw format, algebraic relations between features should exist in the original feature space.

\textbf{Univariate Operations.} \hspace{1.5mm}
Related autoFE libraries often include univariate algebraic operators (e.g., log, exp, power, ...) in addition to arithmetic operations. 
For example, openFE, which is based on LightGBM, includes univariate monotonic transformations such as logarithms in its candidate space, despite tree-based models like LightGBM being nearly invariant to monotonic transformations. 
As a result, such features would greatly increase the candidate pool, especially for subsequent 2-order interactions, with a lower chance of improving performance.
Therefore, we avoid monotonic operations because there is not enough evidence that these operations are blind spots for existing models. 

\textbf{Exploiting Randomness to Increase Diversity.} \hspace{1.5mm}
In order to explore different feature candidates under limited time budgets, we recommend to test the generator repeatedly with different feature generation seeds and operators. Due to the limited feature budget, the algorithm may miss helpful features. However, we expect users to treat feature generation as a hyperparameter, where a random seed controls which features are generated. Therefore, during hyperparameter optimization, varying feature sets can be explored. This mitigates the limitation of restricting to single feature sets in budgeted generation. While this may increase the diversity of different model configurations, we show in Appendix \ref{app:seed_sensitivity} that the sensitivity to the feature generation seed is low and different feature sets can improve performance.


\subsection{CatPrep Feature Generators}
Previous automated feature engineering libraries either omitted categorical feature handling (e.g., autofeat \cite{horn2019autofeat}), or allowed the creation of new categorical features (e.g., openFE \cite{zhang2023openfe}) which leads to high-cardinality categorical features and as a result to frequent inefficiencies and out-of-memory failures. 
Moreover, no single encoding strategy for categorical data is flexible enough to explicitly cover all three kinds of categorical feature effects that we explained in \autoref{ssec:catprep}.
Therefore, we propose a generator pipeline (\textbf{CatPrep}) of three complementary techniques, each targeted towards one type of group-conditional effects.

\begin{table}[t]
    \caption{Default hyperparameters for Out-of-fold target encoder.}
    \label{tab:space:oofte}
    \centering
    \begin{tabular}{ll}
    \toprule
    Hyperparameter & Default \\
    \midrule
    n\_splits & 5 \\
    $\alpha$ & 10.0 \\
    random\_state & 42 \\
    keep\_original & False \\
    \bottomrule
    \end{tabular}
\end{table}

\begin{table}[t]
    \caption{Default hyperparameters for categorical interaction generator.}
    \label{tab:space:catint}
    \centering
    \begin{tabular}{ll}
    \toprule
    Hyperparameter & Default \\
    \midrule
    max\_order $R$ & 3 \\
    max\_new\_feats $M$ & 100 \\
    min\_cardinality & 2 \\
    random\_state & 42 \\
    \bottomrule
    \end{tabular}
\end{table}

\subsubsection{Out-of-Fold Target Encoding (OOF-TE)}
\label{sec:oof_te_impl}
We implement a variant of target encoding for high-cardinality categorical variables, mapping each category to a regularized estimate of its conditional target expectation \cite{micci2001preprocessing}. We follow the observation of \citet{pargent2022regularized} that cross-validated target encodings are consistently strong baselines on high-cardinality categorical features. 
Note that out-of-fold target encoding is not a new technique and has already been used in several Kaggle competitions \footnote{e.g., \href{https://www.kaggle.com/competitions/porto-seguro-safe-driver-prediction/discussion/44987}{Porto Seguro’s Safe Driver Prediction Challenge}, \href{https://www.kaggle.com/competitions/bnp-paribas-cardif-claims-management/overview}{BNP Paribas Cardif Claims Management Challenge}}.
Let $c_k \in \mathcal{C}_k$ denote a categorical feature and $y$ the target. For regression, the encoding approximates $\mathbb{E}[y \mid c_k]$; for binary classification, we encode the conditional probability of the positive class; for multiclass classification with $K$ classes, we construct a one-vs-rest target matrix $Y \in \{0,1\}^{n \times K}$ and produce $K$ encoded features per categorical column. Note that while sklearn also implements cross-validated target encoding, we chose to use a custom implementation for alignment with the design of other feature generators and to flexibly add functionalities like dropping vs. keeping original features, filtering features for many-class targets, or filtering redundant encodings. For the implementation we follow the original work on target encoding \cite{micci2001preprocessing}.

\paragraph{Train-time (fit): Out-of-Fold Encodings.}
During fitting, we compute encodings via K-fold cross-validation. For classification tasks, we use stratified folds. For each fold $s \in \{1,\dots,S\}$ with training indices $\mathcal{I}^{(s)}_{\mathrm{tr}}$ and validation indices $\mathcal{I}^{(s)}_{\mathrm{val}}$, we compute category statistics \emph{only} on $\mathcal{I}^{(s)}_{\mathrm{tr}}$ and assign encodings to all $i \in \mathcal{I}^{(s)}_{\mathrm{val}}$. Thus, each training row receives an encoding that was computed without using its own target, preventing target leakage. The resulting out-of-fold encodings are used as representation of the training data, without keeping the original categorical features (default).

For a fixed categorical column and target representation $z_i$, we compute fold-specific per-category counts and target sums on $\mathcal{I}^{(s)}_{\mathrm{tr}}$:
\[
n_c^{(s)} \;=\; \sum_{i \in \mathcal{I}^{(s)}_{\mathrm{tr}}} \indic{c_i = c},
\qquad
t_c^{(s)} \;=\; \sum_{i \in \mathcal{I}^{(s)}_{\mathrm{tr}}} z_i \, \indic{c_i = c},
\]
yielding the empirical mean $\hat{\mu}_c^{(s)} = t_c^{(s)}/n_c^{(s)}$ when $n_c^{(s)} > 0$. Here, $z_i=y_i$ for regression and binary classification, while for multiclass classification the same construction is applied to each one-vs-rest target column. We apply shrinkage with $\alpha > 0$:
\[
\mathrm{TE}^{(s)}(c)
\;=\;
\frac{n_c^{(s)} \, \hat{\mu}_c^{(s)} + \alpha \, \mu^{(s)}_{\mathrm{prior}}}{n_c^{(s)} + \alpha},
\]
where $\mu^{(s)}_{\mathrm{prior}}$ is a fold-specific prior computed as the mean of per-category means over categories observed in $\mathcal{I}^{(s)}_{\mathrm{tr}}$. Categories not observed in the fold training split, as well as missing values, are assigned $\mu^{(s)}_{\mathrm{prior}}$.

\paragraph{Test-time (transform): Full-training Encodings.}
In addition to the out-of-fold encodings, we compute and store a \emph{single} encoding table on the \emph{full} training data, using the same smoothing rule, together with a global prior. At transformation time on unseen data, each category value is mapped using this full-training encoding table. Values that are missing or not present in the stored category vocabulary (unseen categories) fall back to the stored prior. 

\subsubsection{Categorical Interaction Feature Generator}
\label{sec:cat_interactions}
Given categorical features $\mathbf{X}_{\mathrm{cat}}=[\mathbf{c}_1,\dots,\mathbf{c}_{d_{\mathrm{cat}}}]$, a feature budget $M$, and a $\mathrm{max\_order}$ $R$, we generate interaction features by combining categorical columns via cartesian products. For an interaction order $r\in\{2,\dots,R\}$, a selected tuple of columns $(j_1,\dots,j_r)$ defines a new categorical feature
\[
\mathbf{c}_{j_1,\dots,j_r}
\;:=\;
\big(\mathbf{c}_{j_1},\dots,\mathbf{c}_{j_r}\big),
\]
where each value corresponds to the joint category assignment across the involved columns. Thus, each unique combination of base-category values forms one new category. \\
To control complexity, we generate at most $M$ interaction features in total. For each order $r=2,\dots,R$, we randomly sample distinct categorical column tuples of length $r$ until the budget is exhausted. Furthermore, optionally, low-cardinality categorical features can be filtered. Each generated interaction feature is encoded using the out-of-fold target encoding procedure described in Section~\ref{sec:oof_te_impl}. The final interaction representation is obtained by concatenating the encoded interaction features across all sampled orders.\\
At inference time, interaction levels are restricted to combinations observed during training. Previously unseen combinations, as well as missing values, are mapped to a dedicated \textsc{unknown} level before applying the stored target-encoding statistics.

\subsubsection{Relative GroupBy Feature Generator}
\label{sec:groupby-preprocessor}
This preprocessor constructs features that describe each numerical value relative to a category-specific reference distribution. For completeness, we describe the full functionality of the implemented preprocessor, including standard GroupBy operations, although the proposed TabPrep default pipeline relies solely on relative GroupBy operations. Given a dataset $\mathcal{D}=(\mathbf{X},\mathbf{y})$,  The preprocessor proceeds in three steps: (1) feature selection, (2) budgeted pair selection, and (3) feature construction.

\begin{table}[t]
    \caption{Default hyperparameters for relative groupby interaction generator.}
    \label{tab:space:groupby}
    \centering
    \begin{tabular}{ll}
    \toprule
    Hyperparameter & Default \\
    \midrule
    \multicolumn{2}{c}{\textbf{------------------------------Generation control------------------------------}} \\
    aggregations ($\mathcal{A}_{\mathrm{group}}$) & \{\texttt{mean}, \texttt{pct\_rank}\} \\
    relative\_to\_aggs & \{\texttt{mean}\} \\
    relative\_ops & \{\texttt{ratio}\} \\
    drop\_non\_relative & True \\
    \multicolumn{2}{c}{\textbf{--------------------Efficiency and lightweight filtering--------------------}} \\
    max\_features ($M$) & 500 \\
    Num. as cat. cardinality threshold ($\tau_{\mathrm{as\_cat\_card}}$) & 2 \\
    Num. cardinality threshold ($\tau_{\mathrm{num\_card}}$) & 10 \\
    Min. cat. support ($k_{\min}$) & 20 \\
    \bottomrule
    \end{tabular}
\end{table}

\noindent\textbf{Stage 1: Feature selection}
\begin{enumerate}[topsep=2pt,itemsep=2pt,parsep=0pt]
    \item Identify categorical features $\mathcal{C}$ as object- or category-typed columns.
    \item Identify numerical features $\mathcal{N}$ as the remaining numeric columns with sufficiently many distinct values (controlled by $\tau_{\mathrm{num\_card}}$). To reduce the number of numerical columns with low variation for meaningful within-group comparisons, by default, we use $\tau_{\mathrm{num\_card}}=10$. 
    \item Optionally reinterpret low-cardinality numeric columns as categorical (controlled by $\tau_{\mathrm{as\_cat\_card}}$). By default, the TabPrep pipeline does not cast low-cardinality numerical columns to categorical ($\tau_{\mathrm{as\_cat\_card}}=2$) to keep the pipeline simple.
\end{enumerate}

\vspace{4pt}
\noindent\textbf{Stage 2: Budgeted categorical--numerical pairing}
To keep the transformation computationally bounded, pairs are selected in a deterministic order under a feature budget $M$. The ordering of pairs is designed to prioritize categorical features that admit reliable group-level estimates and numeric features that contain high variation.
\begin{enumerate}[topsep=2pt,itemsep=2pt,parsep=0pt]
    \item \textbf{Categorical feature order.} We prioritize categorical features whose categories are sufficiently supported by observations, ensuring stable group statistics, meaning that the lower frequency tail of observations is not too high. 
    For each categorical feature $\mathbf{c}_k \in \mathcal{C}$, we compute the frequency of each level in the training data (ignoring missing values). We aim to prevent features from being discarded due to single or few rare categories. Therefore, we consider the $10$ smallest category frequencies above a minimum support threshold of $k_{\min}=20$. These vectors are compared lexicographically in descending order to rank features, favoring features whose small category groups still contain many observations. 
    Hence, we prioritize features where even the small groups are well supported by counts being not too small to estimate group statistics reliably ($\ge k_{\min}$), thus stressing robustness in the tail of the group-size distribution. 
    \item \textbf{Numerical feature order.} Numerical features $x_j$ are ranked by descending cardinality, prioritizing features with more distinct values, and, hence more potential for informative within-group variation.
    \item \textbf{Generation order.} Finally, the method iterates over categorical features in their priority order, and within each categorical feature, over numerical features in their priority order, adding pairs until the budget constraint $M$ is met.  Each selected pair produces a fixed number of derived features (depending on the chosen aggregations and whether relative features are enabled). New features are added per pair until the feature budget $M$ is reached.
\end{enumerate}

\vspace{4pt}
\noindent\textbf{Stage 3: Generated group-conditional features} 
\begin{enumerate}[topsep=2pt,itemsep=2pt,parsep=0pt]
    \item \textbf{Group reference statistics.}
    For each selected pair $(\mathbf{c}_k,\mathbf{x}_j)\in\mathcal{P}$ and each aggregation operator in a configurable set of aggregations
    $\mathcal{A}_{\mathrm{group}} \in \{\mathrm{mean}, \mathrm{std}, \mathrm{median}, \mathrm{count}, \mathrm{n\_unique}, \mathrm{min}, \mathrm{max}, \mathrm{quantile}_1,\ldots,\mathrm{quantile}_{99}\}$,
    we generate one feature that maps each row to the statistic of its category level. Concretely, all rows that share the same value of $\mathbf{c}_k$ receive the same reference statistic for $\mathbf{x}_j$ (unseen levels at inference are treated as missing). The resulting features capture \emph{between-group variation}.
    
    \item \textbf{Within-group normalized rank.}
    If enabled, $\mathcal{A}_{\mathrm{group}}$ can also contain $\mathrm{pct\_rank}$ to generate a feature that encodes the relative position of $x_{i,j}$ \emph{within} its category group defined by $\mathbf{c}_k$.
    This feature acts as a percentile rank: it is close to $0$ for values near the lower tail of the group's empirical distribution and close to $1$ for values near the upper tail. For unseen groups or missing values, it defaults to $0.5$.

    \item \textbf{Relative intra-group transformations.}
    To emphasize category-conditional shifts in numerical feature behavior, we generate features that re-express $x_{i,j}$ relative to a group-specific reference level from $\mathcal{A}_{\mathrm{group}}$. The parameter $\mathrm{relative\_to\_aggs}$ specifies which group aggregations from $\mathcal{A}_{\mathrm{group}}$ to use, and the parameter $\mathrm{relative\_ops}$ specifies which relative operators to use. We use two variants to express features relative to reference statistics:
    (i) a \emph{group-centered} value (difference to the reference) and
    (ii) a \emph{group-normalized} value (ratio to the reference). By default, the group reference statistics are only used to compute relative features and are not returned ($\mathrm{drop\_non\_relative=True}$). These features, as well as the normalized rank, capture \textit{within-group variation}. 
\end{enumerate}

\subsection{Random Subset Feature Compression Generator}
\begin{table}[ht]
    \caption{Default hyperparameters for random subset feature compression generator.}
    \label{tab:space:rsfc}
    \centering
    \begin{tabular}{ll}
    \toprule
    Hyperparameter & Default \\
    \midrule
    n\_subsets & 50 \\
    subset\_size & None \\
    min\_subset\_size & 2 \\
    max\_subset\_size & None \\
    max\_base\_feats & 150 \\
    feature\_generation\_seed & 42 \\
    max\_cardinality & None \\
    round\_numerical & 2 \\


    \bottomrule
    \end{tabular}
\end{table}

We propose Random Subset Feature Compression (RSFC), a novel generator targeting the improvement of a model's capabilities to learn
\textbf{(1)} pseudo-categorical interactions of numerical features and
\textbf{(2)} higher-order value-based interactions of low-cardinality features in general.
RSFC generates sets of continuous features that summarize target variation in modal regions defined by subsets of features. 
It does so by turning randomly chosen feature subsets into discrete cluster keys and applying
out-of-fold target encoding (OOF-TE) to these keys, thereby approximating latent category-conditional target
effects while avoiding target leakage. 

\textbf{Motivation.} \hspace{1.5mm}
Our motivation to develop the RSFC generator comes from a chain of observations made by analyzing the difference between numerical and categorical features:
\begin{enumerate}[topsep=2pt,itemsep=2pt,parsep=0pt]
\item The most characteristic property of categorical features is that single feature values influence the target and interact with other features, while for numerical features, smooth interpolation between feature values is necessary.
\item If categorical features are ordinal-encoded, they induce a high-frequency dependence of $y$, which is a known challenge for many models \cite{grinsztajn2022tree}. Consequently, numerical features that induce high-frequency target dependence effectively behave like categorical features from a model’s perspective, because local smoothness and interpolation across nearby values are not meaningful.
\item Current models heavily rely on interpolation between numerical feature values. Therefore, if numeric features behave like categorical features in the sense that assigning effects to single values is more important than interpolation, models have troubles spotting that and treating a feature as categorical can be beneficial.
\item Although for a low-cardinality numeric feature, this is not an issue, it becomes a serious issue with increasing cardinality. E.g., tree-based models require a high number of splits to treat each value individually. Similarly, neural models require high capacity to isolate distinct feature value effects since models rely on dense layers biased towards smoothing the values.
\item These observations can be extended to modal regions spanned by multiple features. Similarly, as relevant single values, clusters of samples having the same value on multiple features can show distinct, generalizable target effects. The more features are involved in such an effect, the harder such multidimensional, pseudo-categorical patterns are to learn. Furthermore, pseudo-categorical numerical features may interact with other pseudo-categorical numerical, but also with categorical features. 
\end{enumerate}
Based on these insights, we conclude that 
\textbf{(1)} Pseudo-categorical numerical value effects require special treatment.
\textbf{(2)} It is important to tackle pseudo-categorical numerical features at the interaction level, 
\textbf{(3)} Since we never know whether such patterns are even present in an unseen dataset, it is crucial to preserve the default behavior of models to interpolate, while being conservative with allowing value interactions. Like with other generators, RSFC applies prefiltering of features followed by ordered feature generation.


\textbf{Candidate Feature Space.} \hspace{1.5mm}
RSFC operates on feature values that can define repeated groups of samples.
Therefore, the generator first constructs a candidate feature set whose values are likely to induce sufficiently supported modal regions.
Categorical and integer features are used in their original discrete representation, while numerical features are optionally discretized by rounding them to a fixed number of decimal places.
Candidate selection is guided by the observation that RSFC is most useful when sampled feature subsets form groups with enough observations to estimate stable target behavior.
Consequently, RSFC prioritizes features whose values are likely to reoccur across samples.
The priority order is:
\begin{enumerate}[topsep=2pt,itemsep=2pt,parsep=0pt]
    \item categorical features, since they already define discrete groups and are especially informative when interacting with numerical features;
    \item binary numerical features, because they have high support per value and can participate in many stable higher-order interactions;
    \item integer-valued numerical features, ordered by cardinality, since low-cardinality integer features often behave similarly to ordinal or coded categorical variables;
    \item floating-point numerical features after rounding, since repeated values only become likely after discretization.
\end{enumerate}
If the number of available features exceeds the specified budget \texttt{max\_base\_feats\_to\_consider}, RSFC keeps only the highest-priority candidates.
This limits the interaction search space and stabilizes runtime, while preserving a preference for features that can plausibly define pseudo-categorical or low-cardinality interaction structure.
In addition, features with cardinality above a user-specified threshold can be discarded.
For numerical features, \texttt{round\_numerical} controls the rounding precision used before constructing interaction keys.
The default configuration rounds numerical features to two decimal places and limits the candidate set to at most $150$ features.

\textbf{Generation Procedure.}  \hspace{1.5mm}
After constructing the candidate feature set, RSFC generates compressed interaction features by repeatedly sampling feature subsets, converting them into categorical keys, and encoding these keys with OOF-TE. The generator proceeds as follows:
\begin{enumerate}[topsep=2pt,itemsep=2pt,parsep=0pt]
    \item \textbf{Determine the number of generated subsets.}
    RSFC generates up to \texttt{n\_subsets} compressed interaction features.
    Each generated feature compresses one sampled feature into a single candidate features.

    \item \textbf{Determine the subset size.}
    For each generated feature, the interaction order is either fixed through \texttt{subset\_size} or sampled from the interval defined by \texttt{min\_subset\_size} and \texttt{max\_subset\_size}.
    If \texttt{subset\_size} is not specified, RSFC samples the subset size from this interval.
    If \texttt{max\_subset\_size} is not specified, the maximum possible subset size is determined by the number of available candidate features. 

    \item \textbf{Sample candidate features.}
    Given the selected subset size $k$, RSFC samples a subset
    $\mathcal{F}^{(s)} = \{f_1,\dots,f_k\}$
    uniformly at random without replacement from the candidate feature set $\mathcal{F}$. Whenever possible, the first subset consists of all candidate features. By default, subsequent subsets are of randomly samples size.
    The sampling is controlled by $\mathrm{feature\_generation\_seed}$. Each resulting subset defines one value-based partition of the dataset.

    \item \textbf{Construct a categorical interaction key.}
    For each sample $i$, RSFC combines the values of the selected features row-wise into a categorical key
    \[
        z_i^{(s)} = \mathrm{key}(x_{i,f_1}, \dots, x_{i,f_k}) .
    \]
    Samples with identical joint values receive the same key and therefore belong to the same induced group.

    \item \textbf{Encode the key with OOF-TE.}
    The key column $z^{(s)}$ is transformed into continuous features using out-of-fold target encoding.
    For regression and binary classification, this produces one continuous feature per sampled subset.
    For multiclass classification, the encoding can be applied in a one-vs-rest manner and produces one feature per class. 
\end{enumerate}

The resulting features can be interpreted as target-informed summaries of value-based regions in the input space.
If a sampled subset captures a meaningful value-based interaction, the encoded feature exposes this structure explicitly to the downstream model.
If the subset is uninformative, the feature is expected to have limited predictive value and can be ignored by the model.
Thus, RSFC transforms the problem of learning potentially high-frequency, higher-order value interactions into the easier problem of selecting among a small number of compressed interaction features.

\textbf{Practical Considerations.} \hspace{1.5mm}
RSFC is deliberately designed as a conservative feature generator.
It does not replace numerical features by categorical versions and does not enumerate all possible discretized interactions.
Instead, it preserves the original model input and adds only a limited number of target-encoded summaries of randomly sampled value partitions.
This allows the downstream model to retain its usual interpolation behavior while also receiving explicit information about non-smooth or pseudo-categorical structure. \\
At the same time, RSFC can be more dataset-sensitive than the other TabPrep generators.
If the sampled partitions do not correspond to stable target structure, the generated target-informed features may add noise or increase the risk of overfitting.
For this reason, RSFC should be evaluated as part of a validation or hyperparameter selection procedure rather than applied unconditionally.
As with all target-informed transformations in TabPrep, the OOF-TE step must be nested within the training folds to avoid target leakage.

\section{Analysis of Model Inductive Biases under the Identified Data Structures} \label{appendix:ind_bias}
In this appendix section, we provide diagnostic evidence for the structural patterns targeted by TabPrep. We study settings in which the target function contains arithmetic structure, group-conditional categorical effects, or pseudo-categorical numerical interactions, and analyze whether common tabular model classes can learn these patterns from the raw representation alone. These experiments motivate the design of the TabPrep generators.

\subsection{Details on the datasets used for uncovering model inductive biases}
We use two types of datasets to study how different model classes behave under the structural patterns targeted by TabPrep. First, we rely on existing benchmark datasets with known or interpretable structural properties to study arithmetic structure and pseudo-categorical numerical behavior. Second, for group-conditional categorical structure, we generate synthetic data under controlled conditions, since it was not possible to identify real-world datasets where the corresponding effects can be isolated without substantial confounding.
For arithmetic structure, we use toy and diagnostic datasets from TabZilla \cite{mcelfresh2023neural}. For pseudo-categorical numerical behavior, we use TabZilla datasets with known leakage or low-cardinality value interactions that induce categorical-like behavior in numerical features. The datasets used in these experiments are:

\begin{enumerate}[topsep=2pt,itemsep=2pt,parsep=0pt]
    \item \textbf{balance-scale} \cite{shultz1994modeling}: A three-class classification dataset where the target is determined by comparing two products corresponding to $(x_1 x_2) > (x_3 x_4)$. Hence, the task can be perfectly approximated using arithmetic operations. OpenML-ID: \href{https://www.openml.org/search?type=data&status=active&id=11}{11}.

    \item \textbf{jungle\_chess} \cite{van2014endgame}: A game-derived dataset where the target can be approximated by deterministic rules. Such rule-based structure can often be represented through combinations and interactions of the input variables, making the dataset useful for diagnosing whether models benefit from explicit feature construction. OpenML-ID: \href{https://www.openml.org/search?type=data&status=active&id=41027}{41027}.

    \item \textbf{vehicle} \cite{siebert1987vehicle}: A dataset whose features are manually constructed numerical descriptors extracted from images. Since several features are themselves arithmetic summaries of lower-level measurements, additional arithmetic interactions between the available features can recover useful structure that is not directly represented in the provided feature set. OpenML-ID: \href{https://www.openml.org/search?type=data&status=active&id=54}{54}.


    \item \textbf{artificial-characters} \cite{botta1991learning}: A dataset with numerical features only. As reported by \citet{rubachev2024tabred}, the feature \texttt{V7} contains a leakage pattern where individual floating-point values are strongly associated with the target. Consequently, this numerical feature behaves similarly to a categorical feature, making the dataset suitable for studying pseudo-categorical numerical behavior. OpenML-ID: \href{https://www.openml.org/search?type=data&status=active&id=1459}{1459}.

    \item \textbf{ada\_agnostic}: A dataset containing many low-cardinality features. Such features can benefit from higher-order value combinations, making the dataset useful for evaluating whether random subset compression can expose interaction structure that is difficult to learn from the original representation alone. OpenML-ID: \href{https://www.openml.org/search?type=data&status=active&id=40993}{40993}.

\end{enumerate}

\subsubsection{Categorical Data Generation}
For group-conditional categorical structure, we use synthetic regression datasets. This allows us to isolate the specific mechanisms studied in Section~\ref{ssec:catprep} while controlling their relative importance.

\textbf{Overview.}  \hspace{1.5mm}
We generate synthetic regression datasets with $d_{\mathrm{num}}$ numerical features, $d_{\mathrm{cat}}$ categorical features, and a target constructed as the sum of four controlled mechanisms:
(1) a numerical base effect, 
(2) additive effects of individual categorical features, 
(3) additive effects of categorical feature combinations, and 
(4) category-conditional numerical effects based on within-group deviations. 

\textbf{Numerical Features.} \hspace{1.5mm}
For each observation $i=1,\dots,n$ and numerical feature $j=1,\dots,d_{\mathrm{num}}$, we sample
\[
x_{ij} \sim \mathcal{N}(0,1),
\qquad
\mathbf{x}_i = (x_{i1},\dots,x_{id_{\mathrm{num}}})^\top.
\]

\textbf{Categorical Features.} \hspace{1.5mm}

Each categorical feature $k=1,\dots,d_{\mathrm{cat}}$ takes values in a finite set $\mathcal{C}_k$ with cardinality $K_k = |\mathcal{C}_k|$, where $K_k \ge 2$. For each observation, we sample
\[
c_{ik} \sim \mathrm{Uniform}(\mathcal{C}_k),
\]
independently across features and observations.

\textbf{Numerical Base Effect.} \hspace{1.5mm}
The numerical signal is generated as
\[
y_i^{(\mathrm{num})} = f(\mathbf{x}_i) + \varepsilon_i^{(\mathrm{num})},
\]
where $f$ denotes a fixed prediction model applied to the numerical features and $\varepsilon_i^{(\mathrm{num})}$ is Gaussian noise. In our experiments, $f$ is a multilayer perceptron with ReLU activations. The resulting predictions are standardized to zero mean and unit variance before being added to the final target.

\textbf{Single-Category Additive Effects.} \hspace{1.5mm}
To generate direct categorical effects, each categorical level receives a level-specific offset. For each categorical feature $k=1,\dots,d_{\mathrm{cat}}$ and level $l \in \mathcal{C}_k$, we sample
\[
\alpha_{kl} \sim \mathcal{N}(0,\sigma_{\mathrm{offset}}^2).
\]
For observation $i$, the raw contribution of feature $k$ is
\[
u_{ik} = \alpha_{k,c_{ik}}.
\]
We standardize these effects feature-wise and define the total single-category contribution as
\[
y_i^{(\mathrm{single})}
=
\frac{w_{\mathrm{single}}}{d_{\mathrm{cat}}}
\sum_{k=1}^{d_{\mathrm{cat}}}
\left(
u_{ik} + \varepsilon_{ik}^{(\mathrm{single})}
\right),
\]
where the noise terms are Gaussian with variance proportional to the scale of the corresponding weighted component and the weight $w_{\mathrm{single}}$ controls the overall impact of direct categorical effects, while categorical features are weighted equally.

\textbf{Categorical Combination Effects.} \hspace{1.5mm}
To introduce interactions between categorical features, we randomly select a subset
\[
\mathcal{K} \subseteq \{1,\dots,d_{\mathrm{cat}}\},
\qquad
|\mathcal{K}| \ge 2.
\]
The selected features jointly define a latent categorical feature. For each observed joint category level $l_{\mathrm{cmb}}$, we sample a latent offset
\[
\beta_{l_{\mathrm{cmb}}}
\sim
\mathcal{N}(0,\sigma_{\mathrm{combo}}^2).
\]
For observation $i$, let $c_{i,\mathrm{cmb}}$ denote the realized joint category level across the features in $\mathcal{K}$. The raw combination effect is
\[
v_i^{\mathrm{raw}} = \beta_{c_{i,\mathrm{cmb}}}.
\]
After standardizing $v_i^{\mathrm{raw}}$ to obtain $v_i$, the corresponding target contribution is
\[
y_i^{(\mathrm{combo})}
=
w_{\mathrm{combo}} v_i
+
\varepsilon_i^{(\mathrm{combo})}.
\]



\textbf{Category-Conditional Within-Group Numerical Effects.} \hspace{1.5mm}
To generate category-conditional numerical effects, we randomly select one categorical feature $k$ and one numerical feature $j$. For each level $l \in \mathcal{C}_k$, we sample a group-specific mean
\[
\mu_l \sim \mathcal{N}(0,\sigma_\mu^2).
\]
We then overwrite the selected numerical feature as
\[
x_{ij} = \mu_{c_{ik}} + \epsilon_i,
\qquad
\epsilon_i \sim \mathcal{N}(0,\sigma_\epsilon^2).
\]
Thus, $x_{ij}$ contains between-group variation through $\mu_{c_{ik}}$ and within-group variation through $\epsilon_i$. The target contribution depends only on the within-group deviation $\epsilon_i$, which is generally not directly available to the model. We standardize this deviation as
\[
r_i = \frac{\epsilon_i}{\mathrm{sd}(\epsilon)}.
\]
The raw centered signal is then constructed as
\[
h_i^{\mathrm{raw}} = \tanh(1.5 r_i).
\]
After standardizing $h_i^{\mathrm{raw}}$ to obtain $h_i$, the centered contribution is
\[
y_i^{(\mathrm{catnum})}
=
w_{\mathrm{catnum}} h_i
+
\varepsilon_i^{(\mathrm{catnum})}.
\]
The resulting numerical feature that is available to the model then affects the target in two ways:
(1) through the global numerical model, possibly interacting with other numerical features, and (2)
through the nonlinearly transformed deviation from a category-dependent reference value.

\textbf{Final Target.} \hspace{1.5mm}
The final target is the sum of all weighted components:
\[
y_i
=
y_i^{(\mathrm{num})}
+
y_i^{(\mathrm{single})}
+
y_i^{(\mathrm{combo})}
+
y_i^{(\mathrm{catnum})}.
\]
This construction aligns with our general viewpoint of functional decomposition in tabular datasets and allows us to control the relative importance of numerical structure, additive categorical effects, categorical interactions, and categorical-numerical interactions in the form of category-conditional numerical effects with within-group variation. After generation, 25\% of each dataset is used as test data. The exact hyperparameters used for generating data in our experiments are in \autoref{tab:cat_data_generation_params}.

\begin{table}[ht]
\centering
\caption{Parameters of the synthetic categorical data generator used for the group-conditional structure experiments. }
\label{tab:cat_data_generation_params}
\resizebox{\textwidth}{!}{%
\begin{tabular}{
    p{0.2\textwidth}
    p{0.4\textwidth}
    p{0.13\textwidth}
    p{0.13\textwidth}
    p{0.13\textwidth}
}
\toprule
\textbf{Parameter} 
& \textbf{Description} 
& \textbf{Direct cat. effects}
& \textbf{Cross-cat. effects}
& \textbf{Cat.-cond. num. effects} \\
\midrule

$n$ 
& Number of observations used to generate the dataset 
& 1000
& 1000 
& 1000 \\

$d_{\mathrm{num}}$ 
& Number of numerical input features 
& 5 
& 5 
& 5 \\

$f$ 
& Numerical base function used to generate $y_i^{(\mathrm{num})}$ 
& 2-layer ReLU MLP  
& 2-layer ReLU MLP 
& 2-layer ReLU MLP \\

$d_{\mathrm{cat}}$ 
& Number of categorical input features 
& 3 
& 2 
& 1 \\

$K$ 
& List of cardinalities of categorical features $k$, i.e., $|\mathcal{C}_k|$ 
& $[200,200,200]$ 
& $[50,50]$
& $[100]$ \\

$[w_{\mathrm{num}}, w_{\mathrm{single}}$, $w_{\mathrm{combo}}$, $w_{\mathrm{catnum}}]$
& Weights controlling the relative importance of the numerical, single-category, categorical-combination, and category-conditional numerical target components
& $[0.5, 0.5, 0., 0.]$ 
& $[0.4, 0.2, 0.4, 0.]$ 
& $[0.4, 0.2, 0., 0.4]$  \\

$[\sigma^{(\mathrm{num})}, \sigma^{(\mathrm{single})}$, $\sigma^{(\mathrm{combo})}, \sigma^{(\mathrm{catnum})}]$
& Noise scales for the numerical, single-category, categorical-combination, and centered numerical target components
& $[0.1, 0.1, 0., 0.]$
& $[0.1, 0.1, 0.1, 0.]$
& $[0.2, 0.2, 0., 0.2]$ \\

\bottomrule
\end{tabular}%
}
\end{table}

\subsection{Inductive Bias Alignment for Arithmetic Structure} \label{appendix:ind_bias_arithmetic}
\begin{figure}[h]
    \centering
    \includegraphics[width=0.5\columnwidth]{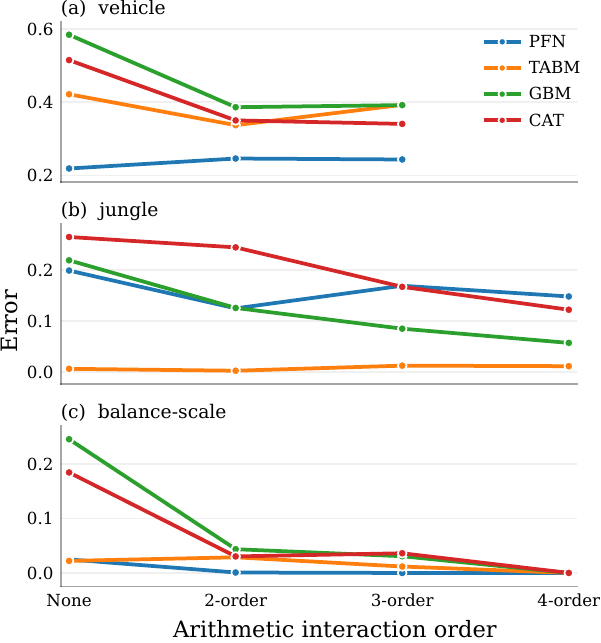}
    \caption{\textbf{Model comparison on data with algebraic structure.} Each plot shows model error, where lower is better. We show the error for the default model and for variants augmented with arithmetic interactions up to order $r$ using Ordered Arithmetic Feature Expansion with a budget of up to 2000 generated features. The displayed models are TabPFN-2.5 (PFN), TabM, LightGBM (GBM), and CatBoost (CAT). \texttt{jungle} represents a medium-sized task ($\approx 40{,}000$ samples), while \texttt{vehicle} and \texttt{balance-scale} are small tasks ($<1000$ samples).}
    \label{fig:arithmetic}
\end{figure}

\textbf{Motivation.} \hspace{1.5mm}
Our motivation to study arithmetic patterns stems from a closer look at the TabZilla-hard benchmark \cite{mcelfresh2023neural}. It contains several datasets whose predictive structure plausibly involves algebraic relationships: 
(1) tasks with deterministic or rule-based target functions, such as \texttt{balance-scale}, \texttt{jungle-chess}, \texttt{poker-hand}, and \texttt{monks-problems-2}; 
(2) datasets with image-derived features that are themselves constructed from lower-level measurements, such as \texttt{vehicle}, \texttt{mfeat-zernike}, and \texttt{mfeat-fourier}; and
(3) simulated datasets from domains governed by physical laws, such as \texttt{MiniBooNE} and \texttt{higgs}.
In such settings, target-relevant relationships may be more naturally expressed through arithmetic combinations of numerical features than through the raw feature representation alone. These datasets therefore provide useful diagnostic settings for testing whether common tabular model classes can recover algebraic structure without explicit feature construction.
We select three of these datasets for which the relevant structure is easy to interpret without additional domain knowledge and analyze model performance before and after augmenting the input with TabPrep's Ordered Arithmetic Feature Expansion generator.

\paragraph{All models can benefit from arithmetic features, especially tree-based models.}
\autoref{fig:arithmetic} shows that model classes differ substantially in their ability to exploit arithmetic feature combinations when these combinations are not explicitly represented. Tree-based models perform poorly relative to neural models on these tasks, which is consistent with their axis-aligned partitioning bias: smooth algebraic relationships generally require many splits to approximate when expressed only in the original coordinate system. Explicitly adding arithmetic features substantially reduces this mismatch, especially for LightGBM and CatBoost. The improvement of tree-based models also increases gradually as the maximum interaction order $r$ is raised. This supports the structural-component view of feature engineering: externalizing a larger set of arithmetic subcomponents progressively simplifies the learning problem, because the downstream model no longer has to approximate these patterns through axis-aligned partitions alone. \\
Interestingly, the benefit of arithmetic feature engineering is not limited to tree-based models. TabM and TabPFN-2.5 also improve on some datasets when low-order arithmetic interactions are added, indicating that even more flexible neural and foundation models do not always infer the relevant algebraic substructure reliably from limited data. At the same time, their gains are less monotonic: as the interaction order increases and more candidate features are added, their error can stagnate or even increase. This is consistent with the weaker feature-selection bias of neural models compared to tree-based models \cite{grinsztajn2022tree}. The results further illustrate the interaction between model class and data regime. TabPFN-2.5 performs best on the two small datasets, whereas TabM is strongest on the medium-sized \texttt{jungle} dataset, likely reflecting the stronger small-data bias of tabular foundation models. Overall, these results support the central premise of TabPrep: feature engineering can improve generalization by externalizing structural components of the target function, thereby reducing the burden on the downstream model to discover these components implicitly. The less aligned the inductive bias is with structural patterns, the higher the performance gains from externalizing the patterns via feature engineering.

\subsection{Inductive Bias Alignment for Group-Conditional Structure} \label{appendix:ind_bias_categorical}

\textbf{Motivation.} \hspace{1.5mm}
Our motivation to study group-conditional structure stems from two observations: 
\textbf{(1)} Many real-world tabular datasets contain high-cardinality categorical features and the long-standing competitiveness of CatBoost on tabular benchmarks is largely attributable to model-inherent feature engineering, that is absent from other models  \citet{tschalzev2024data}. This suggests that categorical variables often encode group-level effects that are difficult to recover from generic numerical encodings. 
\textbf{(2)} In traditional statistics, categorical variables are frequently treated as indicators of grouped or hierarchical data structure and mixed-effects models provide a classical framework for such settings by decomposing the target into population-level effects shared across all observations and group-specific deviations associated with categorical grouping variables \cite{bates2010lme4,zuur2009mixed, bolker2009generalized}. This motivates revisiting the perspective that a categorical feature is not merely a discrete input, but can indicate that target relationships change across groups, which directly aligns with our notion of group-conditional structure in tabular data. 

\textbf{LightGBM and TabPFN-2.5 lack a bias for high-cardinality categorical features.} \hspace{1.5mm}
\begin{figure}[!b]
    \centering
    \includegraphics[width=0.7\columnwidth]{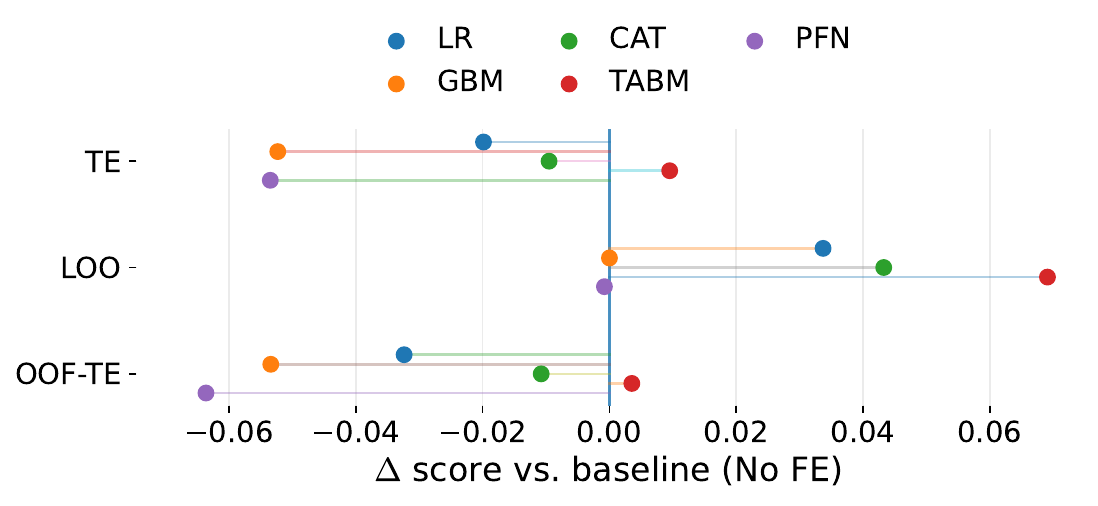}
   \caption{\textbf{Model comparison on data with cross-categorical target effects} (lower is better). We compare the models' default categorical feature handling (\textit{None}), out-of-fold target encoding (\textit{OOF-TE}), appending OOF-TE while keeping the original categorical features (\textit{OOF-TE\_APPEND}), adding the Cartesian product of the categorical features (\textit{CATINT}), and applying OOF-TE to categorical interactions (\textit{CATINT\_OOFTE}). The simulated dataset has 1000 samples and the regression target depends on $5$ numerical features. In addition, the target is shifted by a category-dependent constant defined by the combination of both categorical features. Performance is measured as MSE, with colors normalized per model. The displayed models are TabPFN-2.5 (PFN), TabM, LightGBM (GBM), CatBoost (CAT), and a linear model (LR).}
        
    \label{fig:single-highcard}
\end{figure}
\autoref{fig:single-highcard} shows that TabPFN-2.5 and LightGBM benefit substantially from replacing their default categorical treatment with target-based encodings. Among the studied encodings, OOF-TE is generally the most reliable, while leave-one-out encoding can even degrade performance. TabM, in contrast, benefits little from OOF-TE, suggesting that embeddings can be sufficient for direct categorical effects in this controlled setting. CatBoost also performs well without additional preprocessing, in line with \citet{tschalzev2024data}, since target statistics are already integrated into its categorical handling. Nevertheless, even CatBoost can still gain from explicit OOF-TE. Overall, these results indicate that direct high-cardinality categorical effects remain a weakness of TabPFN-2.5 and LightGBM, motivating the use of OOF-TE as a component of CatPrep. Since high-cardinality categorical features are largely absent from the synthetic priors used by recent tabular foundation models, this limitation may also affect other foundation models \cite{qu2025tabicl,zhang2025limix}.

\begin{figure}[t]
    \centering
    \includegraphics[width=0.7\columnwidth]{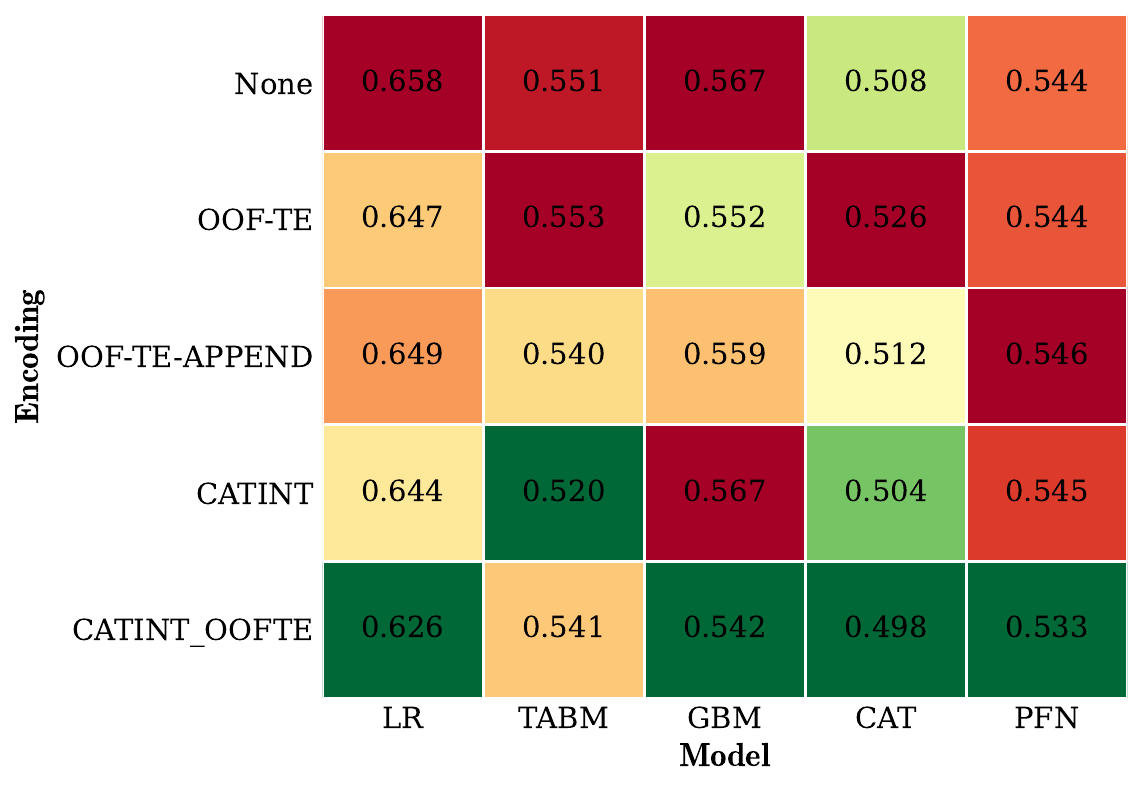}
    \caption{\textbf{Model comparison on data with group-conditional numerical feature effects} (lower is better). We compare the models' default categorical feature handling (\textit{No FE}), absolute GroupByThenMean interactions (\textit{GroupBy}), and relative GroupByThenMean interactions obtained by subtracting the group mean from the base numerical feature (\textit{Diff-GroupBy}). The simulated dataset has 1000 samples, and the regression target depends on $5$ numerical features and one categorical feature. For one numerical feature, the target-relevant signal is shifted by a category-dependent offset. Performance is measured as MSE, with colors normalized globally. The displayed models are TabPFN-2.5 (PFN), TabM, LightGBM (GBM), CatBoost (CAT), and a linear model (LR).}
    
    \label{fig:joint_cat}
\end{figure}

\textbf{Joint categorical effects are challenging for most models.} \hspace{1.5mm}
Since all studied tabular models rely on numerical encodings, while joint categorical effects exist in a symbolic space, learning such effects is challenging. \autoref{fig:joint_cat} shows that such interactions are hard to recover when only the individual categorical features are represented, either through compressed numerical encodings such as OOF-TE or through embeddings. Therefore, joint effects of categorical features with high-cardinality are difficult to approximate under typical inductive biases of weighted-sum-based or partition-based models.The only model that explicitly computes categorical combinations is CatBoost, which results in strong performance for tasks with joint categorical effects. Noteworthy, while TabM cannot learn such patterns on its own, it is the only model that performs well, even without the addition of OOF-TE, as seen in \autoref{fig:single-highcard}. On the other hand, LightGBM and TabPFN-2.5 cannot benefit from the addition of the combination feature, due to poor high-cardinality categorical feature handling, as discussed above.

\begin{figure}[!h]
    \centering
    \includegraphics[width=0.7\columnwidth]{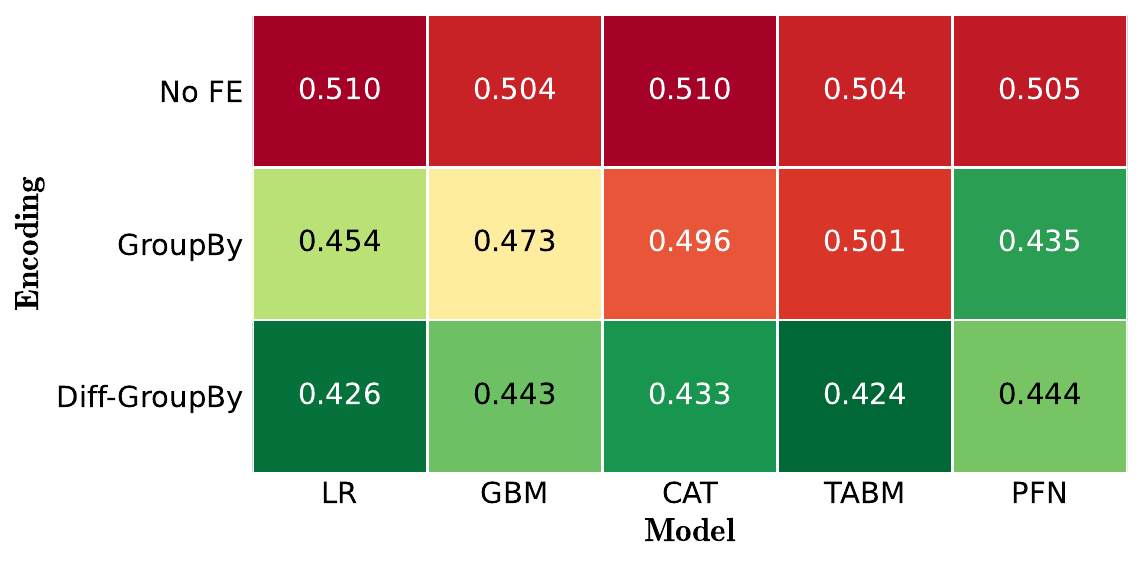}
    \caption{\textbf{Model comparison on data with group-conditional numerical feature effects} (lower is better). We compare the models default categorical feature handling (\textit{No FE}), absolute GroupByThenMean interactions (\textit{GroupBy}), and GroupByThenMean interactions subtracted from the base numerical feature (\textit{Diff-GroupBy}). The simulated dataset has 1000 samples and the regression target depends on $5$ numerical features and one categorical feature, but for one of the numeric features, the target effect is shifted by a category dependent offset (slope). The performance is measured as MSE, with the colors being normalized globally. TabPFN-2.5 (PFN), TabM, LightGBM (GBM), CatBoost (CAT), and a linear model (LR)}
    \label{fig:groupby}
\end{figure}

\textbf{No model is able to learn group-conditional numerical effects} \hspace{1.5mm}, as can be seen in \autoref{fig:groupby}. 
\autoref{fig:groupby} shows that none of the studied models reliably captures group-conditional numerical effects from the raw representation alone. Hence, none of the models has an inductive bias allowing to estimate within group shifts in a numerical feature induced by a categorical feature.
Remarkably, for TabPFN it is sufficient to add the GroupByThenMean feature. This can be explained by the fact that after adding this feature, the model only has to learn an arithmetic interaction between the original and the group-specific feature, which we have shown to be a strength of TabPFN-2.5 in the small data regime. This observation underscores our functional decomposition hypothesis, showing that focusing only on feature engineering covering particular weaknesses of a model can be sufficient.
\autoref{fig:groupby} shows that none of the studied models reliably captures group-conditional numerical effects from the raw representation alone. 

\subsection{Inductive Bias Alignment for Pseudo-Categorical Numerical Features} \label{appendix:ind_bias_pseudocat}

\begin{figure}[!h]
    \centering
    \includegraphics[width=\columnwidth]{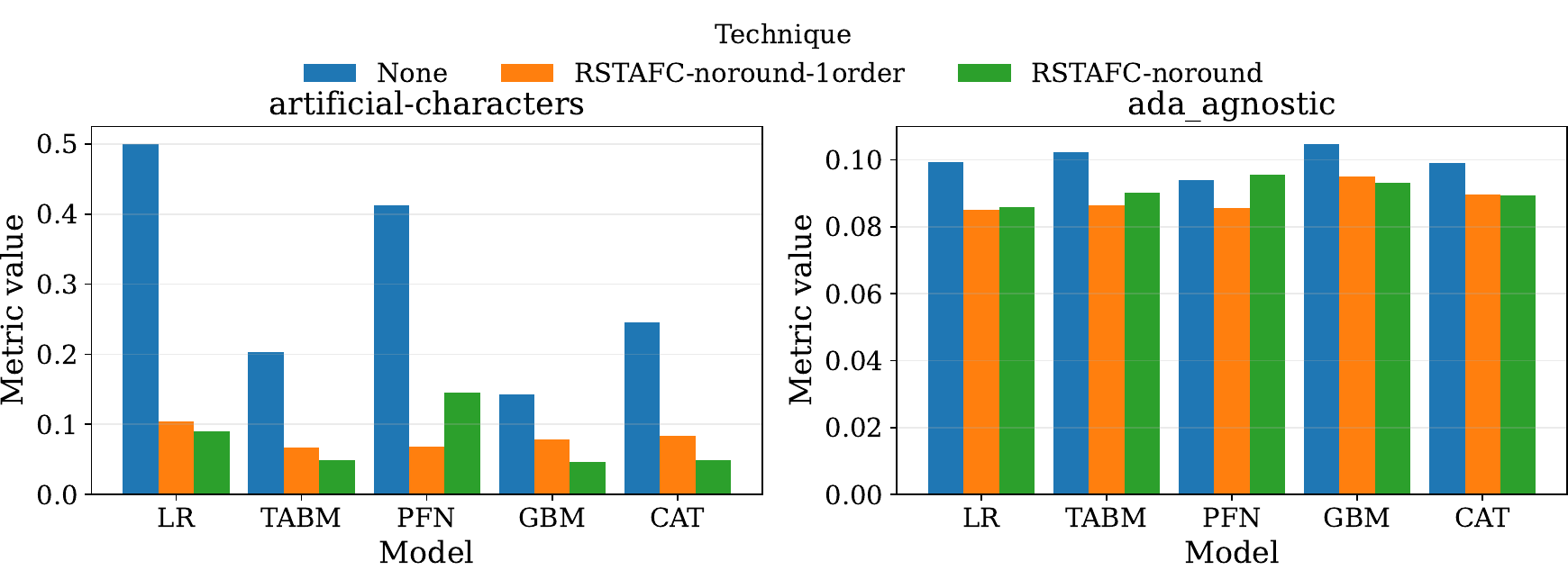}
    \caption{\textbf{Model comparison with Random Subset Feature Compression.} \texttt{artificial-characters} (left) is a dataset containing pseudo-categorical numerical features, while \texttt{ada\_agnostic} (right) contains many low-cardinality features. We evaluate performance using log loss, where lower is better. \textit{None} denotes the model without feature engineering. \textit{RSFC-noround-1-order} denotes Random Subset Feature Compression with $\mathrm{max\_order}=1$ and rounding of numerical features disabled, while \textit{RSFC-noround} allows higher-order interactions.}
    
    \label{fig:pseudocategorical}
\end{figure}

\textbf{All models can benefit from handling pseudo-categorical numerical features.} \hspace{1.5mm}
\autoref{fig:pseudocategorical} (left) shows that all studied models benefit from explicitly handling pseudo-categorical numerical features. Without such preprocessing, models have to learn target variation that depends on individual numerical values or small value regions, rather than on smooth changes in the numerical input space. This creates high-frequency patterns in the target function, for which neural models are known to have an unfavorable inductive bias, while tree-based models are comparatively better aligned \cite{grinsztajn2022tree}. Consistent with this view, LightGBM performs strongly in the default setting, whereas TabPFN-2.5 underperforms despite the small dataset size. This suggests that TabPFN-2.5 can inherit the same limitations as neural models when the relevant signal is value-wise or high-frequency rather than smooth.
Moreover, for most models, interaction-based RSFC variants outperform univariate encodings, supporting our hypothesis that pseudo-categorical numerical structure often appears through feature interactions rather than through isolated features alone. \autoref{fig:pseudocategorical} (right) further shows that, in the presence of many low-cardinality features, RSFC can benefit all studied models. This extends the insight from the analysis of joint categorical effects: when useful information is encoded in higher-order combinations of discrete or discretized values, explicitly compressing such combinations through target encoding can make the relevant structure easier to exploit.



\section{Extended Comparison of TabPrep vs. AutoFE Libraries}
\subsection{Budgeted Pattern-Informed Generation vs. Expand-and-Reduce} \label{appendix:expand_and_reduce}
In this subsection, we distinguish our approach from the dominant expand-and-reduce paradigm in automated feature engineering. In this paradigm, candidate features are first generated (often in large numbers) and then pruned through a subsequent selection procedure \cite{zhang2023openfe,kanter2015deep,shi2020safe,lam2021automated,horn2019autofeat,fan2010generalized,yuanfei2019autocross}. While conceptually appealing, this strategy is difficult to reconcile with large-scale, systematic benchmarking: generating and selecting from large candidate sets can be computationally expensive, hard to control in high-dimensional settings, dependent on specific model classes or pipeline designs, and susceptible to overfitting when selection is supervised.

We therefore adopt a different objective. Rather than aiming to identify the strongest possible dataset-specific feature set, we seek a simple, effective, and scalable feature-engineering baseline that can be integrated reliably into standard tuning pipelines across diverse datasets. This requires computational predictability, bounded feature growth, and robustness across heterogeneous datasets, feature types, and models. Motivated by these requirements, we introduce simplifying design principles that deliberately distinguish our framework from most existing autoFE libraries.

\begin{table}[h]
\centering
\small
\begin{tabular}{p{0.20\linewidth} p{0.33\linewidth} p{0.37\linewidth}}
\toprule
 & \textbf{Expand-and-reduce (OpenFE)} & \textbf{TabPrep} \\
\midrule
\textbf{Search unit}
& Concrete transformations of individual features or feature tuples
& Generator-level choices limited to generators for specific structural patterns \\[0.4em]
\midrule
\textbf{Feature generation mechanism}
& Exhaustively enumerates candidates before deciding which are useful
& Generates a bounded set of candidates using ordering rules or random sampling \\[0.4em]
\midrule
\textbf{Feature selection mechanism}
& Supervised model-based selection to prune the generated candidate pool
& Delegates usefulness assessment to the downstream learner and HPO procedure \\[0.4em]
\midrule
\textbf{Compute control}
& Memory and runtime depend on the dataset and size of the generated pool and selection model
& Per-generator feature budgets bound dimensionality before generation starts \\[0.4em]
\midrule
\textbf{Operator scope}
& Broad operation catalogs relying on selection to remove weak variants
& Includes only operators motivated by identified model blind spots and structural patterns \\[0.4em]
\midrule
\textbf{Operator representation}
& May create raw categorical features or many similar numerical variants
& Uses numerical encodings for generated categorical features and keeps only expressive variants whenever possible \\[0.4em]
\midrule
\textbf{Pipeline role}
& Usually acts as a standalone autoFE stage before model training
& Acts as a tunable preprocessing component inside the evaluation pipeline \\[0.4em]
\bottomrule
\end{tabular}
\caption{Comparison of TabPrep and the expand-and-reduce framework openFE \cite{zhang2023openfe}. We use openFE \cite{zhang2023openfe} as a prominent representative of expand-and-reduce approaches. TabPrep reformulates automated feature engineering for benchmarking by replacing expand-and-reduce with budgeted pattern-informed generation.}
\label{tab:budgeted_generation_shift}
\end{table}

\textbf{Search unit.} \hspace{1.5mm}
Expand-and-reduce methods typically search over concrete feature transformations, such as applying a specific operation to a specific feature or feature tuple.
This leads to a search space whose size grows rapidly with the number of input features and the number of supported operations.
TabPrep instead searches at the level of generators.
Each generator represents a family of related transformations designed to expose one recurring structural pattern, such as arithmetic structure, group-conditional categorical effects, or pseudo-categorical value interactions.
This distinction is important because different transformation families target different structural mismatches between the data-generating process and model inductive biases.
Treating them as one homogeneous candidate pool obscures these differences and leads to unnecessarily broad search spaces.
TabPrep therefore changes the search problem from selecting among many individual operations to deciding which pattern-targeted generators are useful for a dataset.

\textbf{Feature generation mechanism.} \hspace{1.5mm}
In expand-and-reduce pipelines, the expansion step typically creates a broad candidate pool before the method determines which generated features are useful.
This makes the generation phase expensive, because many features must be materialized even if most of them are later discarded.
TabPrep avoids exhaustive enumeration.
Instead, each generator produces only a bounded set of candidates using operation-specific ordering rules, lightweight prefiltering, or random sampling.
The goal is to increase the chance of generating useful features without attempting to cover the full combinatorial space.

\textbf{Feature selection mechanism.} \hspace{1.5mm}
Expand-and-reduce methods rely on supervised model-based feature selection to prune the generated candidate pool.
While this can identify predictive generated features, it introduces an additional supervised learning problem before the downstream model is trained.
This increases computational cost, creates dependence on the chosen selection model, and can increase the risk of overfitting during preprocessing.
TabPrep does not include a separate supervised reduction stage.
Instead, generated features are passed to the downstream learner, and their usefulness is assessed through the same model training and hyperparameter optimization procedure used in a given benchmark design.
Thus, feature selection is not performed as an additional preprocessing step, but absorbed into the existing evaluation pipeline.

\textbf{Compute control.} \hspace{1.5mm}
A central practical limitation of expand-and-reduce methods is that memory usage and runtime depend on the size of the generated candidate pool and on the cost of the selection model.
Because the candidate pool is often constructed before pruning, computational problems can occur before the reduction stage has a chance to remove features.
TabPrep controls this cost before generation starts.
Each generator is assigned an explicit feature budget, which bounds the dimensionality of the expanded representation in advance.
This makes memory use and runtime more predictable, which is essential for repeated evaluations across datasets, folds, and hyperparameter configurations.

\textbf{Operator scope.}  \hspace{1.5mm}
Expand-and-reduce frameworks often include broad operation catalogs from many different families, where weak or redundant transformations are removed via selection.
This can be inefficient when operations are not aligned with the limitations of the downstream models.
For example, OpenFE \cite{zhang2023openfe} includes univariate monotonic transformations such as logarithms, even though its selection model is based on LightGBM and tree-based models are largely insensitive to monotonic transformations.
Such operations increase the candidate pool without a realistic expectation of improving predictive performance.
TabPrep therefore narrows the operation family scope before generation.
We design generators only for operation families that are motivated by identified structural patterns and by evidence that current tabular models may fail to learn these patterns reliably.
The resulting operator set is not intended to be exhaustive, but focused on transformations that are likely to provide complementary information in the modern model landscape.

\textbf{Operator representation.} \hspace{1.5mm}
Some generated operations can be represented in several closely related ways, and not all representations are equally useful or safe.
For example, raw categorical crosses can produce high-cardinality categorical features, while large collections of similar aggregation statistics can inflate the feature space without adding much distinct information.
TabPrep therefore uses conservative representations of generated features.
New categorical keys are converted into leakage-safe numerical encodings before being passed to the downstream model.
Moreover, when several operators target similar structural patterns, TabPrep retains the more expressive variants instead of enumerating many near-duplicates.
For GroupBy-based feature engineering, we therefore focus on relative GroupBy interactions rather than adding many aggregation statistics such as minimum, maximum, or median.
For arithmetic feature generation, we impose an order on the considered operations rather than treating all related variants as equally useful candidates.
This keeps the generated feature space smaller while preserving the structural information targeted by the generators.

\textbf{Pipeline role.}  \hspace{1.5mm}
Expand-and-reduce methods are usually designed as standalone autoFE stages that are applied before model training.
This is difficult to reconcile with large-scale benchmarking, where models are evaluated under repeated resampling and hyperparameter optimization.
TabPrep is instead designed as a tunable preprocessing component inside the evaluation pipeline.
Feature expansion can be enabled, disabled, or varied as part of the model configuration, reflecting the fact that engineered features are useful on some datasets but not universally beneficial.
Moreover, features that are beneficial under one model configuration may not remain beneficial under another.
This makes it important to evaluate feature expansion jointly with model hyperparameters rather than as a fixed preprocessing step.

\subsection{Analysis of the Practical Usefulness of TabPrep vs. AutoFE Libraries for Benchmarking} \label{appendix:autofe_time}
In the previous subsection, we discussed the methodological differences between TabPrep and expand-and-reduce autoFE.
Here, we analyze the practical consequences of these design choices in the TabArena benchmark setting.
For a feature engineering method to be useful as a benchmark baseline, it must not only improve predictive performance, but also run reliably under repeated evaluation across datasets, folds, and hyperparameter configurations.
We therefore compare TabPrep to autofeat and OpenFE along three practical dimensions: failure rate, training-time overhead, and the effect of using a reduced operator candidate set.

\textbf{Failure Analysis.} \hspace{1.5mm} 
autofeat failed on 33 datasets and only ran successfully on datasets with up to approximately $22{,}000$ rows and up to $60$ features.
OpenFE failed on 22 datasets and only ran successfully on datasets with up to approximately $36{,}000$ rows and up to $94$ features.
For both libraries, most failures occurred due to out-of-memory errors, while on some large but low- to medium-dimensional datasets the training ran out of the time limit.
In some runs, OpenFE did not even complete feature generation within the time limit, even though the process remained active.
In contrast, TabPrep's budgeted feature generation mechanism prevents exhaustive generation of a large number of features upfront and avoids out-of-time issues by not iterating over all possible operations on large feature sets.
Furthermore, TabPrep is implemented in AutoGluon \cite{erickson-arxiv20a}, which includes memory estimation and time-based early stopping of models.
Therefore, TabPrep is robust to failures and can be considered a practical solution by current standards \cite{schafer2025usable}.

\begin{figure}[!t]
    \centering
    \begin{subfigure}[t]{0.49\columnwidth}
        \centering
        \includegraphics[width=\linewidth]{figures/autoFE_boxplots_subset.pdf}
        \label{fig:autofe_copy}
    \end{subfigure}
    \hfill
    \begin{subfigure}[t]{0.49\columnwidth}
        \centering
        \includegraphics[width=\linewidth]{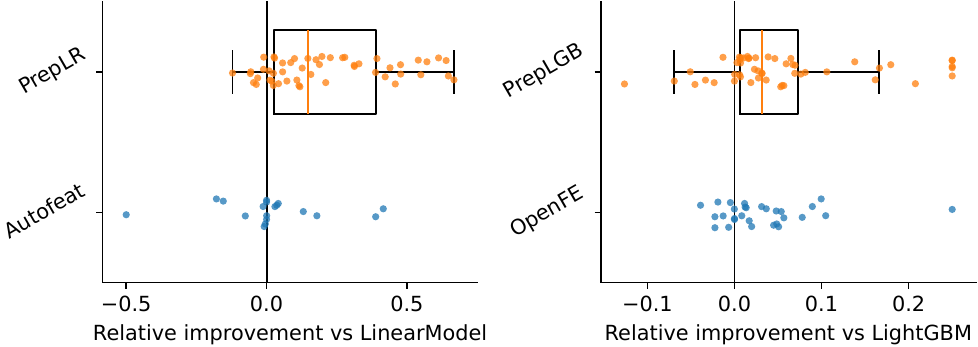}
        \label{fig:autofe_all_dats_2}
    \end{subfigure}
    \caption{\textbf{Performance comparison of TabPrep-augmented models and autoFE libraries.}
    The left panel includes only datasets on which autofeat and OpenFE ran successfully.
    The right panel includes all datasets.
    Improvements are measured relative to the corresponding model without feature engineering, i.e., a linear model for autofeat and LightGBM for OpenFE.}
    \label{fig:combined_side_by_side}
\end{figure}

\textbf{Training and HPO Time.}  \hspace{1.5mm} 
\autoref{tab:autofe_traintime} shows that, as expected with feature engineering, all approaches increase the training time compared to the default model.
However, even in the worst case, the time increase for TabPrep is comparatively low, while the other libraries often lead to impractical runtimes.
Note that this does not include the datasets where OpenFE or autofeat fail.
Therefore, the actual practical time increase would be even higher, since failed runs often also take a long time before breaking.
Hence, failed runs greatly increase the real application time of existing autoFE libraries without adding anything to benchmark performance, while TabPrep robustly finishes training.
In the main paper, we display the autoFE models in a favorable way by limiting our comparison to datasets where they ran successfully.
However, in \autoref{fig:combined_side_by_side}, we also show the picture when including all datasets, where TabPrep outperforms these libraries by an even larger margin under actual benchmark conditions.
It is a well-established fact that accurate benchmarking requires hyperparameter optimization \cite{erickson2025tabarena,gorishniy2021revisiting}.
Existing libraries already increase the training time by a large margin for a single model configuration, often without delivering any gains.
Therefore, we conclude that existing libraries following the expand-and-reduce paradigm cannot be reliably integrated into modern benchmarks and their hyperparameter optimization workflows.

\textbf{Does TabPrep Miss Something Compared to AutoFE?} \hspace{1.5mm}
The compared autoFE libraries also search over operators that we explicitly did not include in TabPrep due to a lack of evidence that they still matter for modern models.
autofeat computes univariate numerical feature transformations along with arithmetic interactions.
OpenFE computes the same transformations and additionally adds several more:
\begin{enumerate*}[label=(\textbf{\arabic*})]
    \item various kinds of GroupBy interactions, which we limited to the most informative ones on purpose;
    \item categorical feature interactions without encoding them further, which we explicitly prevent;
    \item categorical feature counts, which we exclude assuming that direct category effects are covered by our inclusion of OOF-TE;
    \item minimum and maximum operations between sets of numerical features, which we did not include due to a lack of empirical support in prior work.
\end{enumerate*}
Despite using a greatly reduced number of candidate features, TabPrep is not outperformed by the other methods.
This is a surprising result because, at least for linear models, univariate transformations of numerical features could be expected to take effect.
Overall, our results indicate that supervised selection does not offer advantages over informed, budgeted feature generation in this benchmark setting, contrary to assumptions in prior work \cite{zhang2023openfe}.

\begin{table}[ht]
    \centering
    \begin{tabular}{lcccc}
        \toprule
        & \multicolumn{2}{c}{Linear} & \multicolumn{2}{c}{LightGBM} \\
        \cmidrule(lr){2-3} \cmidrule(lr){4-5}
        & TabPrep & autofeat & TabPrep & OpenFE \\
        \midrule
        min  & 0.9  & 0.8    & 0.8  & 2.1   \\
        25\% & 10.3 & 34.2   & 2.2  & 6.9   \\
        50\% & 16.7 & 179.5  & 6.0  & 19.2  \\
        75\% & 22.9 & 464.2  & 10.5 & 92.8  \\
        max  & 52.6 & 4499.9 & 32.8 & 264.7 \\
        \bottomrule
    \end{tabular}
    \caption{\textbf{Training-time comparison of TabPrep, autofeat, and OpenFE.}
    Lower is better.
    Values show the training-time increase relative to the corresponding default model without feature engineering.
    For example, $2$ means that training is $2\times$ slower than the default model.}
    \label{tab:autofe_traintime}
\end{table}

\section{Broader Societal Impact, Baseline Framing, and Possible Extensions}

\subsection{Broader Societal Impact}
\label{app:broader-impact}

TabPrep is intended as a benchmarking baseline for tabular machine learning, and its primary positive impact is to make evaluations more complete and transparent by adding performance gains from feature-engineering steps and comparing them to performance gains from improved modeling. By providing a lightweight, configurable and generally applicable preprocessing pipeline, TabPrep may also make stronger tabular prediction performance more accessible to researchers and practitioners without requiring specialized data science knowledge, expensive model-centric innovations or specialized hardware.

TabPrep does not collect or release new sensitive data, or targets other improvements beyond improving predictive performance in standard supervised tabular learning. Therefore, we expect its potential negative societal impacts to be similar to those of general improvements in machine learning accuracy: if applied in sensitive domains such as credit, insurance, hiring, healthcare, or education, stronger predictors could improve or worsen downstream decisions depending on the data, objective, and governance process. In particular, as with any ML method, feature engineering may allow models to exploit biases, proxy variables, leakage, or spurious correlations already present in the data. TabPrep itself does not address fairness, privacy, causal validity, or deployment governance, and should therefore be complemented by domain-specific validation, fairness and privacy audits, and checks for leakage before use in high-stakes applications.

\subsection{Why is TabPrep Framed as a Benchmarking Baseline?}
\label{app:tabprep_as_baseline}
Given the substantial performance improvements, one might ask why we framed TabPrep as a benchmarking baseline instead of as an autoFE solution.
The goal of this work is not to exhaust the potential of feature engineering for tabular data, but to establish that feature engineering can be made sufficiently general, reliable, and efficient to be included in modern benchmarking pipelines. 
Therefore, although TabPrep achieves strong empirical results, we intentionally frame it as a baseline rather than as a final or optimal feature engineering method. 
In this sense, TabPrep serves as a simple, reproducible point of comparison that exposes an important missing dimension in current evaluations and fills a longstanding gap.

\textbf{TabPrep as a Benchmarking Baseline.} \hspace{1.5mm}
\label{app:tabprep_benchmarking_baseline}
There are several reasons why we understand TabPrep as a benchmarking baseline, all related to the fact that deployment-ready feature engineering tools have different requirements than feature engineering tools meant to expose potential performance gains in benchmarking. The main reasons are: \\
\begin{enumerate*}[label=(\textbf{\arabic*})]
    \item \textbf{The pipeline intentionally targets three recurring structural patterns}, selected because we were able to find evidence that they occur across datasets, and are not consistently captured by current model classes. However, they do not cover the full space of possible feature engineering operations. Domain-specific transformations, temporal or spatial structure, relational information, text-derived features, and many other sources of signal are outside the scope of the current pipeline. Consequently, the performance gains reported in this paper should be interpreted as a lower bound on what feature engineering could contribute under more specialized or adaptive approaches. 
    \\
    \item \textbf{TabPrep deliberately avoids exhaustive feature search and keeps the hyperparameter space intentionally simple}. This simplicity is important for benchmarking: a feature engineering baseline should be easy to add to existing evaluation protocols without requiring an expensive search procedure. However, it also leaves clear room for future work on adaptive generator selection, better feature budgeting, dataset-dependent operation choices, and improved interaction discovery. \\
    \item \textbf{TabPrep is designed to expose model blind spots rather than to replace model development}. The results show that current models benefit from explicitly representing patterns. This provides evidence that future tabular foundation models should incorporate stronger priors of training distributions for such patterns. A future model that learns these structures directly could reduce or eliminate the need for some of the current generators. Thus, the strong performance of TabPrep is precisely why it is useful as a baseline: it raises the standard for tabular evaluation while providing a reference point for specialized approaches to be evaluated against TabPrep in the future. \\
    \item \textbf{TabPrep was not optimized specifically for deployment}. This deliberate design choice is aligned with the benchmark setting studied in this paper. In large-scale tabular benchmarking, training time is typically the more relevant bottleneck than deployment latency. We therefore prioritize a pipeline that can be integrated into hyperparameter optimization and benchmark evaluation with manageable training overhead, rather than optimizing for a particular production latency target. The compute--performance analysis shows that TabPrep-augmented models can nevertheless lie on the Pareto frontier of performance versus inference cost, indicating practical usefulness in deployment. Finally, deployment settings vary substantially across applications: users may face different constraints on latency, memory, throughput, hardware, and retraining frequency. A single deployment-optimized configuration of feature engineering methods would therefore be unlikely to match the requirements of all use cases. Instead, we focus on demonstrating that there is substantial room for improvement from feature engineering under a controlled benchmark protocol, and leave application-specific inference-time optimization to users. 
\end{enumerate*}

\textbf{TabPrep for Deployment Scenarios.} \hspace{1.5mm}
\label{app:tabprep_for_deployment}
Our results suggest that TabPrep can be practically useful despite not being optimized for deployment: \\
\begin{enumerate*}[label=(\textbf{\arabic*})]
    \item TabPrep models already lie on the Pareto frontier of performance vs. inference time, despite not being optimized for inference time efficiency. E.g., the tuned PrepTabPFN-2.5 model has a median throughput of $\sim500$ samples/second making it practical for most settings, including all datasets in TabArena. Hence, inference time is not a bottleneck for deploying TabPrep models. \\
    \item The TabPrep generators are stateful and fully integrated within AutoGluon's AbstractFeatureGenerator framework \cite{erickson-arxiv20a}. Hence, the methods already follow common implementation practice making them usable in practice. Furthermore, the same properties that make TabPrep suitable for benchmarking (model-agnostic, fixed budgets, deterministic after fitting, manageable compute overhead) also make it a practical starting point for real-world tabular pipelines. \\
    \item TabPrep can narrow the gap between lightweight CPU-based models and more expensive GPU-dependent methods, which is directly relevant in deployment scenarios where hardware cost, throughput, or operational simplicity matter. Thus, while TabPrep increases feature dimensionality, it can also enable simpler downstream models to reach stronger performance. \\
    \item The current preprocessing logic can be reimplemented and optimized for running on GPU to reduce the inference overhead from feature generation.
    \item Simple postprocessing on top of TabPrep would already improve inference time meaningfully. Users interested in optimizing inference time could do several optimizations, e.g. reduce feature budgets, disable individual generators, prune generated features after validation, or restrict the pipeline to transformations that satisfy their operational constraints. In particular, applying feature selection via attribution methods post-validation can lead to significantly faster models at inference time. \\
\end{enumerate*}
We therefore view TabPrep as immediately usable for deployment in the settings covered in the TabArena benchmark, and further adaptable for deployment for efficiency-critical scenarios. 

\textbf{Intended Use in Future Work.} \hspace{1.5mm}
We intend TabPrep to be used as a standard feature-engineering baseline in future tabular benchmarking studies. In particular, new tabular models can not only be compared against existing model classes under minimal preprocessing, but also against models augmented with TabPrep. Including TabPrep in future evaluations can make the feature-engineering gap explicit by testing whether new methods remain competitive against baselines augmented with budgeted, pattern-informed feature generation. Conversely, TabPrep can be used as a diagnostic tool: if a new model closes the gap to TabPrep-augmented baselines without requiring explicit feature generation, this provides evidence that the model has learned inductive biases aligned with the recurring structural patterns targeted by TabPrep.
TabPrep is not meant to replace domain-specific feature engineering or to serve as an optimal automated feature-engineering system. Instead, it is intended as a lightweight, reproducible, and budget-controlled reference point for benchmarking. 
To conclude, future benchmark papers should include TabPrep as an additional preprocessing condition, future model papers should report whether their methods remain competitive against TabPrep-augmented baselines, and future feature-engineering work should use TabPrep as a baseline to demonstrate improvements over budgeted pattern-informed generation.

\subsection{Considerations on Extending TabPrep to Datasets Beyond the Scope of the Paper} \label{app:beyond_ta_scope}
TabPrep is implemented inside AutoGluon’s feature-generation framework \cite{erickson-arxiv20a}. This ensures an apples-to-apples comparison to the TabArena models, which likewise rely on AutoGluon's preprocessing. In addition, this makes TabPrep easily integrable into existing Machine Learning frameworks and workflows. Each preprocessor is implemented inheriting from the AbstractFeatureGenerator in AutoGluon and the overall TabPrep pipeline is constructed using AutoGluon's BulkFeatureGenerator. 

\paragraph{Tables with text fields.}
On tables with text fields, TabPrep works on top of AutoGluon-extracted features and natively ingests AutoGluon's FeatureMetadata information that describes which features are suitable for each preprocessor. More generally, TabPrep is orthogonal to text feature handling and compatible with methods such as Skrub, whose TableVectorizer could be added as an additional preprocessing step. Datasets with rich free-text fields are out of scope of our study. Nevertheless, TabPrep can be applied in such settings, potentially improving the categorical/numerical part of the table, while text columns are passed to be handled by a model’s usual text-processing pipeline. If text columns contain repeated values (high-cardinality text features), TabPrep can additionally exploit them as categorical features.

\paragraph{Non-IID data.}
In principle, TabPrep is orthogonal to the data split. In cases with strong distribution shifts, we recommend changing the out-of-fold cross-validation strategy in target encoding to a strategy representing the domain split. However, this is out-of-scope, and we restrict the evaluation to the IID benchmark TabArena. For temporal tabular data, the most relevant existing benchmark TabRed \cite{rubachev2024tabred} already consists of datasets that have been feature engineered, preventing a fair evaluation of our approach. In the future, we expect TabPrep to be evaluated on more heterogeneous data as well.

\paragraph{Tiny data problems.}
For limited train sizes lower than the TabArena limits ($<500$ samples), we do not recommend feature engineering due to the high risk of overfitting, since patterns are not well supported. In this scenario, simple baselines have historically worked well and, currently, foundation models have been shown to deliver state-of-the-art performance \cite{hollmann2022tabpfn}. 

\paragraph{Classification targets with high cardinality.}
Higher target cardinality only affects target encoding in TabPrep. Since these datasets are largely absent from benchmarks, we did not include dedicated handling of such cases. In practice, we would recommend limiting the target encoding to the most frequent target values.

\paragraph{Relational Data.}
Although our current experiments focus on single-table tabular data, TabPrep could be a natural and promising fit for relational settings \cite{leskovec2023databases,gan2024graph,fey2025kumorfm,cucumides2025features}, because the core structures targeted by TabPrep frequently arise in relational data. \\
In the simplest case, given a prediction entity (e.g., a customer or transaction), the task can be converted into an entity-centric table, and TabPrep can then be applied directly, as in the single-table setting used in TabArena. For a more integrated approach, the TabPrep components can also be incorporated into the relational structure itself to expand the available feature space. Each of the utilized techniques can, with minimal effort, be incorporated into a pipeline operating on top of a relational database, either within individual tables or on top of joins and relational aggregates: \\
Ordered Arithmetic Feature Expansion is generally applicable to all combinations of numerical features that can be retrieved for the prediction unit via database keys. This can also be done across tables by applying the preprocessor on top of joins.
OOF target encoding and target-encoded combinations are especially natural for relational data, which often contains many high-cardinality features (e.g., customer, merchant, or product IDs). In practice, these features can be incorporated by attaching categorical attributes to the prediction entity through joins with applying the techniques on top.
Relative GroupBy features are particularly well suited to relational data, where features such as 'value relative to customer mean' or 'price relative to category mean' can be highly predictive.
Random Subset Feature Compression can be applied within individual tables to derive a small number of features capturing higher-order categorical-like interactions. In relational settings, these compressed features can then be associated with the prediction entity through joins or suitable relational aggregation.
Furthermore, data in relational databases is often less curated and less heavily preprocessed than the single-table data used in academic benchmarks. This suggests that TabPrep may be particularly useful in relational settings. 

\section{TabArena-Full Results Per Dataset}
\label{appendix:performance_per_dataset} 
Finally, to assess statistical significance per dataset, we present the performance per dataset for all methods in the TabArena-full benchmark \cite{erickson2025tabarena}. Note that we only evaluated on the full benchmark for LightGBM and the linear model, due to the high cost of training GPU models on the full benchmark. Therefore, only PrepLinearModel, PrepLightGBM, and (Prep)LightGBM are displayed. PrepLightGBM stands for a model always using TabPrep, while (Prep)LightGBM stands for the tuning protocol that was employed for the rest of the paper with 25\% of the configurations utilizing TabPrep.

\input{per_dataset_tables_3folds.tex}


\end{document}

%% file: per_dataset_tables_3folds.tex
\begin{table}[htb]
  \centering
\caption{\textbf{Average performance per dataset with the standard deviation over 3 outer folds.}  \
                    Performance is shown in three tuning regimes: \texttt{Default}, \texttt{Tuned}, and post-hoc ensembled (\texttt{Tuned + Ens.}). \
                    We highlight the best-performing methods on three levels:  \
                    (1) \textcolor{green!50!black}{Green}: The best performing method on average; \
                    (2) \textbf{Bold}: Methods that are not worse than the best method on at least one fold. \
                    (3) \underline{Underlined}: Methods that are not worse than the best method on atleast one fold in the same pipeline regime (\texttt{Default}, \texttt{Tuned}, or \texttt{Tuned + Ens.}).}
  \begin{subtable}[t]{0.49\textwidth}
    \centering
    \scriptsize
    \caption*{APSFailure (AUC $\uparrow$)}
    \vspace{-1ex}    \label{tab:1}
    \begin{minipage}[t][][t]{\linewidth}
      \vspace{0pt}      \resizebox{\linewidth}{!}{%
\begin{tabular}{l@{\quad}lll}
      \toprule
       & Default & Tuned & Tuned + Ens. \\
      \midrule
      Linear & 0.987 $\pm$ 0.004 & 0.988 $\pm$ 0.003 & 0.990 $\pm$ 0.002 \\
      KNN & 0.959 $\pm$ 0.006 & 0.984 $\pm$ 0.002 & 0.989 $\pm$ 0.002 \\
      ExtraTrees & 0.990 $\pm$ 0.003 & 0.989 $\pm$ 0.002 & 0.991 $\pm$ 0.003 \\
      RF & 0.990 $\pm$ 0.003 & 0.990 $\pm$ 0.003 & 0.991 $\pm$ 0.002 \\
      XGBoost & 0.993 $\pm$ 0.002 & 0.992 $\pm$ 0.003 & \underline{0.993 $\pm$ 0.002} \\
      LightGBM & 0.992 $\pm$ 0.002 & 0.992 $\pm$ 0.003 & 0.992 $\pm$ 0.002 \\
      CatBoost & 0.992 $\pm$ 0.002 & 0.992 $\pm$ 0.001 & 0.992 $\pm$ 0.002 \\
\bottomrule
      \end{tabular}
      }
    \end{minipage}
  \end{subtable} \hfill
  \begin{subtable}[t]{0.49\textwidth}
    \centering
    \scriptsize
    \caption*{Amazon\_employee (AUC $\uparrow$)}
    \vspace{-1ex}    \label{tab:2}
    \begin{minipage}[t][][t]{\linewidth}
      \vspace{0pt}      \resizebox{\linewidth}{!}{%
      \begin{tabular}{l@{\quad}lll}
      \toprule
       & Default & Tuned & Tuned + Ens. \\
      \midrule
      Linear & 0.850 $\pm$ 0.010 & 0.852 $\pm$ 0.010 & 0.853 $\pm$ 0.009 \\
      KNN & 0.751 $\pm$ 0.006 & 0.818 $\pm$ 0.006 & 0.853 $\pm$ 0.006 \\
      ExtraTrees & 0.835 $\pm$ 0.003 & 0.840 $\pm$ 0.007 & 0.846 $\pm$ 0.002 \\
      RF & 0.841 $\pm$ 0.001 & 0.841 $\pm$ 0.001 & 0.850 $\pm$ 0.002 \\
      XGBoost & 0.838 $\pm$ 0.009 & 0.858 $\pm$ 0.007 & 0.862 $\pm$ 0.009 \\
      LightGBM & 0.845 $\pm$ 0.008 & 0.852 $\pm$ 0.008 & 0.858 $\pm$ 0.011 \\
      CatBoost & \underline{\textbf{0.886 $\pm$ 0.012}} & \underline{\textbf{0.887 $\pm$ 0.011}} & \underline{0.886 $\pm$ 0.011} \\
\bottomrule
      \end{tabular}
      }
    \end{minipage}
  \end{subtable}
  \medskip

  \begin{subtable}[t]{0.49\textwidth}
    \centering
    \scriptsize
    \caption*{Fiat-500 (rmse $\downarrow$)}
    \vspace{-1ex}    \label{tab:3}
    \begin{minipage}[t][][t]{\linewidth}
      \vspace{0pt}      \resizebox{\linewidth}{!}{%
      \begin{tabular}{l@{\quad}lll}
      \toprule
       & Default & Tuned & Tuned + Ens. \\
      \midrule
      Linear & 792.8 $\pm$ 29.5 & 761.3 $\pm$ 34.0 & 761.2 $\pm$ 33.9 \\
      KNN & 822.3 $\pm$ 40.0 & 796.3 $\pm$ 44.4 & 786.6 $\pm$ 40.5 \\
      ExtraTrees & 738.0 $\pm$ 37.0 & 733.5 $\pm$ 39.8 & 730.0 $\pm$ 34.6 \\
      RF & 756.1 $\pm$ 32.1 & 734.6 $\pm$ 34.1 & 734.7 $\pm$ 34.9 \\
      XGBoost & 755.2 $\pm$ 23.1 & 735.7 $\pm$ 27.7 & 732.9 $\pm$ 27.3 \\
      LightGBM & 748.5 $\pm$ 23.2 & 741.3 $\pm$ 30.8 & 727.1 $\pm$ 28.6 \\
      CatBoost & 735.4 $\pm$ 25.5 & 735.3 $\pm$ 25.2 & 731.0 $\pm$ 28.1 \\
      EBM & 745.9 $\pm$ 26.1 & 748.2 $\pm$ 24.6 & 741.8 $\pm$ 26.8 \\
      FastAIMLP & 758.7 $\pm$ 31.1 & 760.7 $\pm$ 32.7 & 757.0 $\pm$ 37.1 \\
      TorchMLP & 767.8 $\pm$ 25.2 & 762.6 $\pm$ 31.9 & 759.5 $\pm$ 33.2 \\
      MNCA & 746.9 $\pm$ 41.9 & 751.4 $\pm$ 46.5 & 730.4 $\pm$ 42.3 \\
      TabM & 748.4 $\pm$ 24.0 & 744.8 $\pm$ 27.6 & 741.8 $\pm$ 26.5 \\
      RealMLP & 752.8 $\pm$ 38.5 & 724.0 $\pm$ 28.4 & 719.2 $\pm$ 27.9 \\
      xRFM & 759.7 $\pm$ 37.9 & 739.7 $\pm$ 34.2 & 731.9 $\pm$ 33.8 \\
      TabDPT & \underline{719.8 $\pm$ 39.0} & 722.2 $\pm$ 39.0 & 716.6 $\pm$ 40.8 \\
      RealTabPFN-2.5 & \underline{726.6 $\pm$ 39.3} & 728.1 $\pm$ 36.3 & 720.3 $\pm$ 35.5 \\
      SAP-RPT-OSS & 729.3 $\pm$ 29.2 & - & - \\
      TabPFN-2.6 & 724.0 $\pm$ 34.3 & - & - \\
      TabICL-V2 & \underline{712.9 $\pm$ 26.7} & - & - \\
      PrepLinear & 751.5 $\pm$ 36.8 & 757.0 $\pm$ 27.3 & 746.0 $\pm$ 31.7 \\
      PrepLightGBM & 721.0 $\pm$ 32.5 & 722.1 $\pm$ 20.2 & 714.3 $\pm$ 31.9 \\
      PrepTabM & 729.0 $\pm$ 28.4 & 738.1 $\pm$ 23.0 & 725.6 $\pm$ 30.9 \\
      PrepTabPFN-2.5 & \underline{726.6 $\pm$ 39.3} & \underline{706.9 $\pm$ 35.8} & \underline{\textbf{\textcolor{green!50!black}{700.6 $\pm$ 35.7}}} \\
      \bottomrule
      \end{tabular}
      }
    \end{minipage}
  \end{subtable} \hfill
  \begin{subtable}[t]{0.49\textwidth}
    \centering
    \scriptsize
    \caption*{Bank\_Customer\_Churn (AUC $\uparrow$)}
    \vspace{-1ex}    \label{tab:4}
    \begin{minipage}[t][][t]{\linewidth}
      \vspace{0pt}      \resizebox{\linewidth}{!}{%
      \begin{tabular}{l@{\quad}lll}
      \toprule
       & Default & Tuned & Tuned + Ens. \\
      \midrule
      Linear & 0.772 $\pm$ 0.007 & 0.773 $\pm$ 0.006 & 0.773 $\pm$ 0.006 \\
      KNN & 0.814 $\pm$ 0.011 & 0.828 $\pm$ 0.007 & 0.841 $\pm$ 0.007 \\
      ExtraTrees & 0.852 $\pm$ 0.006 & 0.859 $\pm$ 0.008 & 0.862 $\pm$ 0.008 \\
      RF & 0.852 $\pm$ 0.002 & 0.855 $\pm$ 0.005 & 0.858 $\pm$ 0.005 \\
      XGBoost & 0.866 $\pm$ 0.011 & 0.869 $\pm$ 0.011 & 0.868 $\pm$ 0.010 \\
      LightGBM & 0.866 $\pm$ 0.009 & 0.868 $\pm$ 0.010 & 0.869 $\pm$ 0.010 \\
      CatBoost & 0.871 $\pm$ 0.011 & 0.871 $\pm$ 0.010 & 0.871 $\pm$ 0.010 \\
      EBM & 0.863 $\pm$ 0.008 & 0.867 $\pm$ 0.011 & 0.867 $\pm$ 0.010 \\
      FastAIMLP & 0.861 $\pm$ 0.006 & 0.865 $\pm$ 0.007 & 0.865 $\pm$ 0.007 \\
      TorchMLP & 0.861 $\pm$ 0.009 & 0.868 $\pm$ 0.010 & 0.867 $\pm$ 0.010 \\
      MNCA & 0.866 $\pm$ 0.007 & 0.871 $\pm$ 0.010 & 0.871 $\pm$ 0.009 \\
      TabM & 0.872 $\pm$ 0.011 & 0.873 $\pm$ 0.012 & 0.873 $\pm$ 0.011 \\
      RealMLP & 0.868 $\pm$ 0.010 & 0.872 $\pm$ 0.010 & 0.872 $\pm$ 0.010 \\
      xRFM & 0.836 $\pm$ 0.004 & 0.864 $\pm$ 0.007 & 0.865 $\pm$ 0.008 \\
      TabDPT & 0.866 $\pm$ 0.009 & 0.867 $\pm$ 0.008 & 0.867 $\pm$ 0.008 \\
      RealTabPFN-2.5 & \textbf{0.874 $\pm$ 0.010} & 0.875 $\pm$ 0.011 & \textbf{0.875 $\pm$ 0.010} \\
      SAP-RPT-OSS & 0.869 $\pm$ 0.009 & - & - \\
      TabPFN-2.6 & 0.873 $\pm$ 0.010 & - & - \\
      TabICL-V2 & 0.874 $\pm$ 0.010 & - & - \\
      PrepLinear & 0.855 $\pm$ 0.007 & 0.860 $\pm$ 0.007 & 0.861 $\pm$ 0.006 \\
      PrepLightGBM & 0.867 $\pm$ 0.007 & 0.869 $\pm$ 0.009 & 0.870 $\pm$ 0.009 \\
      PrepTabM & 0.861 $\pm$ 0.007 & 0.873 $\pm$ 0.012 & 0.873 $\pm$ 0.011 \\
      PrepTabPFN-2.5 & \underline{\textbf{0.874 $\pm$ 0.010}} & \underline{\textbf{\textcolor{green!50!black}{0.875 $\pm$ 0.011}}} & \underline{\textbf{0.875 $\pm$ 0.011}} \\
      \bottomrule
      \end{tabular}
      }
    \end{minipage}
  \end{subtable}
  \medskip

\end{table}

\begin{table}[htb]
  \centering
  \begin{subtable}[t]{0.49\textwidth}
    \centering
    \scriptsize
    \caption*{Bioresponse (AUC $\uparrow$)}
    \vspace{-1ex}    \label{tab:5}
    \begin{minipage}[t][][t]{\linewidth}
      \vspace{0pt}      \resizebox{\linewidth}{!}{%
      \begin{tabular}{l@{\quad}lll}
      \toprule
       & Default & Tuned & Tuned + Ens. \\
      \midrule
      Linear & 0.789 $\pm$ 0.015 & 0.808 $\pm$ 0.016 & 0.815 $\pm$ 0.016 \\
      KNN & 0.815 $\pm$ 0.026 & 0.836 $\pm$ 0.018 & 0.845 $\pm$ 0.018 \\
      ExtraTrees & 0.867 $\pm$ 0.013 & 0.868 $\pm$ 0.014 & 0.871 $\pm$ 0.014 \\
      RF & 0.871 $\pm$ 0.011 & 0.871 $\pm$ 0.011 & 0.875 $\pm$ 0.010 \\
      XGBoost & \underline{0.875 $\pm$ 0.013} & \underline{0.875 $\pm$ 0.014} & \underline{\textbf{0.876 $\pm$ 0.014}} \\
      LightGBM & \underline{0.875 $\pm$ 0.013} & 0.875 $\pm$ 0.013 & 0.876 $\pm$ 0.014 \\
      CatBoost & \underline{0.872 $\pm$ 0.015} & \underline{0.874 $\pm$ 0.015} & 0.874 $\pm$ 0.014 \\
\bottomrule
      \end{tabular}
      }
    \end{minipage}
  \end{subtable} \hfill
  \begin{subtable}[t]{0.49\textwidth}
    \centering
    \scriptsize
    \caption*{Diabetes130US (AUC $\uparrow$)}
    \vspace{-1ex}    \label{tab:6}
    \begin{minipage}[t][][t]{\linewidth}
      \vspace{0pt}      \resizebox{\linewidth}{!}{%
      \begin{tabular}{l@{\quad}lll}
      \toprule
       & Default & Tuned & Tuned + Ens. \\
      \midrule
      Linear & 0.645 $\pm$ 0.008 & 0.656 $\pm$ 0.011 & 0.656 $\pm$ 0.010 \\
      KNN & 0.539 $\pm$ 0.003 & 0.634 $\pm$ 0.006 & 0.651 $\pm$ 0.005 \\
      ExtraTrees & 0.622 $\pm$ 0.005 & 0.651 $\pm$ 0.006 & 0.653 $\pm$ 0.007 \\
      RF & 0.632 $\pm$ 0.005 & 0.657 $\pm$ 0.008 & 0.658 $\pm$ 0.007 \\
      XGBoost & 0.661 $\pm$ 0.013 & 0.668 $\pm$ 0.011 & 0.669 $\pm$ 0.012 \\
      LightGBM & 0.647 $\pm$ 0.009 & 0.668 $\pm$ 0.009 & 0.671 $\pm$ 0.010 \\
      CatBoost & \underline{0.671 $\pm$ 0.009} & \underline{0.671 $\pm$ 0.009} & 0.671 $\pm$ 0.010 \\
      EBM & 0.657 $\pm$ 0.009 & 0.659 $\pm$ 0.007 & 0.663 $\pm$ 0.008 \\
      FastAIMLP & 0.647 $\pm$ 0.009 & 0.653 $\pm$ 0.009 & 0.661 $\pm$ 0.009 \\
      TorchMLP & 0.655 $\pm$ 0.009 & 0.663 $\pm$ 0.011 & 0.667 $\pm$ 0.012 \\
      MNCA & 0.656 $\pm$ 0.010 & 0.660 $\pm$ 0.011 & 0.666 $\pm$ 0.009 \\
      TabM & 0.659 $\pm$ 0.008 & 0.661 $\pm$ 0.011 & 0.661 $\pm$ 0.010 \\
      RealMLP & 0.658 $\pm$ 0.011 & \underline{0.671 $\pm$ 0.011} & 0.671 $\pm$ 0.010 \\
      xRFM & 0.629 $\pm$ 0.004 & 0.665 $\pm$ 0.010 & 0.666 $\pm$ 0.008 \\
      TabDPT & 0.622 $\pm$ 0.006 & 0.638 $\pm$ 0.015 & 0.643 $\pm$ 0.009 \\
      RealTabPFN-2.5 & 0.658 $\pm$ 0.009 & 0.656 $\pm$ 0.010 & 0.657 $\pm$ 0.010 \\
      SAP-RPT-OSS & 0.653 $\pm$ 0.007 & - & - \\
      TabPFN-2.6 & \underline{0.669 $\pm$ 0.011} & - & - \\
      TabICL-V2 & 0.664 $\pm$ 0.009 & - & - \\
      PrepLinear & 0.662 $\pm$ 0.008 & 0.665 $\pm$ 0.011 & 0.668 $\pm$ 0.010 \\
      PrepLightGBM & \underline{0.670 $\pm$ 0.010} & \underline{0.672 $\pm$ 0.009} & \underline{\textbf{\textcolor{green!50!black}{0.673 $\pm$ 0.010}}} \\
      PrepTabM & 0.665 $\pm$ 0.010 & 0.669 $\pm$ 0.010 & 0.670 $\pm$ 0.010 \\
      PrepTabPFN-2.5 & 0.658 $\pm$ 0.009 & 0.657 $\pm$ 0.010 & 0.660 $\pm$ 0.009 \\
      \bottomrule
      \end{tabular}
      }
    \end{minipage}
  \end{subtable}
  \medskip

  \begin{subtable}[t]{0.49\textwidth}
    \centering
    \scriptsize
    \caption*{E-Commerce (AUC $\uparrow$)}
    \vspace{-1ex}    \label{tab:7}
    \begin{minipage}[t][][t]{\linewidth}
      \vspace{0pt}      \resizebox{\linewidth}{!}{%
      \begin{tabular}{l@{\quad}lll}
      \toprule
       & Default & Tuned & Tuned + Ens. \\
      \midrule
      Linear & 0.703 $\pm$ 0.007 & 0.721 $\pm$ 0.004 & 0.722 $\pm$ 0.005 \\
      KNN & 0.703 $\pm$ 0.000 & 0.735 $\pm$ 0.005 & 0.734 $\pm$ 0.005 \\
      ExtraTrees & 0.740 $\pm$ 0.002 & \underline{0.742 $\pm$ 0.003} & 0.742 $\pm$ 0.002 \\
      RF & 0.742 $\pm$ 0.004 & \underline{0.739 $\pm$ 0.008} & \underline{\textbf{0.742 $\pm$ 0.008}} \\
      XGBoost & 0.741 $\pm$ 0.004 & \underline{\textbf{0.744 $\pm$ 0.004}} & \underline{\textbf{0.744 $\pm$ 0.004}} \\
      LightGBM & \underline{\textbf{0.741 $\pm$ 0.007}} & \underline{0.742 $\pm$ 0.002} & \underline{0.743 $\pm$ 0.002} \\
      CatBoost & \underline{\textbf{0.744 $\pm$ 0.006}} & \textbf{0.743 $\pm$ 0.006} & \textbf{0.743 $\pm$ 0.007} \\
\bottomrule
      \end{tabular}
      }
    \end{minipage}
  \end{subtable} \hfill
  \begin{subtable}[t]{0.49\textwidth}
    \centering
    \scriptsize
    \caption*{Fitness\_Club (AUC $\uparrow$)}
    \vspace{-1ex}    \label{tab:8}
    \begin{minipage}[t][][t]{\linewidth}
      \vspace{0pt}      \resizebox{\linewidth}{!}{%
      \begin{tabular}{l@{\quad}lll}
      \toprule
       & Default & Tuned & Tuned + Ens. \\
      \midrule
      Linear & 0.817 $\pm$ 0.013 & 0.818 $\pm$ 0.016 & 0.818 $\pm$ 0.016 \\
      KNN & 0.731 $\pm$ 0.018 & 0.785 $\pm$ 0.013 & 0.802 $\pm$ 0.009 \\
      ExtraTrees & 0.773 $\pm$ 0.011 & 0.819 $\pm$ 0.013 & 0.818 $\pm$ 0.014 \\
      RF & 0.787 $\pm$ 0.012 & 0.806 $\pm$ 0.017 & 0.806 $\pm$ 0.015 \\
      XGBoost & 0.803 $\pm$ 0.014 & 0.810 $\pm$ 0.014 & 0.811 $\pm$ 0.013 \\
      LightGBM & 0.802 $\pm$ 0.011 & 0.815 $\pm$ 0.014 & 0.816 $\pm$ 0.013 \\
      CatBoost & 0.814 $\pm$ 0.013 & 0.812 $\pm$ 0.012 & 0.814 $\pm$ 0.012 \\
\bottomrule
      \end{tabular}
      }
    \end{minipage}
  \end{subtable}
  \medskip

\end{table}

\begin{table}[htb]
  \centering
  \begin{subtable}[t]{0.49\textwidth}
    \centering
    \scriptsize
    \caption*{Food\_Delivery (rmse $\downarrow$)}
    \vspace{-1ex}    \label{tab:9}
    \begin{minipage}[t][][t]{\linewidth}
      \vspace{0pt}      \resizebox{\linewidth}{!}{%
      \begin{tabular}{l@{\quad}lll}
      \toprule
       & Default & Tuned & Tuned + Ens. \\
      \midrule
      Linear & 8.565 $\pm$ 0.054 & 8.337 $\pm$ 0.059 & 8.275 $\pm$ 0.048 \\
      KNN & 8.585 $\pm$ 0.042 & 8.046 $\pm$ 0.029 & 7.916 $\pm$ 0.034 \\
      ExtraTrees & 8.205 $\pm$ 0.037 & 7.764 $\pm$ 0.060 & 7.758 $\pm$ 0.057 \\
      RF & 7.860 $\pm$ 0.027 & 7.590 $\pm$ 0.033 & 7.593 $\pm$ 0.035 \\
      XGBoost & 7.400 $\pm$ 0.044 & 7.400 $\pm$ 0.044 & 7.403 $\pm$ 0.044 \\
      LightGBM & 7.613 $\pm$ 0.051 & 7.381 $\pm$ 0.040 & 7.377 $\pm$ 0.040 \\
      CatBoost & 7.378 $\pm$ 0.045 & 7.369 $\pm$ 0.036 & 7.370 $\pm$ 0.039 \\
      EBM & 7.439 $\pm$ 0.040 & 7.436 $\pm$ 0.045 & 7.410 $\pm$ 0.034 \\
      FastAIMLP & 8.183 $\pm$ 0.030 & 8.087 $\pm$ 0.056 & 8.064 $\pm$ 0.042 \\
      TorchMLP & 7.737 $\pm$ 0.049 & 7.580 $\pm$ 0.029 & 7.559 $\pm$ 0.031 \\
      MNCA & 7.490 $\pm$ 0.037 & 7.422 $\pm$ 0.016 & 7.381 $\pm$ 0.026 \\
      TabM & 7.760 $\pm$ 0.047 & 7.651 $\pm$ 0.046 & 7.653 $\pm$ 0.048 \\
      RealMLP & 7.938 $\pm$ 0.048 & 7.408 $\pm$ 0.024 & 7.404 $\pm$ 0.030 \\
      xRFM & 7.637 $\pm$ 0.034 & 7.484 $\pm$ 0.017 & 7.448 $\pm$ 0.020 \\
      TabDPT & 7.468 $\pm$ 0.036 & 7.421 $\pm$ 0.025 & 7.415 $\pm$ 0.025 \\
      RealTabPFN-2.5 & 7.537 $\pm$ 0.050 & 7.399 $\pm$ 0.032 & 7.390 $\pm$ 0.025 \\
      SAP-RPT-OSS & 7.925 $\pm$ 0.050 & - & - \\
      TabPFN-2.6 & 7.379 $\pm$ 0.028 & - & - \\
      TabICL-V2 & 7.603 $\pm$ 0.061 & - & - \\
      PrepLinear & 7.430 $\pm$ 0.017 & 7.392 $\pm$ 0.017 & 7.377 $\pm$ 0.014 \\
      PrepLightGBM & \underline{7.220 $\pm$ 0.009} & 7.220 $\pm$ 0.007 & 7.218 $\pm$ 0.006 \\
      PrepTabM & 7.249 $\pm$ 0.009 & 7.242 $\pm$ 0.015 & 7.239 $\pm$ 0.011 \\
      PrepTabPFN-2.5 & 7.537 $\pm$ 0.050 & \underline{7.204 $\pm$ 0.005} & \underline{\textbf{\textcolor{green!50!black}{7.199 $\pm$ 0.004}}} \\
      \bottomrule
      \end{tabular}
      }
    \end{minipage}
  \end{subtable} \hfill
  \begin{subtable}[t]{0.49\textwidth}
    \centering
    \scriptsize
    \caption*{GiveMeSomeCredit (AUC $\uparrow$)}
    \vspace{-1ex}    \label{tab:10}
    \begin{minipage}[t][][t]{\linewidth}
      \vspace{0pt}      \resizebox{\linewidth}{!}{%
      \begin{tabular}{l@{\quad}lll}
      \toprule
       & Default & Tuned & Tuned + Ens. \\
      \midrule
      Linear & 0.841 $\pm$ 0.003 & 0.841 $\pm$ 0.003 & 0.841 $\pm$ 0.003 \\
      KNN & 0.748 $\pm$ 0.006 & 0.858 $\pm$ 0.003 & 0.859 $\pm$ 0.003 \\
      ExtraTrees & 0.840 $\pm$ 0.005 & 0.857 $\pm$ 0.003 & 0.857 $\pm$ 0.004 \\
      RF & 0.848 $\pm$ 0.005 & 0.863 $\pm$ 0.004 & 0.863 $\pm$ 0.004 \\
      XGBoost & 0.865 $\pm$ 0.003 & 0.866 $\pm$ 0.003 & 0.866 $\pm$ 0.003 \\
      LightGBM & 0.865 $\pm$ 0.003 & 0.866 $\pm$ 0.004 & 0.867 $\pm$ 0.004 \\
      CatBoost & 0.866 $\pm$ 0.003 & 0.867 $\pm$ 0.003 & 0.867 $\pm$ 0.003 \\
      EBM & 0.866 $\pm$ 0.004 & 0.866 $\pm$ 0.004 & 0.866 $\pm$ 0.004 \\
      FastAIMLP & 0.830 $\pm$ 0.002 & 0.843 $\pm$ 0.006 & 0.847 $\pm$ 0.002 \\
      TorchMLP & 0.864 $\pm$ 0.002 & 0.864 $\pm$ 0.003 & 0.865 $\pm$ 0.002 \\
      MNCA & 0.866 $\pm$ 0.003 & 0.867 $\pm$ 0.003 & 0.867 $\pm$ 0.003 \\
      TabM & 0.866 $\pm$ 0.003 & \underline{0.867 $\pm$ 0.003} & \underline{\textbf{0.867 $\pm$ 0.003}} \\
      RealMLP & 0.866 $\pm$ 0.003 & 0.866 $\pm$ 0.003 & 0.866 $\pm$ 0.003 \\
      xRFM & 0.746 $\pm$ 0.009 & 0.863 $\pm$ 0.003 & 0.860 $\pm$ 0.003 \\
      TabDPT & 0.845 $\pm$ 0.004 & 0.862 $\pm$ 0.005 & 0.865 $\pm$ 0.004 \\
      RealTabPFN-2.5 & 0.866 $\pm$ 0.003 & 0.866 $\pm$ 0.003 & 0.867 $\pm$ 0.003 \\
      SAP-RPT-OSS & 0.862 $\pm$ 0.003 & - & - \\
      TabPFN-2.6 & \underline{\textbf{0.867 $\pm$ 0.003}} & - & - \\
      TabICL-V2 & \underline{\textbf{0.867 $\pm$ 0.003}} & - & - \\
      PrepLinear & 0.862 $\pm$ 0.004 & 0.864 $\pm$ 0.003 & 0.865 $\pm$ 0.004 \\
      PrepLightGBM & 0.866 $\pm$ 0.003 & 0.866 $\pm$ 0.004 & 0.867 $\pm$ 0.004 \\
      PrepTabM & 0.866 $\pm$ 0.003 & 0.867 $\pm$ 0.003 & \underline{\textbf{\textcolor{green!50!black}{0.868 $\pm$ 0.003}}} \\
      PrepTabPFN-2.5 & 0.866 $\pm$ 0.003 & 0.866 $\pm$ 0.003 & 0.867 $\pm$ 0.003 \\
      \bottomrule
      \end{tabular}
      }
    \end{minipage}
  \end{subtable}
  \medskip

  \begin{subtable}[t]{0.49\textwidth}
    \centering
    \scriptsize
    \caption*{HR\_Analytics (AUC $\uparrow$)}
    \vspace{-1ex}    \label{tab:11}
    \begin{minipage}[t][][t]{\linewidth}
      \vspace{0pt}      \resizebox{\linewidth}{!}{%
      \begin{tabular}{l@{\quad}lll}
      \toprule
       & Default & Tuned & Tuned + Ens. \\
      \midrule
      Linear & 0.796 $\pm$ 0.008 & 0.797 $\pm$ 0.006 & 0.797 $\pm$ 0.007 \\
      KNN & 0.747 $\pm$ 0.009 & 0.788 $\pm$ 0.010 & 0.798 $\pm$ 0.010 \\
      ExtraTrees & 0.783 $\pm$ 0.010 & 0.800 $\pm$ 0.011 & 0.801 $\pm$ 0.011 \\
      RF & 0.787 $\pm$ 0.009 & 0.802 $\pm$ 0.009 & 0.802 $\pm$ 0.009 \\
      XGBoost & \underline{\textbf{0.806 $\pm$ 0.009}} & \underline{0.804 $\pm$ 0.007} & \underline{\textbf{0.805 $\pm$ 0.007}} \\
      LightGBM & 0.802 $\pm$ 0.009 & \underline{0.804 $\pm$ 0.009} & \underline{0.805 $\pm$ 0.008} \\
      CatBoost & \underline{\textbf{0.805 $\pm$ 0.007}} & \textbf{0.805 $\pm$ 0.007} & 0.804 $\pm$ 0.008 \\
\bottomrule
      \end{tabular}
      }
    \end{minipage}
  \end{subtable} \hfill
  \begin{subtable}[t]{0.49\textwidth}
    \centering
    \scriptsize
    \caption*{Is-good-customer (AUC $\uparrow$)}
    \vspace{-1ex}    \label{tab:12}
    \begin{minipage}[t][][t]{\linewidth}
      \vspace{0pt}      \resizebox{\linewidth}{!}{%
      \begin{tabular}{l@{\quad}lll}
      \toprule
       & Default & Tuned & Tuned + Ens. \\
      \midrule
      Linear & 0.733 $\pm$ 0.011 & 0.724 $\pm$ 0.012 & 0.728 $\pm$ 0.010 \\
      KNN & 0.663 $\pm$ 0.040 & 0.739 $\pm$ 0.020 & 0.729 $\pm$ 0.011 \\
      ExtraTrees & 0.693 $\pm$ 0.021 & 0.702 $\pm$ 0.025 & 0.707 $\pm$ 0.025 \\
      RF & 0.713 $\pm$ 0.044 & \underline{0.734 $\pm$ 0.032} & \underline{\textbf{0.726 $\pm$ 0.042}} \\
      XGBoost & 0.724 $\pm$ 0.021 & 0.738 $\pm$ 0.021 & \underline{\textbf{0.740 $\pm$ 0.025}} \\
      LightGBM & 0.718 $\pm$ 0.024 & \underline{\textbf{0.735 $\pm$ 0.039}} & \underline{\textbf{0.746 $\pm$ 0.024}} \\
      CatBoost & 0.746 $\pm$ 0.020 & 0.740 $\pm$ 0.015 & \textbf{0.746 $\pm$ 0.007} \\
\bottomrule
      \end{tabular}
      }
    \end{minipage}
  \end{subtable}
  \medskip

\end{table}

\begin{table}[htb]
  \centering
  \begin{subtable}[t]{0.49\textwidth}
    \centering
    \scriptsize
    \caption*{MIC (logloss $\downarrow$)}
    \vspace{-1ex}    \label{tab:13}
    \begin{minipage}[t][][t]{\linewidth}
      \vspace{0pt}      \resizebox{\linewidth}{!}{%
      \begin{tabular}{l@{\quad}lll}
      \toprule
       & Default & Tuned & Tuned + Ens. \\
      \midrule
      Linear & 0.594 $\pm$ 0.042 & 0.473 $\pm$ 0.027 & 0.474 $\pm$ 0.030 \\
      KNN & 1.409 $\pm$ 0.099 & 0.673 $\pm$ 0.044 & 0.592 $\pm$ 0.037 \\
      ExtraTrees & 0.501 $\pm$ 0.023 & 0.480 $\pm$ 0.025 & 0.469 $\pm$ 0.025 \\
      RF & 0.507 $\pm$ 0.057 & 0.485 $\pm$ 0.024 & 0.477 $\pm$ 0.023 \\
      XGBoost & 0.475 $\pm$ 0.026 & 0.448 $\pm$ 0.023 & 0.449 $\pm$ 0.024 \\
      LightGBM & 0.509 $\pm$ 0.021 & 0.457 $\pm$ 0.027 & 0.458 $\pm$ 0.026 \\
      CatBoost & 0.458 $\pm$ 0.018 & 0.453 $\pm$ 0.015 & 0.450 $\pm$ 0.016 \\
\bottomrule
      \end{tabular}
      }
    \end{minipage}
  \end{subtable} \hfill
  \begin{subtable}[t]{0.49\textwidth}
    \centering
    \scriptsize
    \caption*{Marketing\_Campaign (AUC $\uparrow$)}
    \vspace{-1ex}    \label{tab:14}
    \begin{minipage}[t][][t]{\linewidth}
      \vspace{0pt}      \resizebox{\linewidth}{!}{%
      \begin{tabular}{l@{\quad}lll}
      \toprule
       & Default & Tuned & Tuned + Ens. \\
      \midrule
      Linear & 0.909 $\pm$ 0.013 & \underline{0.909 $\pm$ 0.013} & 0.908 $\pm$ 0.014 \\
      KNN & 0.848 $\pm$ 0.005 & 0.862 $\pm$ 0.023 & 0.886 $\pm$ 0.023 \\
      ExtraTrees & 0.889 $\pm$ 0.018 & 0.894 $\pm$ 0.023 & 0.893 $\pm$ 0.023 \\
      RF & 0.884 $\pm$ 0.025 & 0.880 $\pm$ 0.031 & 0.883 $\pm$ 0.027 \\
      XGBoost & 0.898 $\pm$ 0.027 & 0.908 $\pm$ 0.024 & 0.908 $\pm$ 0.024 \\
      LightGBM & 0.907 $\pm$ 0.024 & 0.915 $\pm$ 0.021 & 0.916 $\pm$ 0.021 \\
      CatBoost & 0.909 $\pm$ 0.026 & 0.904 $\pm$ 0.022 & 0.904 $\pm$ 0.027 \\
\bottomrule
      \end{tabular}
      }
    \end{minipage}
  \end{subtable}
  \medskip

  \begin{subtable}[t]{0.49\textwidth}
    \centering
    \scriptsize
    \caption*{NATICUSdroid (AUC $\uparrow$)}
    \vspace{-1ex}    \label{tab:15}
    \begin{minipage}[t][][t]{\linewidth}
      \vspace{0pt}      \resizebox{\linewidth}{!}{%
      \begin{tabular}{l@{\quad}lll}
      \toprule
       & Default & Tuned & Tuned + Ens. \\
      \midrule
      Linear & 0.981 $\pm$ 0.001 & 0.981 $\pm$ 0.001 & 0.981 $\pm$ 0.001 \\
      KNN & 0.950 $\pm$ 0.003 & 0.978 $\pm$ 0.003 & 0.980 $\pm$ 0.004 \\
      ExtraTrees & 0.979 $\pm$ 0.001 & 0.982 $\pm$ 0.002 & 0.982 $\pm$ 0.002 \\
      RF & 0.979 $\pm$ 0.002 & 0.981 $\pm$ 0.002 & 0.981 $\pm$ 0.002 \\
      XGBoost & 0.985 $\pm$ 0.002 & 0.986 $\pm$ 0.002 & 0.986 $\pm$ 0.002 \\
      LightGBM & 0.985 $\pm$ 0.002 & 0.986 $\pm$ 0.001 & 0.986 $\pm$ 0.002 \\
      CatBoost & 0.986 $\pm$ 0.002 & 0.986 $\pm$ 0.001 & 0.986 $\pm$ 0.001 \\
      EBM & 0.984 $\pm$ 0.001 & 0.984 $\pm$ 0.002 & 0.985 $\pm$ 0.001 \\
      FastAIMLP & 0.985 $\pm$ 0.002 & 0.985 $\pm$ 0.001 & 0.986 $\pm$ 0.001 \\
      TorchMLP & 0.985 $\pm$ 0.001 & 0.985 $\pm$ 0.000 & 0.986 $\pm$ 0.001 \\
      MNCA & 0.983 $\pm$ 0.001 & 0.985 $\pm$ 0.001 & 0.986 $\pm$ 0.001 \\
      TabM & 0.986 $\pm$ 0.001 & 0.986 $\pm$ 0.001 & 0.986 $\pm$ 0.001 \\
      RealMLP & 0.986 $\pm$ 0.001 & 0.986 $\pm$ 0.001 & 0.987 $\pm$ 0.001 \\
      xRFM & 0.976 $\pm$ 0.001 & 0.984 $\pm$ 0.001 & 0.985 $\pm$ 0.001 \\
      TabDPT & 0.984 $\pm$ 0.001 & 0.987 $\pm$ 0.001 & 0.987 $\pm$ 0.001 \\
      RealTabPFN-2.5 & 0.986 $\pm$ 0.001 & 0.986 $\pm$ 0.001 & 0.986 $\pm$ 0.001 \\
      SAP-RPT-OSS & 0.985 $\pm$ 0.001 & - & - \\
      TabPFN-2.6 & 0.986 $\pm$ 0.001 & - & - \\
      TabICL-V2 & 0.988 $\pm$ 0.001 & - & - \\
      PrepLinear & 0.986 $\pm$ 0.002 & 0.987 $\pm$ 0.002 & \underline{\textbf{0.989 $\pm$ 0.001}} \\
      PrepLightGBM & \underline{0.989 $\pm$ 0.001} & \underline{0.989 $\pm$ 0.001} & \underline{\textbf{0.989 $\pm$ 0.001}} \\
      PrepTabM & \underline{0.989 $\pm$ 0.000} & 0.988 $\pm$ 0.000 & \underline{\textbf{0.990 $\pm$ 0.000}} \\
      PrepTabPFN-2.5 & 0.986 $\pm$ 0.001 & \underline{\textbf{0.989 $\pm$ 0.000}} & \underline{\textbf{\textcolor{green!50!black}{0.990 $\pm$ 0.000}}} \\
      \bottomrule
      \end{tabular}
      }
    \end{minipage}
  \end{subtable} \hfill
  \begin{subtable}[t]{0.49\textwidth}
    \centering
    \scriptsize
    \caption*{QSAR-TID-11 (rmse $\downarrow$)}
    \vspace{-1ex}    \label{tab:16}
    \begin{minipage}[t][][t]{\linewidth}
      \vspace{0pt}      \resizebox{\linewidth}{!}{%
      \begin{tabular}{l@{\quad}lll}
      \toprule
       & Default & Tuned & Tuned + Ens. \\
      \midrule
      Linear & 1.015 $\pm$ 0.047 & 0.939 $\pm$ 0.052 & 0.935 $\pm$ 0.052 \\
      KNN & 1.000 $\pm$ 0.058 & 0.885 $\pm$ 0.071 & 0.858 $\pm$ 0.074 \\
      ExtraTrees & 0.799 $\pm$ 0.055 & 0.799 $\pm$ 0.054 & 0.786 $\pm$ 0.057 \\
      RF & 0.799 $\pm$ 0.056 & 0.799 $\pm$ 0.056 & 0.791 $\pm$ 0.057 \\
      XGBoost & 0.787 $\pm$ 0.062 & 0.760 $\pm$ 0.055 & 0.759 $\pm$ 0.058 \\
      LightGBM & 0.771 $\pm$ 0.058 & 0.756 $\pm$ 0.057 & 0.755 $\pm$ 0.057 \\
      CatBoost & 0.774 $\pm$ 0.064 & 0.773 $\pm$ 0.062 & 0.771 $\pm$ 0.063 \\
      EBM & 0.835 $\pm$ 0.056 & 0.810 $\pm$ 0.052 & 0.809 $\pm$ 0.052 \\
      FastAIMLP & 0.781 $\pm$ 0.061 & 0.771 $\pm$ 0.064 & 0.766 $\pm$ 0.066 \\
      TorchMLP & 0.775 $\pm$ 0.068 & 0.760 $\pm$ 0.058 & 0.747 $\pm$ 0.068 \\
      MNCA & 0.763 $\pm$ 0.053 & \underline{0.745 $\pm$ 0.054} & \underline{\textbf{\textcolor{green!50!black}{0.731 $\pm$ 0.051}}} \\
      TabM & 0.765 $\pm$ 0.062 & 0.758 $\pm$ 0.060 & 0.757 $\pm$ 0.060 \\
      RealMLP & 0.766 $\pm$ 0.073 & 0.753 $\pm$ 0.065 & 0.755 $\pm$ 0.066 \\
      xRFM & 0.816 $\pm$ 0.064 & 0.778 $\pm$ 0.054 & 0.764 $\pm$ 0.058 \\
      TabDPT & 0.761 $\pm$ 0.058 & \underline{\textbf{0.737 $\pm$ 0.060}} & \underline{\textbf{0.732 $\pm$ 0.059}} \\
      RealTabPFN-2.5 & 0.759 $\pm$ 0.064 & 0.747 $\pm$ 0.054 & 0.740 $\pm$ 0.056 \\
      SAP-RPT-OSS & 0.835 $\pm$ 0.065 & - & - \\
      TabPFN-2.6 & \underline{0.752 $\pm$ 0.061} & - & - \\
      TabICL-V2 & 0.802 $\pm$ 0.063 & - & - \\
      PrepLinear & 0.884 $\pm$ 0.054 & 0.891 $\pm$ 0.046 & 0.877 $\pm$ 0.056 \\
      PrepLightGBM & 0.827 $\pm$ 0.060 & 0.759 $\pm$ 0.054 & 0.754 $\pm$ 0.056 \\
      PrepTabM & 0.802 $\pm$ 0.058 & 0.760 $\pm$ 0.059 & 0.755 $\pm$ 0.061 \\
      PrepTabPFN-2.5 & 0.759 $\pm$ 0.064 & 0.749 $\pm$ 0.051 & 0.740 $\pm$ 0.057 \\
      \bottomrule
      \end{tabular}
      }
    \end{minipage}
  \end{subtable}
  \medskip

\end{table}

\begin{table}[htb]
  \centering
  \begin{subtable}[t]{0.49\textwidth}
    \centering
    \scriptsize
    \caption*{QSAR\_fish\_toxicity (rmse $\downarrow$)}
    \vspace{-1ex}    \label{tab:17}
    \begin{minipage}[t][][t]{\linewidth}
      \vspace{0pt}      \resizebox{\linewidth}{!}{%
      \begin{tabular}{l@{\quad}lll}
      \toprule
       & Default & Tuned & Tuned + Ens. \\
      \midrule
      Linear & 0.948 $\pm$ 0.059 & 0.948 $\pm$ 0.059 & 0.948 $\pm$ 0.058 \\
      KNN & 0.900 $\pm$ 0.096 & 0.900 $\pm$ 0.089 & 0.884 $\pm$ 0.091 \\
      ExtraTrees & 0.875 $\pm$ 0.082 & 0.866 $\pm$ 0.083 & \underline{0.866 $\pm$ 0.083} \\
      RF & 0.896 $\pm$ 0.078 & 0.881 $\pm$ 0.079 & 0.875 $\pm$ 0.079 \\
      XGBoost & 0.896 $\pm$ 0.070 & 0.872 $\pm$ 0.070 & 0.872 $\pm$ 0.067 \\
      LightGBM & 0.888 $\pm$ 0.063 & 0.883 $\pm$ 0.064 & 0.878 $\pm$ 0.062 \\
      CatBoost & 0.868 $\pm$ 0.082 & 0.874 $\pm$ 0.067 & 0.868 $\pm$ 0.075 \\
\bottomrule
      \end{tabular}
      }
    \end{minipage}
  \end{subtable} \hfill
  \begin{subtable}[t]{0.49\textwidth}
    \centering
    \scriptsize
    \caption*{SDSS17 (logloss $\downarrow$)}
    \vspace{-1ex}    \label{tab:18}
    \begin{minipage}[t][][t]{\linewidth}
      \vspace{0pt}      \resizebox{\linewidth}{!}{%
      \begin{tabular}{l@{\quad}lll}
      \toprule
       & Default & Tuned & Tuned + Ens. \\
      \midrule
      Linear & 0.147 $\pm$ 0.001 & 0.146 $\pm$ 0.001 & 0.135 $\pm$ 0.002 \\
      KNN & 0.379 $\pm$ 0.031 & 0.222 $\pm$ 0.002 & 0.153 $\pm$ 0.003 \\
      ExtraTrees & 0.131 $\pm$ 0.001 & 0.081 $\pm$ 0.001 & 0.078 $\pm$ 0.001 \\
      RF & 0.085 $\pm$ 0.001 & 0.073 $\pm$ 0.001 & 0.073 $\pm$ 0.001 \\
      XGBoost & 0.074 $\pm$ 0.001 & 0.074 $\pm$ 0.001 & 0.074 $\pm$ 0.000 \\
      LightGBM & 0.087 $\pm$ 0.001 & 0.073 $\pm$ 0.001 & 0.073 $\pm$ 0.001 \\
      CatBoost & 0.075 $\pm$ 0.001 & 0.074 $\pm$ 0.001 & 0.074 $\pm$ 0.001 \\
      EBM & 0.076 $\pm$ 0.001 & 0.073 $\pm$ 0.001 & 0.072 $\pm$ 0.001 \\
      FastAIMLP & 0.135 $\pm$ 0.002 & 0.109 $\pm$ 0.003 & 0.112 $\pm$ 0.003 \\
      TorchMLP & 0.095 $\pm$ 0.001 & 0.081 $\pm$ 0.001 & 0.081 $\pm$ 0.002 \\
      MNCA & 0.082 $\pm$ 0.001 & 0.076 $\pm$ 0.001 & 0.075 $\pm$ 0.001 \\
      TabM & 0.096 $\pm$ 0.001 & 0.084 $\pm$ 0.001 & 0.083 $\pm$ 0.001 \\
      RealMLP & 0.103 $\pm$ 0.003 & 0.086 $\pm$ 0.001 & 0.085 $\pm$ 0.001 \\
      xRFM & 0.147 $\pm$ 0.008 & 0.094 $\pm$ 0.001 & 0.091 $\pm$ 0.001 \\
      TabDPT & 0.084 $\pm$ 0.001 & 0.084 $\pm$ 0.001 & 0.080 $\pm$ 0.001 \\
      RealTabPFN-2.5 & 0.076 $\pm$ 0.002 & 0.067 $\pm$ 0.001 & 0.066 $\pm$ 0.001 \\
      SAP-RPT-OSS & 0.084 $\pm$ 0.002 & - & - \\
      TabPFN-2.6 & 0.069 $\pm$ 0.001 & - & - \\
      TabICL-V2 & 0.072 $\pm$ 0.002 & - & - \\
      PrepLinear & 0.101 $\pm$ 0.002 & 0.079 $\pm$ 0.001 & 0.078 $\pm$ 0.001 \\
      PrepLightGBM & \underline{0.065 $\pm$ 0.001} & \underline{0.065 $\pm$ 0.001} & 0.065 $\pm$ 0.001 \\
      PrepTabM & 0.077 $\pm$ 0.015 & 0.067 $\pm$ 0.001 & 0.067 $\pm$ 0.001 \\
      PrepTabPFN-2.5 & 0.076 $\pm$ 0.002 & \underline{0.064 $\pm$ 0.001} & \underline{\textbf{\textcolor{green!50!black}{0.063 $\pm$ 0.001}}} \\
      \bottomrule
      \end{tabular}
      }
    \end{minipage}
  \end{subtable}
  \medskip

  \begin{subtable}[t]{0.49\textwidth}
    \centering
    \scriptsize
    \caption*{airfoil\_self\_noise (rmse $\downarrow$)}
    \vspace{-1ex}    \label{tab:19}
    \begin{minipage}[t][][t]{\linewidth}
      \vspace{0pt}      \resizebox{\linewidth}{!}{%
      \begin{tabular}{l@{\quad}lll}
      \toprule
       & Default & Tuned & Tuned + Ens. \\
      \midrule
      Linear & 5.363 $\pm$ 0.069 & 4.829 $\pm$ 0.088 & 4.792 $\pm$ 0.087 \\
      KNN & 3.381 $\pm$ 0.136 & 2.496 $\pm$ 0.144 & 2.244 $\pm$ 0.074 \\
      ExtraTrees & 1.759 $\pm$ 0.052 & 1.652 $\pm$ 0.070 & 1.647 $\pm$ 0.045 \\
      RF & 1.840 $\pm$ 0.078 & 1.840 $\pm$ 0.078 & 1.803 $\pm$ 0.057 \\
      XGBoost & 1.505 $\pm$ 0.121 & 1.395 $\pm$ 0.113 & 1.405 $\pm$ 0.113 \\
      LightGBM & 1.519 $\pm$ 0.071 & 1.456 $\pm$ 0.076 & 1.411 $\pm$ 0.081 \\
      CatBoost & 1.534 $\pm$ 0.058 & 1.262 $\pm$ 0.096 & 1.265 $\pm$ 0.100 \\
      EBM & 1.920 $\pm$ 0.054 & 1.899 $\pm$ 0.023 & 1.871 $\pm$ 0.049 \\
      FastAIMLP & 2.359 $\pm$ 0.065 & 1.612 $\pm$ 0.077 & 1.635 $\pm$ 0.078 \\
      TorchMLP & 1.513 $\pm$ 0.036 & 1.355 $\pm$ 0.092 & 1.372 $\pm$ 0.111 \\
      MNCA & 1.540 $\pm$ 0.078 & 1.475 $\pm$ 0.077 & 1.418 $\pm$ 0.101 \\
      TabM & 1.217 $\pm$ 0.117 & 1.107 $\pm$ 0.106 & 1.098 $\pm$ 0.104 \\
      RealMLP & 1.184 $\pm$ 0.062 & 1.091 $\pm$ 0.056 & 1.071 $\pm$ 0.059 \\
      xRFM & 1.585 $\pm$ 0.101 & 1.267 $\pm$ 0.078 & 1.267 $\pm$ 0.079 \\
      TabDPT & 1.127 $\pm$ 0.120 & 1.087 $\pm$ 0.092 & 1.064 $\pm$ 0.101 \\
      RealTabPFN-2.5 & 1.106 $\pm$ 0.114 & \underline{\textbf{1.000 $\pm$ 0.083}} & \underline{\textbf{\textcolor{green!50!black}{0.980 $\pm$ 0.095}}} \\
      SAP-RPT-OSS & 1.330 $\pm$ 0.122 & - & - \\
      TabPFN-2.6 & \underline{1.050 $\pm$ 0.109} & - & - \\
      TabICL-V2 & \underline{1.072 $\pm$ 0.049} & - & - \\
      PrepLinear & 2.528 $\pm$ 0.068 & 2.447 $\pm$ 0.057 & 2.352 $\pm$ 0.007 \\
      PrepLightGBM & 1.727 $\pm$ 0.084 & 1.355 $\pm$ 0.107 & 1.324 $\pm$ 0.079 \\
      PrepTabM & 1.375 $\pm$ 0.060 & 1.107 $\pm$ 0.106 & 1.090 $\pm$ 0.097 \\
      PrepTabPFN-2.5 & 1.106 $\pm$ 0.114 & \underline{1.039 $\pm$ 0.141} & \underline{\textbf{0.983 $\pm$ 0.118}} \\
      \bottomrule
      \end{tabular}
      }
    \end{minipage}
  \end{subtable} \hfill
  \begin{subtable}[t]{0.49\textwidth}
    \centering
    \scriptsize
    \caption*{anneal (logloss $\downarrow$)}
    \vspace{-1ex}    \label{tab:20}
    \begin{minipage}[t][][t]{\linewidth}
      \vspace{0pt}      \resizebox{\linewidth}{!}{%
      \begin{tabular}{l@{\quad}lll}
      \toprule
       & Default & Tuned & Tuned + Ens. \\
      \midrule
      Linear & 0.085 $\pm$ 0.015 & 0.041 $\pm$ 0.034 & 0.031 $\pm$ 0.022 \\
      KNN & 0.141 $\pm$ 0.035 & 0.060 $\pm$ 0.023 & 0.055 $\pm$ 0.024 \\
      ExtraTrees & 0.066 $\pm$ 0.013 & 0.021 $\pm$ 0.009 & 0.022 $\pm$ 0.008 \\
      RF & 0.045 $\pm$ 0.007 & 0.025 $\pm$ 0.014 & 0.021 $\pm$ 0.006 \\
      XGBoost & 0.030 $\pm$ 0.017 & 0.023 $\pm$ 0.015 & 0.024 $\pm$ 0.014 \\
      LightGBM & 0.050 $\pm$ 0.024 & 0.026 $\pm$ 0.018 & 0.028 $\pm$ 0.018 \\
      CatBoost & 0.040 $\pm$ 0.011 & 0.016 $\pm$ 0.014 & 0.015 $\pm$ 0.012 \\
\bottomrule
      \end{tabular}
      }
    \end{minipage}
  \end{subtable}
  \medskip

\end{table}

\begin{table}[htb]
  \centering
  \begin{subtable}[t]{0.49\textwidth}
    \centering
    \scriptsize
    \caption*{bank-marketing (AUC $\uparrow$)}
    \vspace{-1ex}    \label{tab:21}
    \begin{minipage}[t][][t]{\linewidth}
      \vspace{0pt}      \resizebox{\linewidth}{!}{%
      \begin{tabular}{l@{\quad}lll}
      \toprule
       & Default & Tuned & Tuned + Ens. \\
      \midrule
      Linear & 0.748 $\pm$ 0.007 & 0.748 $\pm$ 0.007 & 0.749 $\pm$ 0.008 \\
      KNN & 0.716 $\pm$ 0.005 & 0.747 $\pm$ 0.009 & 0.755 $\pm$ 0.008 \\
      ExtraTrees & 0.722 $\pm$ 0.005 & 0.758 $\pm$ 0.007 & 0.758 $\pm$ 0.007 \\
      RF & 0.728 $\pm$ 0.005 & 0.762 $\pm$ 0.008 & 0.762 $\pm$ 0.008 \\
      XGBoost & 0.763 $\pm$ 0.008 & 0.765 $\pm$ 0.009 & 0.765 $\pm$ 0.009 \\
      LightGBM & 0.762 $\pm$ 0.007 & 0.765 $\pm$ 0.007 & 0.766 $\pm$ 0.008 \\
      CatBoost & 0.765 $\pm$ 0.008 & 0.766 $\pm$ 0.008 & 0.766 $\pm$ 0.008 \\
      EBM & 0.763 $\pm$ 0.008 & 0.762 $\pm$ 0.008 & 0.763 $\pm$ 0.008 \\
      FastAIMLP & 0.759 $\pm$ 0.009 & 0.761 $\pm$ 0.009 & 0.762 $\pm$ 0.009 \\
      TorchMLP & 0.756 $\pm$ 0.010 & 0.758 $\pm$ 0.011 & 0.759 $\pm$ 0.011 \\
      MNCA & 0.762 $\pm$ 0.007 & 0.763 $\pm$ 0.010 & 0.764 $\pm$ 0.009 \\
      TabM & 0.764 $\pm$ 0.009 & 0.764 $\pm$ 0.008 & 0.765 $\pm$ 0.008 \\
      RealMLP & 0.761 $\pm$ 0.010 & 0.764 $\pm$ 0.008 & 0.765 $\pm$ 0.009 \\
      xRFM & 0.701 $\pm$ 0.011 & 0.763 $\pm$ 0.007 & 0.763 $\pm$ 0.009 \\
      TabDPT & 0.763 $\pm$ 0.008 & 0.762 $\pm$ 0.007 & 0.764 $\pm$ 0.007 \\
      RealTabPFN-2.5 & 0.765 $\pm$ 0.008 & 0.765 $\pm$ 0.009 & 0.766 $\pm$ 0.009 \\
      SAP-RPT-OSS & 0.761 $\pm$ 0.009 & - & - \\
      TabPFN-2.6 & 0.765 $\pm$ 0.009 & - & - \\
      TabICL-V2 & 0.765 $\pm$ 0.010 & - & - \\
      PrepLinear & 0.769 $\pm$ 0.007 & 0.772 $\pm$ 0.005 & 0.774 $\pm$ 0.006 \\
      PrepLightGBM & \underline{0.771 $\pm$ 0.008} & \underline{\textbf{0.779 $\pm$ 0.008}} & \underline{\textbf{\textcolor{green!50!black}{0.780 $\pm$ 0.008}}} \\
      PrepTabM & \underline{0.771 $\pm$ 0.007} & 0.774 $\pm$ 0.007 & 0.777 $\pm$ 0.007 \\
      PrepTabPFN-2.5 & 0.765 $\pm$ 0.008 & 0.773 $\pm$ 0.006 & 0.774 $\pm$ 0.008 \\
      \bottomrule
      \end{tabular}
      }
    \end{minipage}
  \end{subtable} \hfill
  \begin{subtable}[t]{0.49\textwidth}
    \centering
    \scriptsize
    \caption*{blood-transfusion (AUC $\uparrow$)}
    \vspace{-1ex}    \label{tab:22}
    \begin{minipage}[t][][t]{\linewidth}
      \vspace{0pt}      \resizebox{\linewidth}{!}{%
      \begin{tabular}{l@{\quad}lll}
      \toprule
       & Default & Tuned & Tuned + Ens. \\
      \midrule
      Linear & 0.733 $\pm$ 0.032 & \underline{0.753 $\pm$ 0.021} & \underline{\textbf{\textcolor{green!50!black}{0.757 $\pm$ 0.022}}} \\
      KNN & 0.682 $\pm$ 0.016 & \underline{\textbf{0.728 $\pm$ 0.062}} & \underline{\textbf{0.734 $\pm$ 0.067}} \\
      ExtraTrees & 0.688 $\pm$ 0.026 & \textbf{0.717 $\pm$ 0.065} & 0.715 $\pm$ 0.054 \\
      RF & 0.680 $\pm$ 0.025 & 0.698 $\pm$ 0.031 & 0.697 $\pm$ 0.035 \\
      XGBoost & 0.697 $\pm$ 0.034 & 0.724 $\pm$ 0.047 & 0.717 $\pm$ 0.040 \\
      LightGBM & 0.706 $\pm$ 0.042 & \underline{\textbf{0.745 $\pm$ 0.047}} & 0.735 $\pm$ 0.032 \\
      CatBoost & 0.731 $\pm$ 0.042 & \textbf{0.732 $\pm$ 0.044} & 0.728 $\pm$ 0.038 \\
\bottomrule
      \end{tabular}
      }
    \end{minipage}
  \end{subtable}
  \medskip

  \begin{subtable}[t]{0.49\textwidth}
    \centering
    \scriptsize
    \caption*{churn (AUC $\uparrow$)}
    \vspace{-1ex}    \label{tab:23}
    \begin{minipage}[t][][t]{\linewidth}
      \vspace{0pt}      \resizebox{\linewidth}{!}{%
      \begin{tabular}{l@{\quad}lll}
      \toprule
       & Default & Tuned & Tuned + Ens. \\
      \midrule
      Linear & 0.785 $\pm$ 0.033 & 0.823 $\pm$ 0.005 & 0.823 $\pm$ 0.006 \\
      KNN & 0.863 $\pm$ 0.001 & 0.895 $\pm$ 0.009 & 0.898 $\pm$ 0.013 \\
      ExtraTrees & 0.917 $\pm$ 0.014 & 0.919 $\pm$ 0.014 & 0.922 $\pm$ 0.014 \\
      RF & 0.918 $\pm$ 0.014 & 0.914 $\pm$ 0.014 & 0.915 $\pm$ 0.014 \\
      XGBoost & 0.923 $\pm$ 0.012 & 0.921 $\pm$ 0.017 & 0.919 $\pm$ 0.017 \\
      LightGBM & 0.916 $\pm$ 0.014 & 0.924 $\pm$ 0.016 & 0.923 $\pm$ 0.017 \\
      CatBoost & 0.926 $\pm$ 0.010 & 0.921 $\pm$ 0.013 & 0.923 $\pm$ 0.011 \\
\bottomrule
      \end{tabular}
      }
    \end{minipage}
  \end{subtable} \hfill
  \begin{subtable}[t]{0.49\textwidth}
    \centering
    \scriptsize
    \caption*{coil2000 (AUC $\uparrow$)}
    \vspace{-1ex}    \label{tab:24}
    \begin{minipage}[t][][t]{\linewidth}
      \vspace{0pt}      \resizebox{\linewidth}{!}{%
      \begin{tabular}{l@{\quad}lll}
      \toprule
       & Default & Tuned & Tuned + Ens. \\
      \midrule
      Linear & 0.741 $\pm$ 0.006 & 0.740 $\pm$ 0.011 & 0.742 $\pm$ 0.011 \\
      KNN & 0.660 $\pm$ 0.014 & 0.721 $\pm$ 0.004 & 0.729 $\pm$ 0.003 \\
      ExtraTrees & 0.693 $\pm$ 0.017 & 0.746 $\pm$ 0.012 & 0.748 $\pm$ 0.016 \\
      RF & 0.688 $\pm$ 0.010 & 0.741 $\pm$ 0.006 & 0.741 $\pm$ 0.007 \\
      XGBoost & 0.755 $\pm$ 0.017 & 0.756 $\pm$ 0.015 & 0.760 $\pm$ 0.011 \\
      LightGBM & 0.750 $\pm$ 0.007 & 0.755 $\pm$ 0.014 & 0.759 $\pm$ 0.013 \\
      CatBoost & 0.755 $\pm$ 0.013 & 0.759 $\pm$ 0.015 & 0.759 $\pm$ 0.012 \\
      EBM & 0.753 $\pm$ 0.009 & 0.751 $\pm$ 0.007 & 0.755 $\pm$ 0.009 \\
      FastAIMLP & 0.726 $\pm$ 0.013 & 0.753 $\pm$ 0.017 & 0.751 $\pm$ 0.019 \\
      TorchMLP & 0.743 $\pm$ 0.015 & 0.753 $\pm$ 0.017 & 0.755 $\pm$ 0.018 \\
      MNCA & 0.758 $\pm$ 0.015 & 0.751 $\pm$ 0.017 & 0.764 $\pm$ 0.014 \\
      TabM & 0.764 $\pm$ 0.013 & \underline{0.766 $\pm$ 0.015} & 0.767 $\pm$ 0.014 \\
      RealMLP & 0.750 $\pm$ 0.009 & 0.761 $\pm$ 0.007 & 0.763 $\pm$ 0.009 \\
      xRFM & 0.699 $\pm$ 0.015 & 0.765 $\pm$ 0.008 & 0.763 $\pm$ 0.004 \\
      TabDPT & 0.724 $\pm$ 0.011 & 0.741 $\pm$ 0.016 & 0.741 $\pm$ 0.015 \\
      RealTabPFN-2.5 & 0.771 $\pm$ 0.012 & \textbf{0.773 $\pm$ 0.015} & \underline{\textbf{0.775 $\pm$ 0.010}} \\
      SAP-RPT-OSS & 0.749 $\pm$ 0.015 & - & - \\
      TabPFN-2.6 & \underline{\textbf{\textcolor{green!50!black}{0.776 $\pm$ 0.011}}} & - & - \\
      TabICL-V2 & \underline{\textbf{0.770 $\pm$ 0.007}} & - & - \\
      PrepLinear & 0.733 $\pm$ 0.018 & 0.742 $\pm$ 0.010 & 0.750 $\pm$ 0.014 \\
      PrepLightGBM & 0.750 $\pm$ 0.013 & 0.756 $\pm$ 0.010 & 0.758 $\pm$ 0.012 \\
      PrepTabM & 0.744 $\pm$ 0.011 & \underline{0.766 $\pm$ 0.015} & 0.767 $\pm$ 0.014 \\
      PrepTabPFN-2.5 & 0.771 $\pm$ 0.012 & \underline{\textbf{0.774 $\pm$ 0.016}} & \underline{\textbf{0.774 $\pm$ 0.009}} \\
      \bottomrule
      \end{tabular}
      }
    \end{minipage}
  \end{subtable}
  \medskip

\end{table}

\begin{table}[htb]
  \centering
  \begin{subtable}[t]{0.49\textwidth}
    \centering
    \scriptsize
    \caption*{concrete (rmse $\downarrow$)}
    \vspace{-1ex}    \label{tab:25}
    \begin{minipage}[t][][t]{\linewidth}
      \vspace{0pt}      \resizebox{\linewidth}{!}{%
      \begin{tabular}{l@{\quad}lll}
      \toprule
       & Default & Tuned & Tuned + Ens. \\
      \midrule
      Linear & 8.379 $\pm$ 0.284 & 8.264 $\pm$ 0.308 & 8.253 $\pm$ 0.339 \\
      KNN & 8.816 $\pm$ 0.304 & 6.495 $\pm$ 0.342 & 6.230 $\pm$ 0.447 \\
      ExtraTrees & 5.071 $\pm$ 0.172 & 5.055 $\pm$ 0.167 & 4.997 $\pm$ 0.212 \\
      RF & 5.227 $\pm$ 0.329 & 5.155 $\pm$ 0.324 & 5.049 $\pm$ 0.337 \\
      XGBoost & 4.753 $\pm$ 0.253 & 4.169 $\pm$ 0.098 & 4.107 $\pm$ 0.182 \\
      LightGBM & 4.340 $\pm$ 0.258 & 4.086 $\pm$ 0.246 & 4.015 $\pm$ 0.193 \\
      CatBoost & 4.151 $\pm$ 0.314 & 4.098 $\pm$ 0.253 & 4.090 $\pm$ 0.257 \\
      EBM & 4.251 $\pm$ 0.242 & 4.276 $\pm$ 0.155 & 4.210 $\pm$ 0.193 \\
      FastAIMLP & 6.156 $\pm$ 0.156 & 5.006 $\pm$ 0.400 & 5.049 $\pm$ 0.349 \\
      TorchMLP & 4.643 $\pm$ 0.149 & 4.370 $\pm$ 0.121 & 4.450 $\pm$ 0.017 \\
      MNCA & 4.769 $\pm$ 0.302 & 4.825 $\pm$ 0.802 & 4.530 $\pm$ 0.559 \\
      TabM & 4.264 $\pm$ 0.222 & 4.303 $\pm$ 0.120 & 4.144 $\pm$ 0.200 \\
      RealMLP & 4.663 $\pm$ 0.309 & 4.175 $\pm$ 0.031 & 4.038 $\pm$ 0.092 \\
      xRFM & 5.300 $\pm$ 0.164 & 4.109 $\pm$ 0.294 & 4.117 $\pm$ 0.219 \\
      TabDPT & 4.059 $\pm$ 0.155 & \underline{3.922 $\pm$ 0.179} & \underline{3.871 $\pm$ 0.174} \\
      RealTabPFN-2.5 & 3.879 $\pm$ 0.063 & \underline{3.777 $\pm$ 0.020} & \underline{3.785 $\pm$ 0.019} \\
      SAP-RPT-OSS & 4.124 $\pm$ 0.080 & - & - \\
      TabPFN-2.6 & 3.757 $\pm$ 0.029 & - & - \\
      TabICL-V2 & \underline{\textbf{\textcolor{green!50!black}{3.682 $\pm$ 0.010}}} & - & - \\
      PrepLinear & 5.414 $\pm$ 0.141 & 5.070 $\pm$ 0.361 & 4.777 $\pm$ 0.219 \\
      PrepLightGBM & 4.366 $\pm$ 0.115 & 4.138 $\pm$ 0.203 & 4.004 $\pm$ 0.134 \\
      PrepTabM & 4.160 $\pm$ 0.136 & 4.334 $\pm$ 0.079 & 4.077 $\pm$ 0.085 \\
      PrepTabPFN-2.5 & 3.879 $\pm$ 0.063 & 3.920 $\pm$ 0.250 & \underline{3.823 $\pm$ 0.083} \\
      \bottomrule
      \end{tabular}
      }
    \end{minipage}
  \end{subtable} \hfill
  \begin{subtable}[t]{0.49\textwidth}
    \centering
    \scriptsize
    \caption*{credit-g (AUC $\uparrow$)}
    \vspace{-1ex}    \label{tab:26}
    \begin{minipage}[t][][t]{\linewidth}
      \vspace{0pt}      \resizebox{\linewidth}{!}{%
      \begin{tabular}{l@{\quad}lll}
      \toprule
       & Default & Tuned & Tuned + Ens. \\
      \midrule
      Linear & \underline{\textbf{0.776 $\pm$ 0.022}} & \underline{\textbf{0.780 $\pm$ 0.024}} & \underline{\textbf{0.779 $\pm$ 0.023}} \\
      KNN & \underline{\textbf{0.769 $\pm$ 0.036}} & \underline{\textbf{0.777 $\pm$ 0.023}} & \underline{\textbf{0.787 $\pm$ 0.024}} \\
      ExtraTrees & 0.775 $\pm$ 0.013 & 0.776 $\pm$ 0.013 & 0.776 $\pm$ 0.009 \\
      RF & 0.778 $\pm$ 0.011 & 0.761 $\pm$ 0.015 & 0.768 $\pm$ 0.012 \\
      XGBoost & \underline{\textbf{0.775 $\pm$ 0.017}} & \underline{0.780 $\pm$ 0.003} & 0.783 $\pm$ 0.002 \\
      LightGBM & 0.764 $\pm$ 0.008 & \underline{\textbf{0.785 $\pm$ 0.022}} & 0.785 $\pm$ 0.006 \\
      CatBoost & 0.784 $\pm$ 0.011 & \underline{0.788 $\pm$ 0.006} & \underline{\textbf{0.791 $\pm$ 0.004}} \\
\bottomrule
      \end{tabular}
      }
    \end{minipage}
  \end{subtable}
  \medskip

  \begin{subtable}[t]{0.49\textwidth}
    \centering
    \scriptsize
    \caption*{credit\_card\_clients (AUC $\uparrow$)}
    \vspace{-1ex}    \label{tab:27}
    \begin{minipage}[t][][t]{\linewidth}
      \vspace{0pt}      \resizebox{\linewidth}{!}{%
      \begin{tabular}{l@{\quad}lll}
      \toprule
       & Default & Tuned & Tuned + Ens. \\
      \midrule
      Linear & 0.746 $\pm$ 0.005 & 0.746 $\pm$ 0.005 & 0.746 $\pm$ 0.005 \\
      KNN & 0.736 $\pm$ 0.005 & 0.759 $\pm$ 0.006 & 0.769 $\pm$ 0.006 \\
      ExtraTrees & 0.766 $\pm$ 0.006 & 0.781 $\pm$ 0.006 & 0.781 $\pm$ 0.006 \\
      RF & 0.765 $\pm$ 0.007 & 0.780 $\pm$ 0.007 & 0.781 $\pm$ 0.006 \\
      XGBoost & 0.783 $\pm$ 0.005 & 0.785 $\pm$ 0.005 & 0.785 $\pm$ 0.005 \\
      LightGBM & 0.784 $\pm$ 0.005 & 0.785 $\pm$ 0.005 & 0.786 $\pm$ 0.005 \\
      CatBoost & 0.785 $\pm$ 0.005 & 0.785 $\pm$ 0.005 & 0.785 $\pm$ 0.005 \\
      EBM & 0.784 $\pm$ 0.005 & 0.784 $\pm$ 0.006 & 0.785 $\pm$ 0.006 \\
      FastAIMLP & 0.782 $\pm$ 0.005 & 0.783 $\pm$ 0.007 & 0.784 $\pm$ 0.006 \\
      TorchMLP & 0.779 $\pm$ 0.002 & 0.784 $\pm$ 0.005 & 0.786 $\pm$ 0.003 \\
      MNCA & 0.782 $\pm$ 0.005 & 0.787 $\pm$ 0.005 & 0.787 $\pm$ 0.005 \\
      TabM & 0.784 $\pm$ 0.006 & 0.788 $\pm$ 0.005 & 0.788 $\pm$ 0.005 \\
      RealMLP & 0.785 $\pm$ 0.005 & 0.785 $\pm$ 0.005 & 0.786 $\pm$ 0.005 \\
      xRFM & 0.738 $\pm$ 0.001 & 0.784 $\pm$ 0.006 & 0.783 $\pm$ 0.005 \\
      TabDPT & 0.784 $\pm$ 0.004 & 0.785 $\pm$ 0.006 & 0.787 $\pm$ 0.004 \\
      RealTabPFN-2.5 & 0.789 $\pm$ 0.005 & 0.789 $\pm$ 0.005 & \underline{0.790 $\pm$ 0.005} \\
      SAP-RPT-OSS & 0.782 $\pm$ 0.005 & - & - \\
      TabPFN-2.6 & \underline{\textbf{0.790 $\pm$ 0.005}} & - & - \\
      TabICL-V2 & \underline{\textbf{\textcolor{green!50!black}{0.791 $\pm$ 0.004}}} & - & - \\
      PrepLinear & 0.776 $\pm$ 0.003 & 0.782 $\pm$ 0.002 & 0.784 $\pm$ 0.003 \\
      PrepLightGBM & 0.788 $\pm$ 0.004 & \underline{0.789 $\pm$ 0.005} & \underline{0.789 $\pm$ 0.005} \\
      PrepTabM & 0.788 $\pm$ 0.005 & 0.788 $\pm$ 0.005 & \underline{0.789 $\pm$ 0.005} \\
      PrepTabPFN-2.5 & 0.789 $\pm$ 0.005 & \underline{0.789 $\pm$ 0.005} & \underline{0.790 $\pm$ 0.005} \\
      \bottomrule
      \end{tabular}
      }
    \end{minipage}
  \end{subtable} \hfill
  \begin{subtable}[t]{0.49\textwidth}
    \centering
    \scriptsize
    \caption*{customer\_satisfaction (AUC $\uparrow$)}
    \vspace{-1ex}    \label{tab:28}
    \begin{minipage}[t][][t]{\linewidth}
      \vspace{0pt}      \resizebox{\linewidth}{!}{%
      \begin{tabular}{l@{\quad}lll}
      \toprule
       & Default & Tuned & Tuned + Ens. \\
      \midrule
      Linear & 0.964 $\pm$ 0.001 & 0.965 $\pm$ 0.001 & 0.965 $\pm$ 0.001 \\
      KNN & 0.985 $\pm$ 0.001 & 0.985 $\pm$ 0.001 & 0.988 $\pm$ 0.000 \\
      ExtraTrees & 0.992 $\pm$ 0.000 & 0.994 $\pm$ 0.000 & 0.994 $\pm$ 0.000 \\
      RF & 0.993 $\pm$ 0.000 & 0.993 $\pm$ 0.000 & 0.993 $\pm$ 0.000 \\
      XGBoost & 0.994 $\pm$ 0.000 & 0.994 $\pm$ 0.000 & 0.994 $\pm$ 0.000 \\
      LightGBM & 0.994 $\pm$ 0.000 & 0.994 $\pm$ 0.000 & 0.995 $\pm$ 0.000 \\
      CatBoost & 0.995 $\pm$ 0.000 & 0.995 $\pm$ 0.000 & 0.995 $\pm$ 0.000 \\
      EBM & 0.989 $\pm$ 0.000 & 0.991 $\pm$ 0.001 & 0.992 $\pm$ 0.001 \\
      FastAIMLP & 0.995 $\pm$ 0.000 & 0.995 $\pm$ 0.000 & 0.995 $\pm$ 0.000 \\
      TorchMLP & 0.993 $\pm$ 0.000 & 0.995 $\pm$ 0.000 & 0.995 $\pm$ 0.000 \\
      MNCA & 0.993 $\pm$ 0.000 & 0.995 $\pm$ 0.000 & 0.995 $\pm$ 0.000 \\
      TabM & 0.995 $\pm$ 0.000 & \underline{0.995 $\pm$ 0.000} & 0.995 $\pm$ 0.000 \\
      RealMLP & 0.995 $\pm$ 0.000 & 0.995 $\pm$ 0.000 & 0.995 $\pm$ 0.000 \\
      xRFM & 0.992 $\pm$ 0.000 & 0.992 $\pm$ 0.000 & 0.993 $\pm$ 0.000 \\
      TabDPT & 0.994 $\pm$ 0.000 & 0.995 $\pm$ 0.000 & 0.995 $\pm$ 0.000 \\
      RealTabPFN-2.5 & 0.995 $\pm$ 0.000 & 0.995 $\pm$ 0.000 & 0.995 $\pm$ 0.000 \\
      SAP-RPT-OSS & 0.989 $\pm$ 0.001 & - & - \\
      TabPFN-2.6 & \underline{0.995 $\pm$ 0.000} & - & - \\
      TabICL-V2 & 0.995 $\pm$ 0.000 & - & - \\
      PrepLinear & 0.986 $\pm$ 0.001 & 0.991 $\pm$ 0.000 & 0.992 $\pm$ 0.000 \\
      PrepLightGBM & 0.994 $\pm$ 0.000 & 0.994 $\pm$ 0.000 & 0.995 $\pm$ 0.000 \\
      PrepTabM & 0.995 $\pm$ 0.000 & 0.995 $\pm$ 0.000 & \underline{\textbf{\textcolor{green!50!black}{0.995 $\pm$ 0.000}}} \\
      PrepTabPFN-2.5 & 0.995 $\pm$ 0.000 & 0.995 $\pm$ 0.000 & 0.995 $\pm$ 0.000 \\
      \bottomrule
      \end{tabular}
      }
    \end{minipage}
  \end{subtable}
  \medskip

\end{table}

\begin{table}[htb]
  \centering
  \begin{subtable}[t]{0.49\textwidth}
    \centering
    \scriptsize
    \caption*{diabetes (AUC $\uparrow$)}
    \vspace{-1ex}    \label{tab:29}
    \begin{minipage}[t][][t]{\linewidth}
      \vspace{0pt}      \resizebox{\linewidth}{!}{%
      \begin{tabular}{l@{\quad}lll}
      \toprule
       & Default & Tuned & Tuned + Ens. \\
      \midrule
      Linear & 0.831 $\pm$ 0.008 & 0.832 $\pm$ 0.006 & 0.830 $\pm$ 0.004 \\
      KNN & 0.820 $\pm$ 0.006 & \underline{\textbf{0.830 $\pm$ 0.023}} & 0.831 $\pm$ 0.012 \\
      ExtraTrees & 0.830 $\pm$ 0.005 & 0.838 $\pm$ 0.006 & 0.840 $\pm$ 0.006 \\
      RF & 0.828 $\pm$ 0.009 & 0.833 $\pm$ 0.015 & 0.833 $\pm$ 0.014 \\
      XGBoost & \underline{0.833 $\pm$ 0.017} & 0.835 $\pm$ 0.012 & 0.835 $\pm$ 0.012 \\
      LightGBM & \underline{0.831 $\pm$ 0.016} & 0.840 $\pm$ 0.009 & 0.839 $\pm$ 0.009 \\
      CatBoost & 0.837 $\pm$ 0.007 & 0.839 $\pm$ 0.004 & 0.840 $\pm$ 0.004 \\
\bottomrule
      \end{tabular}
      }
    \end{minipage}
  \end{subtable} \hfill
  \begin{subtable}[t]{0.49\textwidth}
    \centering
    \scriptsize
    \caption*{diamonds (rmse $\downarrow$)}
    \vspace{-1ex}    \label{tab:30}
    \begin{minipage}[t][][t]{\linewidth}
      \vspace{0pt}      \resizebox{\linewidth}{!}{%
      \begin{tabular}{l@{\quad}lll}
      \toprule
       & Default & Tuned & Tuned + Ens. \\
      \midrule
      Linear & 1562.1 $\pm$ 35.7 & 1131.2 $\pm$ 28.0 & 1131.2 $\pm$ 28.0 \\
      KNN & 956.8 $\pm$ 21.2 & 704.2 $\pm$ 18.9 & 668.4 $\pm$ 18.6 \\
      ExtraTrees & 536.2 $\pm$ 4.7 & 536.1 $\pm$ 4.8 & 534.0 $\pm$ 3.7 \\
      RF & 547.8 $\pm$ 1.3 & 547.8 $\pm$ 1.3 & 545.7 $\pm$ 3.1 \\
      XGBoost & 537.8 $\pm$ 3.0 & 531.6 $\pm$ 4.4 & 529.5 $\pm$ 5.0 \\
      LightGBM & 533.6 $\pm$ 5.4 & 525.3 $\pm$ 3.3 & 521.2 $\pm$ 4.1 \\
      CatBoost & 522.0 $\pm$ 7.6 & 522.0 $\pm$ 7.6 & 522.5 $\pm$ 6.2 \\
      EBM & 599.7 $\pm$ 6.6 & 598.7 $\pm$ 7.7 & 595.8 $\pm$ 6.8 \\
      FastAIMLP & 560.0 $\pm$ 5.9 & 559.8 $\pm$ 7.3 & 548.7 $\pm$ 3.0 \\
      TorchMLP & 627.1 $\pm$ 20.4 & 552.6 $\pm$ 18.0 & 545.6 $\pm$ 14.0 \\
      MNCA & 525.2 $\pm$ 4.9 & 525.4 $\pm$ 3.8 & 510.8 $\pm$ 2.7 \\
      TabM & 522.5 $\pm$ 4.3 & 522.0 $\pm$ 4.5 & 519.1 $\pm$ 5.7 \\
      RealMLP & 531.1 $\pm$ 2.3 & 514.9 $\pm$ 1.2 & 510.0 $\pm$ 2.5 \\
      xRFM & 531.2 $\pm$ 3.7 & 519.4 $\pm$ 7.1 & 512.5 $\pm$ 1.4 \\
      TabDPT & 526.6 $\pm$ 12.3 & 519.6 $\pm$ 12.4 & 516.9 $\pm$ 13.2 \\
      RealTabPFN-2.5 & \underline{508.5 $\pm$ 4.9} & 506.0 $\pm$ 6.1 & 505.1 $\pm$ 3.5 \\
      SAP-RPT-OSS & 563.4 $\pm$ 12.6 & - & - \\
      TabPFN-2.6 & 511.7 $\pm$ 2.8 & - & - \\
      TabICL-V2 & \underline{500.8 $\pm$ 2.4} & - & - \\
      PrepLinear & 761.8 $\pm$ 8.5 & 714.9 $\pm$ 8.6 & 810.2 $\pm$ 162.1 \\
      PrepLightGBM & 508.2 $\pm$ 7.8 & \underline{506.4 $\pm$ 4.5} & 500.7 $\pm$ 3.2 \\
      PrepTabM & 519.4 $\pm$ 8.4 & \underline{512.6 $\pm$ 8.4} & 507.2 $\pm$ 5.9 \\
      PrepTabPFN-2.5 & \underline{508.5 $\pm$ 4.9} & \underline{506.0 $\pm$ 6.1} & \underline{\textbf{\textcolor{green!50!black}{497.3 $\pm$ 3.1}}} \\
      \bottomrule
      \end{tabular}
      }
    \end{minipage}
  \end{subtable}
  \medskip

  \begin{subtable}[t]{0.49\textwidth}
    \centering
    \scriptsize
    \caption*{hazelnut-spread (AUC $\uparrow$)}
    \vspace{-1ex}    \label{tab:31}
    \begin{minipage}[t][][t]{\linewidth}
      \vspace{0pt}      \resizebox{\linewidth}{!}{%
      \begin{tabular}{l@{\quad}lll}
      \toprule
       & Default & Tuned & Tuned + Ens. \\
      \midrule
      Linear & 0.948 $\pm$ 0.005 & 0.951 $\pm$ 0.008 & 0.952 $\pm$ 0.006 \\
      KNN & 0.912 $\pm$ 0.008 & 0.932 $\pm$ 0.002 & 0.943 $\pm$ 0.004 \\
      ExtraTrees & 0.954 $\pm$ 0.003 & 0.961 $\pm$ 0.005 & 0.962 $\pm$ 0.005 \\
      RF & 0.957 $\pm$ 0.004 & 0.958 $\pm$ 0.005 & 0.959 $\pm$ 0.004 \\
      XGBoost & 0.972 $\pm$ 0.004 & 0.974 $\pm$ 0.004 & 0.974 $\pm$ 0.004 \\
      LightGBM & 0.970 $\pm$ 0.003 & 0.975 $\pm$ 0.004 & 0.976 $\pm$ 0.005 \\
      CatBoost & 0.973 $\pm$ 0.004 & 0.974 $\pm$ 0.005 & 0.973 $\pm$ 0.005 \\
      EBM & 0.971 $\pm$ 0.006 & 0.973 $\pm$ 0.006 & 0.974 $\pm$ 0.006 \\
      FastAIMLP & 0.980 $\pm$ 0.004 & 0.985 $\pm$ 0.004 & 0.984 $\pm$ 0.004 \\
      TorchMLP & 0.980 $\pm$ 0.002 & 0.985 $\pm$ 0.001 & 0.986 $\pm$ 0.002 \\
      MNCA & 0.983 $\pm$ 0.004 & 0.986 $\pm$ 0.003 & 0.987 $\pm$ 0.003 \\
      TabM & 0.965 $\pm$ 0.008 & 0.982 $\pm$ 0.003 & 0.982 $\pm$ 0.003 \\
      RealMLP & 0.984 $\pm$ 0.002 & 0.985 $\pm$ 0.002 & 0.985 $\pm$ 0.003 \\
      xRFM & 0.984 $\pm$ 0.003 & 0.985 $\pm$ 0.003 & 0.985 $\pm$ 0.004 \\
      TabDPT & 0.993 $\pm$ 0.003 & \underline{0.991 $\pm$ 0.004} & \underline{0.992 $\pm$ 0.004} \\
      RealTabPFN-2.5 & 0.991 $\pm$ 0.004 & \underline{0.991 $\pm$ 0.004} & 0.991 $\pm$ 0.004 \\
      SAP-RPT-OSS & 0.988 $\pm$ 0.003 & - & - \\
      TabPFN-2.6 & 0.991 $\pm$ 0.004 & - & - \\
      TabICL-V2 & \underline{\textbf{\textcolor{green!50!black}{0.994 $\pm$ 0.003}}} & - & - \\
      PrepLinear & 0.958 $\pm$ 0.011 & 0.969 $\pm$ 0.009 & 0.970 $\pm$ 0.009 \\
      PrepLightGBM & 0.974 $\pm$ 0.003 & 0.978 $\pm$ 0.004 & 0.978 $\pm$ 0.004 \\
      PrepTabM & 0.979 $\pm$ 0.007 & 0.979 $\pm$ 0.004 & 0.981 $\pm$ 0.005 \\
      PrepTabPFN-2.5 & 0.991 $\pm$ 0.004 & \underline{0.991 $\pm$ 0.004} & 0.991 $\pm$ 0.004 \\
      \bottomrule
      \end{tabular}
      }
    \end{minipage}
  \end{subtable} \hfill
  \begin{subtable}[t]{0.49\textwidth}
    \centering
    \scriptsize
    \caption*{healthcare\_insurance (rmse $\downarrow$)}
    \vspace{-1ex}    \label{tab:32}
    \begin{minipage}[t][][t]{\linewidth}
      \vspace{0pt}      \resizebox{\linewidth}{!}{%
      \begin{tabular}{l@{\quad}lll}
      \toprule
       & Default & Tuned & Tuned + Ens. \\
      \midrule
      Linear & 6045.6 $\pm$ 653.7 & 6049.5 $\pm$ 649.7 & 6047.7 $\pm$ 653.0 \\
      KNN & 5745.9 $\pm$ 828.3 & 5086.0 $\pm$ 602.8 & 5069.2 $\pm$ 623.4 \\
      ExtraTrees & 4861.0 $\pm$ 550.4 & 4626.1 $\pm$ 608.8 & 4624.2 $\pm$ 609.4 \\
      RF & 4910.2 $\pm$ 461.9 & 4655.4 $\pm$ 512.7 & 4651.1 $\pm$ 552.3 \\
      XGBoost & 4676.3 $\pm$ 524.9 & 4542.3 $\pm$ 566.2 & 4534.9 $\pm$ 564.2 \\
      LightGBM & 4620.4 $\pm$ 507.4 & 4529.2 $\pm$ 582.4 & 4510.7 $\pm$ 590.2 \\
      CatBoost & 4529.3 $\pm$ 629.0 & \underline{4512.2 $\pm$ 612.2} & 4520.2 $\pm$ 598.0 \\
\bottomrule
      \end{tabular}
      }
    \end{minipage}
  \end{subtable}
  \medskip

\end{table}

\begin{table}[htb]
  \centering
  \begin{subtable}[t]{0.49\textwidth}
    \centering
    \scriptsize
    \caption*{heloc (AUC $\uparrow$)}
    \vspace{-1ex}    \label{tab:33}
    \begin{minipage}[t][][t]{\linewidth}
      \vspace{0pt}      \resizebox{\linewidth}{!}{%
      \begin{tabular}{l@{\quad}lll}
      \toprule
       & Default & Tuned & Tuned + Ens. \\
      \midrule
      Linear & 0.787 $\pm$ 0.006 & 0.787 $\pm$ 0.007 & 0.787 $\pm$ 0.007 \\
      KNN & 0.744 $\pm$ 0.002 & 0.788 $\pm$ 0.008 & 0.789 $\pm$ 0.006 \\
      ExtraTrees & 0.792 $\pm$ 0.008 & 0.793 $\pm$ 0.007 & 0.794 $\pm$ 0.008 \\
      RF & 0.791 $\pm$ 0.006 & 0.793 $\pm$ 0.008 & 0.793 $\pm$ 0.007 \\
      XGBoost & 0.793 $\pm$ 0.008 & 0.797 $\pm$ 0.006 & 0.797 $\pm$ 0.006 \\
      LightGBM & 0.794 $\pm$ 0.007 & 0.799 $\pm$ 0.007 & 0.799 $\pm$ 0.007 \\
      CatBoost & 0.798 $\pm$ 0.007 & 0.798 $\pm$ 0.007 & 0.799 $\pm$ 0.007 \\
\bottomrule
      \end{tabular}
      }
    \end{minipage}
  \end{subtable} \hfill
  \begin{subtable}[t]{0.49\textwidth}
    \centering
    \scriptsize
    \caption*{hiva\_agnostic (logloss $\downarrow$)}
    \vspace{-1ex}    \label{tab:34}
    \begin{minipage}[t][][t]{\linewidth}
      \vspace{0pt}      \resizebox{\linewidth}{!}{%
      \begin{tabular}{l@{\quad}lll}
      \toprule
       & Default & Tuned & Tuned + Ens. \\
      \midrule
      Linear & 0.433 $\pm$ 0.026 & 0.321 $\pm$ 0.011 & 0.321 $\pm$ 0.011 \\
      KNN & 0.462 $\pm$ 0.015 & 0.176 $\pm$ 0.003 & \underline{\textbf{0.175 $\pm$ 0.002}} \\
      ExtraTrees & 0.255 $\pm$ 0.035 & \underline{\textbf{\textcolor{green!50!black}{0.174 $\pm$ 0.000}}} & \underline{0.174 $\pm$ 0.000} \\
      RF & 0.259 $\pm$ 0.028 & 0.174 $\pm$ 0.001 & 0.174 $\pm$ 0.001 \\
      XGBoost & 0.182 $\pm$ 0.001 & 0.179 $\pm$ 0.003 & 0.179 $\pm$ 0.003 \\
      LightGBM & \underline{0.174 $\pm$ 0.000} & 0.175 $\pm$ 0.001 & 0.175 $\pm$ 0.001 \\
      CatBoost & 0.176 $\pm$ 0.002 & 0.178 $\pm$ 0.002 & 0.178 $\pm$ 0.002 \\
\bottomrule
      \end{tabular}
      }
    \end{minipage}
  \end{subtable}
  \medskip

  \begin{subtable}[t]{0.49\textwidth}
    \centering
    \scriptsize
    \caption*{houses (rmse $\downarrow$)}
    \vspace{-1ex}    \label{tab:35}
    \begin{minipage}[t][][t]{\linewidth}
      \vspace{0pt}      \resizebox{\linewidth}{!}{%
      \begin{tabular}{l@{\quad}lll}
      \toprule
       & Default & Tuned & Tuned + Ens. \\
      \midrule
      Linear & 0.325 $\pm$ 0.004 & 0.325 $\pm$ 0.004 & 0.323 $\pm$ 0.003 \\
      KNN & 0.290 $\pm$ 0.002 & 0.281 $\pm$ 0.003 & 0.273 $\pm$ 0.003 \\
      ExtraTrees & 0.242 $\pm$ 0.003 & 0.237 $\pm$ 0.003 & 0.238 $\pm$ 0.003 \\
      RF & 0.232 $\pm$ 0.003 & 0.232 $\pm$ 0.003 & 0.230 $\pm$ 0.003 \\
      XGBoost & 0.215 $\pm$ 0.004 & 0.215 $\pm$ 0.003 & 0.214 $\pm$ 0.003 \\
      LightGBM & 0.217 $\pm$ 0.003 & 0.211 $\pm$ 0.003 & 0.211 $\pm$ 0.004 \\
      CatBoost & 0.211 $\pm$ 0.004 & 0.211 $\pm$ 0.004 & 0.211 $\pm$ 0.004 \\
      EBM & 0.231 $\pm$ 0.003 & 0.229 $\pm$ 0.002 & 0.227 $\pm$ 0.002 \\
      FastAIMLP & 0.245 $\pm$ 0.002 & 0.235 $\pm$ 0.002 & 0.235 $\pm$ 0.002 \\
      TorchMLP & 0.233 $\pm$ 0.004 & 0.227 $\pm$ 0.005 & 0.226 $\pm$ 0.004 \\
      MNCA & 0.206 $\pm$ 0.004 & 0.205 $\pm$ 0.004 & 0.201 $\pm$ 0.004 \\
      TabM & 0.212 $\pm$ 0.003 & 0.208 $\pm$ 0.004 & 0.205 $\pm$ 0.003 \\
      RealMLP & 0.225 $\pm$ 0.002 & \underline{0.199 $\pm$ 0.004} & 0.198 $\pm$ 0.004 \\
      xRFM & 0.230 $\pm$ 0.003 & 0.210 $\pm$ 0.004 & 0.208 $\pm$ 0.003 \\
      TabDPT & 0.206 $\pm$ 0.004 & 0.203 $\pm$ 0.005 & 0.202 $\pm$ 0.004 \\
      RealTabPFN-2.5 & 0.202 $\pm$ 0.004 & 0.197 $\pm$ 0.005 & 0.196 $\pm$ 0.004 \\
      SAP-RPT-OSS & 0.203 $\pm$ 0.003 & - & - \\
      TabPFN-2.6 & 0.197 $\pm$ 0.004 & - & - \\
      TabICL-V2 & \underline{\textbf{0.196 $\pm$ 0.005}} & - & - \\
      PrepLinear & 0.280 $\pm$ 0.003 & 0.271 $\pm$ 0.003 & 0.265 $\pm$ 0.003 \\
      PrepLightGBM & 0.207 $\pm$ 0.003 & 0.206 $\pm$ 0.001 & 0.203 $\pm$ 0.002 \\
      PrepTabM & 0.214 $\pm$ 0.002 & 0.208 $\pm$ 0.004 & 0.203 $\pm$ 0.003 \\
      PrepTabPFN-2.5 & 0.202 $\pm$ 0.004 & \underline{0.197 $\pm$ 0.005} & \underline{\textbf{\textcolor{green!50!black}{0.195 $\pm$ 0.004}}} \\
      \bottomrule
      \end{tabular}
      }
    \end{minipage}
  \end{subtable} \hfill
  \begin{subtable}[t]{0.49\textwidth}
    \centering
    \scriptsize
    \caption*{in\_vehicle (AUC $\uparrow$)}
    \vspace{-1ex}    \label{tab:36}
    \begin{minipage}[t][][t]{\linewidth}
      \vspace{0pt}      \resizebox{\linewidth}{!}{%
      \begin{tabular}{l@{\quad}lll}
      \toprule
       & Default & Tuned & Tuned + Ens. \\
      \midrule
      Linear & 0.735 $\pm$ 0.012 & 0.735 $\pm$ 0.013 & 0.735 $\pm$ 0.013 \\
      KNN & 0.727 $\pm$ 0.009 & 0.777 $\pm$ 0.008 & 0.805 $\pm$ 0.007 \\
      ExtraTrees & 0.795 $\pm$ 0.010 & 0.802 $\pm$ 0.012 & 0.808 $\pm$ 0.012 \\
      RF & 0.809 $\pm$ 0.011 & 0.815 $\pm$ 0.013 & 0.819 $\pm$ 0.012 \\
      XGBoost & 0.830 $\pm$ 0.008 & 0.840 $\pm$ 0.008 & 0.840 $\pm$ 0.008 \\
      LightGBM & 0.833 $\pm$ 0.008 & 0.841 $\pm$ 0.009 & 0.842 $\pm$ 0.009 \\
      CatBoost & 0.838 $\pm$ 0.010 & 0.841 $\pm$ 0.009 & 0.842 $\pm$ 0.010 \\
      EBM & 0.809 $\pm$ 0.007 & 0.808 $\pm$ 0.006 & 0.812 $\pm$ 0.008 \\
      FastAIMLP & 0.809 $\pm$ 0.012 & 0.822 $\pm$ 0.011 & 0.825 $\pm$ 0.010 \\
      TorchMLP & 0.823 $\pm$ 0.005 & 0.827 $\pm$ 0.009 & 0.838 $\pm$ 0.009 \\
      MNCA & 0.810 $\pm$ 0.010 & 0.839 $\pm$ 0.009 & 0.846 $\pm$ 0.009 \\
      TabM & \underline{0.846 $\pm$ 0.008} & 0.849 $\pm$ 0.009 & 0.850 $\pm$ 0.009 \\
      RealMLP & 0.834 $\pm$ 0.007 & 0.845 $\pm$ 0.010 & 0.848 $\pm$ 0.009 \\
      xRFM & 0.818 $\pm$ 0.008 & 0.829 $\pm$ 0.009 & 0.837 $\pm$ 0.009 \\
      TabDPT & 0.812 $\pm$ 0.008 & 0.822 $\pm$ 0.006 & 0.824 $\pm$ 0.007 \\
      RealTabPFN-2.5 & 0.844 $\pm$ 0.010 & 0.842 $\pm$ 0.010 & 0.844 $\pm$ 0.009 \\
      SAP-RPT-OSS & 0.764 $\pm$ 0.010 & - & - \\
      TabPFN-2.6 & \underline{0.848 $\pm$ 0.010} & - & - \\
      TabICL-V2 & 0.840 $\pm$ 0.008 & - & - \\
      PrepLinear & 0.814 $\pm$ 0.010 & 0.830 $\pm$ 0.007 & 0.835 $\pm$ 0.009 \\
      PrepLightGBM & 0.842 $\pm$ 0.010 & 0.840 $\pm$ 0.009 & 0.847 $\pm$ 0.009 \\
      PrepTabM & 0.841 $\pm$ 0.011 & \underline{0.849 $\pm$ 0.009} & \underline{\textbf{\textcolor{green!50!black}{0.851 $\pm$ 0.010}}} \\
      PrepTabPFN-2.5 & 0.844 $\pm$ 0.010 & 0.842 $\pm$ 0.010 & \underline{\textbf{0.850 $\pm$ 0.010}} \\
      \bottomrule
      \end{tabular}
      }
    \end{minipage}
  \end{subtable}
  \medskip

\end{table}

\begin{table}[htb]
  \centering
  \begin{subtable}[t]{0.49\textwidth}
    \centering
    \scriptsize
    \caption*{jm1 (AUC $\uparrow$)}
    \vspace{-1ex}    \label{tab:37}
    \begin{minipage}[t][][t]{\linewidth}
      \vspace{0pt}      \resizebox{\linewidth}{!}{%
      \begin{tabular}{l@{\quad}lll}
      \toprule
       & Default & Tuned & Tuned + Ens. \\
      \midrule
      Linear & 0.724 $\pm$ 0.009 & 0.724 $\pm$ 0.009 & 0.724 $\pm$ 0.009 \\
      KNN & 0.742 $\pm$ 0.006 & 0.749 $\pm$ 0.006 & 0.761 $\pm$ 0.008 \\
      ExtraTrees & 0.758 $\pm$ 0.011 & 0.761 $\pm$ 0.014 & 0.768 $\pm$ 0.010 \\
      RF & 0.753 $\pm$ 0.012 & 0.755 $\pm$ 0.013 & 0.762 $\pm$ 0.012 \\
      XGBoost & 0.747 $\pm$ 0.013 & 0.751 $\pm$ 0.012 & 0.753 $\pm$ 0.012 \\
      LightGBM & 0.748 $\pm$ 0.012 & 0.752 $\pm$ 0.012 & 0.753 $\pm$ 0.011 \\
      CatBoost & 0.744 $\pm$ 0.010 & 0.751 $\pm$ 0.009 & 0.749 $\pm$ 0.009 \\
      EBM & 0.735 $\pm$ 0.011 & 0.736 $\pm$ 0.011 & 0.736 $\pm$ 0.012 \\
      FastAIMLP & 0.729 $\pm$ 0.013 & 0.727 $\pm$ 0.011 & 0.732 $\pm$ 0.012 \\
      TorchMLP & 0.726 $\pm$ 0.006 & 0.733 $\pm$ 0.007 & 0.734 $\pm$ 0.008 \\
      MNCA & 0.763 $\pm$ 0.003 & 0.761 $\pm$ 0.008 & 0.772 $\pm$ 0.007 \\
      TabM & 0.732 $\pm$ 0.012 & 0.737 $\pm$ 0.008 & 0.744 $\pm$ 0.011 \\
      RealMLP & 0.732 $\pm$ 0.014 & 0.747 $\pm$ 0.007 & 0.763 $\pm$ 0.007 \\
      xRFM & 0.742 $\pm$ 0.007 & 0.754 $\pm$ 0.008 & 0.763 $\pm$ 0.009 \\
      TabDPT & \underline{\textbf{0.778 $\pm$ 0.008}} & \underline{\textbf{\textcolor{green!50!black}{0.779 $\pm$ 0.009}}} & \underline{0.776 $\pm$ 0.008} \\
      RealTabPFN-2.5 & 0.758 $\pm$ 0.009 & 0.764 $\pm$ 0.010 & 0.766 $\pm$ 0.010 \\
      SAP-RPT-OSS & 0.743 $\pm$ 0.009 & - & - \\
      TabPFN-2.6 & 0.775 $\pm$ 0.011 & - & - \\
      TabICL-V2 & \underline{\textbf{0.776 $\pm$ 0.007}} & - & - \\
      PrepLinear & 0.763 $\pm$ 0.007 & 0.760 $\pm$ 0.010 & 0.766 $\pm$ 0.008 \\
      PrepLightGBM & 0.768 $\pm$ 0.009 & 0.769 $\pm$ 0.010 & 0.770 $\pm$ 0.009 \\
      PrepTabM & 0.762 $\pm$ 0.010 & 0.761 $\pm$ 0.009 & 0.761 $\pm$ 0.009 \\
      PrepTabPFN-2.5 & 0.758 $\pm$ 0.009 & 0.762 $\pm$ 0.010 & 0.772 $\pm$ 0.010 \\
      \bottomrule
      \end{tabular}
      }
    \end{minipage}
  \end{subtable} \hfill
  \begin{subtable}[t]{0.49\textwidth}
    \centering
    \scriptsize
    \caption*{kddcup09\_appetency (AUC $\uparrow$)}
    \vspace{-1ex}    \label{tab:38}
    \begin{minipage}[t][][t]{\linewidth}
      \vspace{0pt}      \resizebox{\linewidth}{!}{%
      %
      }
    \end{minipage}
  \end{subtable}
  \medskip

  \begin{subtable}[t]{0.49\textwidth}
    \centering
    \scriptsize
    \caption*{maternal\_health\_risk (logloss $\downarrow$)}
    \vspace{-1ex}    \label{tab:39}
    \begin{minipage}[t][][t]{\linewidth}
      \vspace{0pt}      \resizebox{\linewidth}{!}{%
      %
      }
    \end{minipage}
  \end{subtable} \hfill
  \begin{subtable}[t]{0.49\textwidth}
    \centering
    \scriptsize
    \caption*{miami\_housing (rmse $\downarrow$)}
    \vspace{-1ex}    \label{tab:40}
    \begin{minipage}[t][][t]{\linewidth}
      \vspace{0pt}      \resizebox{\linewidth}{!}{%
      %
      }
    \end{minipage}
  \end{subtable}
  \medskip

\end{table}

\begin{table}[htb]
  \centering
  \begin{subtable}[t]{0.49\textwidth}
    \centering
    \scriptsize
    \caption*{online\_shoppers (AUC $\uparrow$)}
    \vspace{-1ex}    \label{tab:41}
    \begin{minipage}[t][][t]{\linewidth}
      \vspace{0pt}      \resizebox{\linewidth}{!}{%
      %
      }
    \end{minipage}
  \end{subtable} \hfill
  \begin{subtable}[t]{0.49\textwidth}
    \centering
    \scriptsize
    \caption*{physiochemical\_protein (rmse $\downarrow$)}
    \vspace{-1ex}    \label{tab:42}
    \begin{minipage}[t][][t]{\linewidth}
      \vspace{0pt}      \resizebox{\linewidth}{!}{%
      %
      }
    \end{minipage}
  \end{subtable}
  \medskip

  \begin{subtable}[t]{0.49\textwidth}
    \centering
    \scriptsize
    \caption*{polish\_companies (AUC $\uparrow$)}
    \vspace{-1ex}    \label{tab:43}
    \begin{minipage}[t][][t]{\linewidth}
      \vspace{0pt}      \resizebox{\linewidth}{!}{%
      %
      }
    \end{minipage}
  \end{subtable} \hfill
  \begin{subtable}[t]{0.49\textwidth}
    \centering
    \scriptsize
    \caption*{qsar-biodeg (AUC $\uparrow$)}
    \vspace{-1ex}    \label{tab:44}
    \begin{minipage}[t][][t]{\linewidth}
      \vspace{0pt}      \resizebox{\linewidth}{!}{%
      %
      }
    \end{minipage}
  \end{subtable}
  \medskip

\end{table}

\begin{table}[htb]
  \centering
  \begin{subtable}[t]{0.49\textwidth}
    \centering
    \scriptsize
    \caption*{seismic-bumps (AUC $\uparrow$)}
    \vspace{-1ex}    \label{tab:45}
    \begin{minipage}[t][][t]{\linewidth}
      \vspace{0pt}      \resizebox{\linewidth}{!}{%
      %
      }
    \end{minipage}
  \end{subtable} \hfill
  \begin{subtable}[t]{0.49\textwidth}
    \centering
    \scriptsize
    \caption*{splice (logloss $\downarrow$)}
    \vspace{-1ex}    \label{tab:46}
    \begin{minipage}[t][][t]{\linewidth}
      \vspace{0pt}      \resizebox{\linewidth}{!}{%
      %
      }
    \end{minipage}
  \end{subtable}
  \medskip

  \begin{subtable}[t]{0.49\textwidth}
    \centering
    \scriptsize
    \caption*{students\_dropout (logloss $\downarrow$)}
    \vspace{-1ex}    \label{tab:47}
    \begin{minipage}[t][][t]{\linewidth}
      \vspace{0pt}      \resizebox{\linewidth}{!}{%
      %
      }
    \end{minipage}
  \end{subtable} \hfill
  \begin{subtable}[t]{0.49\textwidth}
    \centering
    \scriptsize
    \caption*{superconductivity (rmse $\downarrow$)}
    \vspace{-1ex}    \label{tab:48}
    \begin{minipage}[t][][t]{\linewidth}
      \vspace{0pt}      \resizebox{\linewidth}{!}{%
      %
      }
    \end{minipage}
  \end{subtable}
  \medskip

\end{table}

\begin{table}[htb]
  \centering
  \begin{subtable}[t]{0.49\textwidth}
    \centering
    \scriptsize
    \caption*{taiwanese\_bankruptcy (AUC $\uparrow$)}
    \vspace{-1ex}    \label{tab:49}
    \begin{minipage}[t][][t]{\linewidth}
      \vspace{0pt}      \resizebox{\linewidth}{!}{%
      %
      }
    \end{minipage}
  \end{subtable} \hfill
  \begin{subtable}[t]{0.49\textwidth}
    \centering
    \scriptsize
    \caption*{website\_phishing (logloss $\downarrow$)}
    \vspace{-1ex}    \label{tab:50}
    \begin{minipage}[t][][t]{\linewidth}
      \vspace{0pt}      \resizebox{\linewidth}{!}{%
      %
      }
    \end{minipage}
  \end{subtable}
  \medskip

  \begin{subtable}[t]{0.49\textwidth}
    \centering
    \scriptsize
    \caption*{wine\_quality (rmse $\downarrow$)}
    \vspace{-1ex}    \label{tab:51}
    \begin{minipage}[t][][t]{\linewidth}
      \vspace{0pt}      \resizebox{\linewidth}{!}{%
      %
      }
    \end{minipage}
  \end{subtable}
  \medskip

\end{table}

%% file: neurips_2026.bib
@article{bischl-arxiv17a,
  title        = {OpenML Benchmarking Suites and the OpenML100},
  author       = {B. Bischl and G. Casalicchio and M. Feurer and F. Hutter and M. Lang and R. Mantovani and J. N. van Rijn and J. Vanschoren},
  year         = 2019,
  journal      = {arXiv:1708.03731v1 [stat.ML]},
}

@article{bischl2017openml,
  title={Openml benchmarking suites},
  author={Bischl, Bernd and Casalicchio, Giuseppe and Feurer, Matthias and Gijsbers, Pieter and Hutter, Frank and Lang, Michel and Mantovani, Rafael G and van Rijn, Jan N and Vanschoren, Joaquin},
  journal={arXiv preprint arXiv:1708.03731},
  year={2017}
}

@article{breiman-mlj01a,
  title        = {Random Forests},
  author       = {L. Breiman},
  year         = 2001,
  journal      = mlj,
  volume       = 45,
  pages        = {5--32},
}

@inproceedings{chen2016xgboost,
  title={Xgboost: A scalable tree boosting system},
  author={Chen, Tianqi and Guestrin, Carlos},
  booktitle={Proceedings of the 22nd acm sigkdd international conference on knowledge discovery and data mining},
  pages={785--794},
  year={2016}
}

@article{erickson-arxiv20a,
  title        = {AutoGluon-Tabular: Robust and Accurate AutoML for Structured Data},
  author       = {N. Erickson and J. Mueller and A. Shirkov and H. Zhang and P. Larroy and M. Li and A. Smola},
  year         = 2020,
  journal      = {arXiv:2003.06505 [stat.ML]},
}

@article{geurts-ml06a,
  title        = {Extremely randomized trees},
  author       = {P. Geurts and D. Ernst and L. Wehenkel},
  year         = 2006,
  journal      = mlj,
  publisher    = springer,
  volume       = 63,
  number       = 1,
  pages        = {3--42},
}

@article{gorishniy2021revisiting,
  title={Revisiting deep learning models for tabular data},
  author={Gorishniy, Yury and Rubachev, Ivan and Khrulkov, Valentin and Babenko, Artem},
  journal={Advances in neural information processing systems},
  volume={34},
  pages={18932--18943},
  year={2021}
}

@article{grinsztajn2022tree,
  title={Why do tree-based models still outperform deep learning on typical tabular data?},
  author={Grinsztajn, L{\'e}o and Oyallon, Edouard and Varoquaux, Ga{\"e}l},
  journal={Advances in neural information processing systems},
  volume={35},
  pages={507--520},
  year={2022}
}

@article{guyon2003introduction,
  title={An introduction to variable and feature selection},
  author={Guyon, Isabelle and Elisseeff, Andr{\'e}},
  journal={Journal of machine learning research},
  volume={3},
  number={Mar},
  pages={1157--1182},
  year={2003}
}

@article{hollmann2022tabpfn,
  title={Tabpfn: A transformer that solves small tabular classification problems in a second},
  author={Hollmann, Noah and M{\"u}ller, Samuel and Eggensperger, Katharina and Hutter, Frank},
  journal={arXiv preprint arXiv:2207.01848},
  year={2022}
}

@article{hollmann-nature25a,
  title={Accurate predictions on small data with a tabular foundation model},
  author={N. Hollmann and S. M{\"u}ller and L. Purucker and A. Krishnakumar and M. K{\"o}rfer and Shi Bin Hoo and Robin Tibor Schirrmeister and Frank Hutter},
  journal={Nature},
  volume={637},
  number={8045},
  pages={319--326},
  year={2025},
  publisher={Nature Publishing Group UK London}
}

@inproceedings{kanter2015deep,
  title={Deep feature synthesis: Towards automating data science endeavors},
  author={Kanter, James Max and Veeramachaneni, Kalyan},
  booktitle={2015 IEEE international conference on data science and advanced analytics (DSAA)},
  pages={1--10},
  year={2015},
  organization={IEEE}
}

@article{ke2017lightgbm,
  title={Lightgbm: A highly efficient gradient boosting decision tree},
  author={Ke, Guolin and Meng, Qi and Finley, Thomas and Wang, Taifeng and Chen, Wei and Ma, Weidong and Ye, Qiwei and Liu, Tie-Yan},
  journal={Advances in neural information processing systems},
  volume={30},
  year={2017}
}

@inproceedings{kohli2024towards,
  title={Towards quantifying the effect of datasets for benchmarking: A look at tabular machine learning},
  author={Kohli, Ravin and Feurer, Matthias and Eggensperger, Katharina and Bischl, Bernd and Hutter, Frank},
  booktitle={ICLR Workshop},
  volume={2},
  pages={6},
  year={2024}
}

@article{mcelfresh2023neural,
  title={When do neural nets outperform boosted trees on tabular data?},
  author={McElfresh, Duncan and Khandagale, Sujay and Valverde, Jonathan and Prasad C, Vishak and Ramakrishnan, Ganesh and Goldblum, Micah and White, Colin},
  journal={Advances in Neural Information Processing Systems},
  volume={36},
  pages={76336--76369},
  year={2023}
}

@article{popov-arxiv19a,
  title        = {{N}eural {O}blivious {D}ecision {E}nsembles for {D}eep {L}earning on Tabular Data},
  author       = {S. Popov and S. Morozov and A. Babenko},
  year         = 2019,
  journal      = {arXiv:1909.06312v2 [cs.LG]},
}

@article{purucker2023assembled,
  title={Assembled-OpenML: creating efficient benchmarks for ensembles in AutoML with OpenML},
  author={Purucker, Lennart and Beel, Joeran},
  journal={arXiv preprint arXiv:2307.00285},
  year={2023}
}

@article{prokhorenkova2018catboost,
  title={CatBoost: unbiased boosting with categorical features},
  author={Prokhorenkova, Liudmila and Gusev, Gleb and Vorobev, Aleksandr and Dorogush, Anna Veronika and Gulin, Andrey},
  journal={Advances in neural information processing systems},
  volume={31},
  year={2018}
}

@article{scikit-learn,
  title        = {Scikit-learn: Machine Learning in {P}ython},
  author       = {F. Pedregosa and G. Varoquaux and A. Gramfort and V. Michel and B. Thirion and O. Grisel and M. Blondel and P. Prettenhofer and R. Weiss and V. Dubourg and J. Vanderplas and A. Passos and D. Cournapeau and M. Brucher and M. Perrot and E. Duchesnay},
  year         = 2011,
  journal      = jmlr,
  volume       = 12,
  pages        = {2825--2830},
}

@TechReport{mitchell80,
  author = 	 "T. M. Mitchell",
  title = 	 "The Need for Biases in Learning Generalizations",
  institution =  "Computer Science Department, Rutgers University",
  year = 	 "1980",
  address =	 "New Brunswick, MA",
}

@article{holzmuller2024better,
  title={Better by default: Strong pre-tuned mlps and boosted trees on tabular data},
  author={Holzm{\"u}ller, David and Grinsztajn, L{\'e}o and Steinwart, Ingo},
  journal={Advances in Neural Information Processing Systems},
  volume={37},
  pages={26577--26658},
  year={2024}
}

@inproceedings{salinas2024tabrepo,
  title={TabRepo: A Large Scale Repository of Tabular Model Evaluations and its AutoML Applications},
  author={Salinas, David and Erickson, Nick},
  booktitle={AutoML Conference 2024 (ABCD Track)},
  year={2024}
}

@inproceedings{tschalzev2025unreflected,
  title={Unreflected Use of Tabular Data Repositories Can Undermine Research Quality},
  author={Tschalzev, Andrej and Purucker, Lennart and L{\"u}dtke, Stefan and Hutter, Frank and Bartelt, Christian and Stuckenschmidt, Heiner},
  booktitle={The Future of Machine Learning Data Practices and Repositories at ICLR 2025},
    year={2025}
}

@inproceedings{tschalzev2024data,
  title={A Data-Centric Perspective on Evaluating Machine Learning Models for Tabular Data},
  author={Tschalzev, Andrej and Marton, Sascha and L{\"u}dtke, Stefan and Bartelt, Christian and Stuckenschmidt, Heiner},
  booktitle={The Thirty-eight Conference on Neural Information Processing Systems Datasets and Benchmarks Track},
year={2024}
}

@article{rubachev2024tabred,
  title={TabReD: Analyzing Pitfalls and Filling the Gaps in Tabular Deep Learning Benchmarks},
  author={Rubachev, Ivan and Kartashev, Nikolay and Gorishniy, Yury and Babenko, Artem},
  journal={arXiv preprint arXiv:2406.19380},
  year={2024}
}

@inproceedings{zeng2024tabflex,
  title={Tabflex: Scaling tabular learning to millions with linear attention},
  author={Zeng, Yuchen and Kang, Wonjun and Mueller, Andreas C},
  booktitle={NeurIPS 2024 Third Table Representation Learning Workshop},
  year={2024}
}

@article{zabergja2024tabular,
  title={Is Deep Learning finally better than Decision Trees on Tabular Data?},
  author={Zab{\"e}rgja, Guri and Kadra, Arlind and Frey, Christian and Grabocka, Josif},
  journal={arXiv preprint arXiv:2402.03970},
  year={2024}
}

@article{shmuel2024comprehensive,
  title={A Comprehensive Benchmark of Machine and Deep Learning Across Diverse Tabular Datasets},
  author={Shmuel, Assaf and Glickman, Oren and Lazebnik, Teddy},
  journal={arXiv preprint arXiv:2408.14817},
  year={2024}
}

@article{gorishniy2024tabm,
  title={TabM: Advancing tabular deep learning with parameter-efficient ensembling},
  author={Gorishniy, Yury and Kotelnikov, Akim and Babenko, Artem},
  journal={arXiv preprint arXiv:2410.24210},
  year={2024}
}

@article{qu2025tabicl,
  title={TabICL: A Tabular Foundation Model for In-Context Learning on Large Data},
  author={Qu, Jingang and Holzm{\"u}ller, David and Varoquaux, Ga{\"e}l and Morvan, Marine Le},
  journal={arXiv preprint arXiv:2502.05564},
  year={2025}
}

@article{ma2024tabdpt,
  title={Tabdpt: Scaling tabular foundation models},
  author={Ma, Junwei and Thomas, Valentin and Hosseinzadeh, Rasa and Kamkari, Hamidreza and Labach, Alex and Cresswell, Jesse C and Golestan, Keyvan and Yu, Guangwei and Volkovs, Maksims and Caterini, Anthony L},
  journal={arXiv preprint arXiv:2410.18164},
  year={2024}
}

@inproceedings{lou2013accurate,
  title={Accurate intelligible models with pairwise interactions},
  author={Lou, Yin and Caruana, Rich and Gehrke, Johannes and Hooker, Giles},
  booktitle={Proceedings of the 19th ACM SIGKDD international conference on Knowledge discovery and data mining},
  pages={623--631},
  year={2013}
}

@inproceedings{gardner2024large,
  title={Large Scale Transfer Learning for Tabular Data via Language Modeling},
  author={Gardner, Joshua P and Perdomo, Juan Carlos and Schmidt, Ludwig},
  booktitle={The Thirty-eighth Annual Conference on Neural Information Processing Systems},
  year={2024},
}

@article{thielmann2024efficiency,
  title={On the Efficiency of NLP-Inspired Methods for Tabular Deep Learning},
  author={Thielmann, Anton Frederik and Samiee, Soheila},
  journal={arXiv preprint arXiv:2411.17207},
  year={2024}
}

@article{ye2024modern,
  title={Modern neighborhood components analysis: A deep tabular baseline two decades later},
  author={Ye, Han-Jia and Yin, Huai-Hong and Zhan, De-Chuan},
  journal={arXiv preprint arXiv:2407.03257},
  year={2024}
}

@inproceedings{bordtelephants,
  title={Elephants Never Forget: Memorization and Learning of Tabular Data in Large Language Models},
  author={Bordt, Sebastian and Nori, Harsha and Vasconcelos, Vanessa Cristiny Rodrigues and Nushi, Besmira and Caruana, Rich},
  booktitle={First Conference on Language Modeling},
    year={2024},
}

@misc{tunguz2023kaggle,
  author       = {Bojan Tunguz and Dieter and Heads or Tails and Karnika Kapoor and Parul Pandey and Paul Mooney and Phil Culliton and Rob Mulla and Sanyam Bhutani and Will Cukierski},
  title        = {2023 Kaggle AI Report},
  year         = 2023,
  publisher    = {Kaggle},
  url          = {https://kaggle.com/competitions/2023-kaggle-ai-report}
}

@article{joseph2021pytorch,
  title={Pytorch tabular: A framework for deep learning with tabular data},
  author={Joseph, Manu},
  journal={arXiv preprint arXiv:2104.13638},
  year={2021}
}

@article{spinaci2025contexttab,
  title={ConTextTab: A Semantics-Aware Tabular In-Context Learner},
  author={Spinaci, Marco and Polewczyk, Marek and Schambach, Maximilian and Thelin, Sam},
  journal={arXiv preprint arXiv:2506.10707},
  year={2025}
}

@article{arazi2025tabstar,
  title={TabSTAR: A Foundation Tabular Model With Semantically Target-Aware Representations},
  author={Arazi, Alan and Shapira, Eilam and Reichart, Roi},
  journal={arXiv preprint arXiv:2505.18125},
  year={2025}
}

@article{bischl2025openml,
  title={OpenML: Insights from 10 years and more than a thousand papers},
  author={Bischl, Bernd and Casalicchio, Giuseppe and Das, Taniya and Feurer, Matthias and Fischer, Sebastian and Gijsbers, Pieter and Mukherjee, Subhaditya and M{\"u}ller, Andreas C and N{\'e}meth, L{\'a}szl{\'o} and Oala, Luis and others},
  journal={Patterns},
  year={2025},
  publisher={Elsevier}
}

@article{schafer2025usable,
  title={How Usable is Automated Feature Engineering for Tabular Data?},
  author={Sch{\"a}fer, Bastian and Purucker, Lennart and Janowski, Maciej and Hutter, Frank},
  journal={arXiv preprint arXiv:2508.13932},
  year={2025}
}

@article{kuken2024large,
  title={Large language models engineer too many simple features for tabular data},
  author={K{\"u}ken, Jaris and Purucker, Lennart and Hutter, Frank},
  journal={arXiv preprint arXiv:2410.17787},
  year={2024}
}

@inproceedings{zhang2023openfe,
  title={Openfe: Automated feature generation with expert-level performance},
  author={Zhang, Tianping and Zhang, Zheyu Aqa and Fan, Zhiyuan and Luo, Haoyan and Liu, Fengyuan and Liu, Qian and Cao, Wei and Jian, Li},
  booktitle={International Conference on Machine Learning},
  pages={41880--41901},
  year={2023},
  organization={PMLR}
}

@article{erickson2025tabarena,
  title={Tabarena: A living benchmark for machine learning on tabular data},
  author={Erickson, Nick and Purucker, Lennart and Tschalzev, Andrej and Holzm{\"u}ller, David and Desai, Prateek Mutalik and Salinas, David and Hutter, Frank},
  journal={arXiv preprint arXiv:2506.16791},
  year={2025}
}

@article{verdonck2024special,
  title={Special issue on feature engineering editorial},
  author={Verdonck, Tim and Baesens, Bart and {\'O}skarsd{\'o}ttir, Mar{\'\i}a and vanden Broucke, Seppe},
  journal={Machine learning},
  volume={113},
  number={7},
  pages={3917--3928},
  year={2024},
  publisher={Springer}
}

@article{madsen2020neural,
  title={Neural arithmetic units},
  author={Madsen, Andreas and Johansen, Alexander Rosenberg},
  journal={arXiv preprint arXiv:2001.05016},
  year={2020}
}

@article{trask2018neural,
  title={Neural arithmetic logic units},
  author={Trask, Andrew and Hill, Felix and Reed, Scott E and Rae, Jack and Dyer, Chris and Blunsom, Phil},
  journal={Advances in neural information processing systems},
  volume={31},
  year={2018}
}

@article{micci2001preprocessing,
  title={A preprocessing scheme for high-cardinality categorical attributes in classification and prediction problems},
  author={Micci-Barreca, Daniele},
  journal={ACM SIGKDD explorations newsletter},
  volume={3},
  number={1},
  pages={27--32},
  year={2001},
  publisher={ACM New York, NY, USA}
}

@article{marton2023grande,
  title={GRANDE: Gradient-based decision tree ensembles for tabular data},
  author={Marton, Sascha and L{\"u}dtke, Stefan and Bartelt, Christian and Stuckenschmidt, Heiner},
  journal={arXiv preprint arXiv:2309.17130},
  year={2023}
}

@article{pargent2022regularized,
  title={Regularized target encoding outperforms traditional methods in supervised machine learning with high cardinality features},
  author={Pargent, Florian and Pfisterer, Florian and Thomas, Janek and Bischl, Bernd},
  journal={Computational Statistics},
  volume={37},
  number={5},
  pages={2671--2692},
  year={2022},
  publisher={Springer}
}

@article{baxter2000model,
  title={A model of inductive bias learning},
  author={Baxter, Jonathan},
  journal={Journal of artificial intelligence research},
  volume={12},
  pages={149--198},
  year={2000}
}

@article{hornik1989multilayer,
  title={Multilayer feedforward networks are universal approximators},
  author={Hornik, Kurt and Stinchcombe, Maxwell and White, Halbert},
  journal={Neural networks},
  volume={2},
  number={5},
  pages={359--366},
  year={1989},
  publisher={Elsevier}
}

@article{gardner2023benchmarking,
  title={Benchmarking distribution shift in tabular data with tableshift},
  author={Gardner, Josh and Popovic, Zoran and Schmidt, Ludwig},
  journal={Advances in Neural Information Processing Systems},
  volume={36},
  pages={53385--53432},
  year={2023}
}

@article{tschalzev2024enabling,
  title={Enabling mixed effects neural networks for diverse, clustered data using Monte Carlo methods},
  author={Tschalzev, Andrej and Nitschke, Paul and Kirchdorfer, Lukas and L{\"u}dtke, Stefan and Bartelt, Christian and Stuckenschmidt, Heiner},
  journal={arXiv preprint arXiv:2407.01115},
  year={2024}
}

@article{morvan2024imputation,
  title={Imputation for prediction: beware of diminishing returns},
  author={Morvan, Marine Le and Varoquaux, Ga{\"e}l},
  journal={arXiv preprint arXiv:2407.19804},
  year={2024}
}

@article{liu2025climb,
  title={CLIMB: Class-imbalanced Learning Benchmark on Tabular Data},
  author={Liu, Zhining and Li, Zihao and Yang, Ze and Wei, Tianxin and Kang, Jian and Zhu, Yada and Hamann, Hendrik and He, Jingrui and Tong, Hanghang},
  journal={arXiv preprint arXiv:2505.17451},
  year={2025}
}

@inproceedings{horn2019autofeat,
  title={The autofeat python library for automated feature engineering and selection},
  author={Horn, Franziska and Pack, Robert and Rieger, Michael},
  booktitle={Joint European Conference on Machine Learning and Knowledge Discovery in Databases},
  pages={111--120},
  year={2019},
  organization={Springer}
}

@article{galli2021feature,
  title={Feature-engine: A Python package for feature engineering for machine learning},
  author={Galli, Soledad},
  journal={Journal of Open Source Software},
  volume={6},
  number={65},
  pages={3642},
  year={2021}
}

@article{mcginnis2018category,
  title={Category encoders: a scikit-learn-contrib package of transformers for encoding categorical data},
  author={McGinnis, William D and Siu, Chapman and Huang, Hanyu and others},
  journal={Journal of Open Source Software},
  volume={3},
  number={21},
  pages={501},
  year={2018}
}

@article{hollmann2023large,
  title={Large language models for automated data science: Introducing caafe for context-aware automated feature engineering},
  author={Hollmann, Noah and M{\"u}ller, Samuel and Hutter, Frank},
  journal={Advances in Neural Information Processing Systems},
  volume={36},
  pages={44753--44775},
  year={2023}
}

@article{abhyankar2025llm,
  title={LLM-FE: Automated Feature Engineering for Tabular Data with LLMs as Evolutionary Optimizers},
  author={Abhyankar, Nikhil and Shojaee, Parshin and Reddy, Chandan K},
  journal={arXiv preprint arXiv:2503.14434},
  year={2025}
}

@incollection{galecki2012linear,
  title={Linear mixed-effects model},
  author={Ga{\l}ecki, Andrzej and Burzykowski, Tomasz},
  booktitle={Linear mixed-effects models using R: a step-by-step approach},
  pages={245--273},
  year={2012},
  publisher={Springer}
}

@book{zuur2009mixed,
  title={Mixed effects models and extensions in ecology with R},
  author={Zuur, Alain F and Ieno, Elena N and Walker, Neil J and Saveliev, Anatoly A and Smith, Graham M and others},
  volume={574},
  year={2009},
  publisher={Springer}
}

@article{bakheet2023hybrid,
  title={Hybrid bag-of-visual-words and featurewiz selection for content-based visual information retrieval},
  author={Bakheet, Samy and Al-Hamadi, Ayoub and Soliman, Emadeldeen and Heshmat, Mohamed},
  journal={Sensors},
  volume={23},
  number={3},
  pages={1653},
  year={2023},
  publisher={Multidisciplinary Digital Publishing Institute}
}

@article{bonidia2022bioautoml,
  title={BioAutoML: automated feature engineering and metalearning to predict noncoding RNAs in bacteria},
  author={Bonidia, Robson P and Santos, Anderson P Avila and de Almeida, Breno LS and Stadler, Peter F and da Rocha, Ulisses N and Sanches, Danilo S and de Carvalho, Andr{\'e} CPLF},
  journal={Briefings in Bioinformatics},
  volume={23},
  number={4},
  pages={bbac218},
  year={2022},
  publisher={Oxford University Press}
}

@article{kursa2010feature,
  title={Feature selection with the Boruta package},
  author={Kursa, Miron B and Rudnicki, Witold R},
  journal={Journal of statistical software},
  volume={36},
  pages={1--13},
  year={2010}
}

@inproceedings{yu2003feature,
  title={Feature selection for high-dimensional data: A fast correlation-based filter solution},
  author={Yu, Lei and Liu, Huan},
  booktitle={Proceedings of the 20th international conference on machine learning (ICML-03)},
  pages={856--863},
  year={2003}
}

@inproceedings{lin2022adafs,
  title={AdaFS: Adaptive feature selection in deep recommender system},
  author={Lin, Weilin and Zhao, Xiangyu and Wang, Yejing and Xu, Tong and Wu, Xian},
  booktitle={Proceedings of the 28th ACM SIGKDD conference on knowledge discovery and data mining},
  pages={3309--3317},
  year={2022}
}

@article{grinsztajn2025tabpfn,
  title={TabPFN-2.5: Advancing the state of the art in tabular foundation models},
  author={Grinsztajn, L{\'e}o and Fl{\"o}ge, Klemens and Key, Oscar and Birkel, Felix and Jund, Philipp and Roof, Brendan and J{\"a}ger, Benjamin and Safaric, Dominik and Alessi, Simone and Hayler, Adrian and others},
  journal={arXiv preprint arXiv:2511.08667},
  year={2025}
}

@article{van2014endgame,
  title={Endgame analysis of dou shou qi},
  author={van Rijn, Jan N and Vis, Jonathan K},
  journal={ICGA journal},
  volume={37},
  number={2},
  pages={120--124},
  year={2014},
  publisher={SAGE Publications Sage UK: London, England}
}

@article{siebert1987vehicle,
  title={Vehicle recognition using rule based methods},
  author={Siebert, J Paul},
  year={1987},
  publisher={Turing Institute}
}

@article{shultz1994modeling,
  title={Modeling cognitive development on balance scale phenomena},
  author={Shultz, Thomas R and Mareschal, Denis and Schmidt, William C},
  journal={Machine learning},
  volume={16},
  number={1},
  pages={57--86},
  year={1994},
  publisher={Springer}
}

@inproceedings{botta1991learning,
  title={Learning quantitative features in a symbolic environment},
  author={Botta, Marco and Giordana, Attilio},
  booktitle={International Symposium on Methodologies for Intelligent Systems},
  pages={296--305},
  year={1991},
  organization={Springer}
}

@article{zhang2025limix,
  title={Limix: Unleashing structured-data modeling capability for generalist intelligence},
  author={Zhang, Xingxuan and Ren, Gang and Yu, Han and Yuan, Hao and Wang, Hui and Li, Jiansheng and Wu, Jiayun and Mo, Lang and Mao, Li and Hao, Mingchao and others},
  journal={arXiv preprint arXiv:2509.03505},
  year={2025}
}

@inproceedings{shi2020safe,
  title={Safe: Scalable automatic feature engineering framework for industrial tasks},
  author={Shi, Qitao and Zhang, Ya-Lin and Li, Longfei and Yang, Xinxing and Li, Meng and Zhou, Jun},
  booktitle={2020 IEEE 36th International Conference on Data Engineering (ICDE)},
  pages={1645--1656},
  year={2020},
  organization={IEEE}
}

@inproceedings{lam2021automated,
  title={Automated data science for relational data},
  author={Lam, Hoang Thanh and Buesser, Beat and Min, Hong and Minh, Tran Ngoc and Wistuba, Martin and Khurana, Udayan and Bramble, Gregory and Salonidis, Theodoros and Wang, Dakuo and Samulowitz, Horst},
  booktitle={2021 IEEE 37th International Conference on Data Engineering (ICDE)},
  pages={2689--2692},
  year={2021},
  organization={IEEE}
}

@inproceedings{fan2010generalized,
  title={Generalized and heuristic-free feature construction for improved accuracy},
  author={Fan, Wei and Zhong, Erheng and Peng, Jing and Verscheure, Olivier and Zhang, Kun and Ren, Jiangtao and Yan, Rong and Yang, Qiang},
  booktitle={Proceedings of the 2010 SIAM International Conference on Data Mining},
  pages={629--640},
  year={2010},
  organization={SIAM}
}

@inproceedings{nargesian2017learning,
  title={Learning feature engineering for classification.},
  author={Nargesian, Fatemeh and Samulowitz, Horst and Khurana, Udayan and Khalil, Elias B and Turaga, Deepak S},
  booktitle={Ijcai},
  volume={17},
  pages={2529--2535},
  year={2017}
}

@article{yuanfei2019autocross,
  title={Autocross: Automatic feature crossing for tabular data in real-world applications},
  author={Yuanfei, Luo and Mengshuo, Wang and Hao, Zhou and Quanming, Yao and WeiWei, Tu and Yuqiang, Chen and Qiang, Yang and Wenyuan, Dai},
  journal={arXiv preprint arXiv:1904.12857},
  year={2019}
}

@article{nogueira2021investigating,
  title={Investigating the limitations of transformers with simple arithmetic tasks},
  author={Nogueira, Rodrigo and Jiang, Zhiying and Lin, Jimmy},
  journal={arXiv preprint arXiv:2102.13019},
  year={2021}
}

@article{qu2026tabiclv2,
  title={TabICLv2: A better, faster, scalable, and open tabular foundation model},
  author={Qu, Jingang and Holzm{\"u}ller, David and Varoquaux, Ga{\"e}l and Morvan, Marine Le},
  journal={arXiv preprint arXiv:2602.11139},
  year={2026}
}

@misc{priorlabs_tabpfn_2_6,
  title        = {TabPFN-2.6},
  author       = {{Prior Labs}},
  year         = {2025},
  howpublished = {\url{https://huggingface.co/Prior-Labs/tabpfn_2_6}},
  note         = {Hugging Face model card, version v1.0}
}

@article{zhang2025mitra,
  title={Mitra: Mixed synthetic priors for enhancing tabular foundation models},
  author={Zhang, Xiyuan and Maddix, Danielle C and Yin, Junming and Erickson, Nick and Ansari, Abdul Fatir and Han, Boran and Zhang, Shuai and Akoglu, Leman and Faloutsos, Christos and Mahoney, Michael W and others},
  journal={arXiv preprint arXiv:2510.21204},
  year={2025}
}

@article{liu2025tabpfn,
  title={Tabpfn unleashed: A scalable and effective solution to tabular classification problems},
  author={Liu, Si-Yang and Ye, Han-Jia},
  journal={arXiv preprint arXiv:2502.02527},
  year={2025}
}

@article{beaglehole2025xrfm,
  title={xRFM: Accurate, scalable, and interpretable feature learning models for tabular data},
  author={Beaglehole, Daniel and Holzm{\"u}ller, David and Radhakrishnan, Adityanarayanan and Belkin, Mikhail},
  journal={arXiv preprint arXiv:2508.10053},
  year={2025}
}

@inproceedings{gan2024graph,
  title={Graph machine learning meets multi-table relational data},
  author={Gan, Quan and Wang, Minjie and Wipf, David and Faloutsos, Christos},
  booktitle={Proceedings of the 30th ACM SIGKDD Conference on Knowledge Discovery and Data Mining},
  pages={6502--6512},
  year={2024}
}

@inproceedings{leskovec2023databases,
  title={Databases as Graphs: Predictive Queries for Declarative Machine Learning},
  author={Leskovec, Jure},
  booktitle={Proceedings of the 42nd ACM SIGMOD-SIGACT-SIGAI Symposium on Principles of Database Systems},
  pages={1--1},
  year={2023}
}

@misc{fey2025kumorfm,
  title={Kumorfm: A foundation model for in-context learning on relational data},
  author={Fey, Matthias and Kocijan, Vid and Lopez, Federico and Lenssen, J and Leskovec, Jure},
  year={2025}
}

@article{cucumides2025features,
  title={From Features to Structure: Task-Aware Graph Construction for Relational and Tabular Learning with GNNs},
  author={Cucumides, Tamara and Geerts, Floris},
  journal={arXiv preprint arXiv:2506.02243},
  year={2025}
}

@article{bolker2009generalized,
  title={Generalized linear mixed models: a practical guide for ecology and evolution},
  author={Bolker, Benjamin M and Brooks, Mollie E and Clark, Connie J and Geange, Shane W and Poulsen, John R and Stevens, M Henry H and White, Jada-Simone S},
  journal={Trends in ecology \& evolution},
  volume={24},
  number={3},
  pages={127--135},
  year={2009},
  publisher={Elsevier}
}

@misc{bates2010lme4,
  title={lme4: Mixed-effects modeling with R},
  author={Bates, Douglas M},
  year={2010},
  publisher={Springer New York}
}

@article{moslemi2023tutorial,
  title={A tutorial-based survey on feature selection: Recent advancements on feature selection},
  author={Moslemi, Amir},
  journal={Engineering applications of artificial intelligence},
  volume={126},
  pages={107136},
  year={2023},
  publisher={Elsevier}
}

@article{barbieri2024analysis,
  title={Analysis and comparison of feature selection methods towards performance and stability},
  author={Barbieri, Matheus Cezimbra and Grisci, Bruno Iochins and Dorn, M{\'a}rcio},
  journal={Expert Systems with Applications},
  volume={249},
  pages={123667},
  year={2024},
  publisher={Elsevier}
}

@inproceedings{yamada2020feature,
  title={Feature selection using stochastic gates},
  author={Yamada, Yutaro and Lindenbaum, Ofir and Negahban, Sahand and Kluger, Yuval},
  booktitle={International conference on machine learning},
  pages={10648--10659},
  year={2020},
  organization={PMLR}
}

@inproceedings{balin2019concrete,
  title={Concrete autoencoders: Differentiable feature selection and reconstruction},
  author={Bal{\i}n, Muhammed Fatih and Abid, Abubakar and Zou, James},
  booktitle={International conference on machine learning},
  pages={444--453},
  year={2019},
  organization={PMLR}
}

@article{cherepanova2023performance,
  title={A performance-driven benchmark for feature selection in tabular deep learning},
  author={Cherepanova, Valeriia and Levin, Roman and Somepalli, Gowthami and Geiping, Jonas and Bruss, C Bayan and Wilson, Andrew G and Goldstein, Tom and Goldblum, Micah},
  journal={Advances in Neural Information Processing Systems},
  volume={36},
  pages={41956--41979},
  year={2023}
}
